\documentclass[10pt,journal,compsoc]{IEEEtran}

\usepackage{epsfig}
\usepackage{graphicx}
\usepackage{amsmath}
\usepackage{amssymb}
\usepackage[dvipsnames,svgnames]{xcolor}
\usepackage{multirow}
\usepackage{bbding}
\usepackage[utf8]{inputenc}

\DeclareMathOperator*{\argmax}{argmax}
\DeclareMathOperator*{\argmin}{argmin}
\usepackage{diagbox}
\usepackage{booktabs}
\usepackage{colortbl}
\def\uline{\underline}

\usepackage{comment}
\usepackage{enumitem}
\usepackage{mathrsfs}
\usepackage[outline]{contour}
\usepackage{amssymb}
\usepackage{framed}
\usepackage{stfloats}
\usepackage{microtype}
\usepackage[pagebackref=true,breaklinks=true,letterpaper=true,colorlinks,bookmarks=false]{hyperref}
\usepackage{pifont}% http://ctan.org/pkg/pifont
\newcommand{\xmark}{\ding{55}}%

\newcommand{\ie}{{\it i.e.}}
\newcommand{\eg}{{\it e.g.}}
\newcommand{\etal}{{\it et al.}}
\newcommand{\etc}{{\it etc.}}

\renewcommand{\textcolor}[2]{{#2}}

\newcommand{\tabname}{Table}
\newcommand{\secname}{Section}

\begin{document}
\title{Adversarial Attack and Defense in Deep Ranking}

\author{Mo~Zhou,~%~\IEEEmembership{Member,~IEEE,}
        Le~Wang,~\IEEEmembership{Senior Member,~IEEE,}
        Zhenxing~Niu,~\IEEEmembership{Member,~IEEE,}
		Qilin~Zhang,~\IEEEmembership{Member,~IEEE,}
		Nanning Zheng,~\IEEEmembership{Fellow,~IEEE,}
        and~Gang~Hua,~\IEEEmembership{Fellow,~IEEE}% <-this % stops a space
\IEEEcompsocitemizethanks{
\IEEEcompsocthanksitem
Mo Zhou, Le Wang, and Nanning Zheng are with the Institute of Artificial Intelligence
and Robotics, Xi'an Jiaotong University, Xi'an, Shaanxi 710049, China.
(Corresponding author: Le Wang.)\protect\\
Email: cdluminate@gmail.com, \{lewang,nnzheng\}@mail.xjtu.edu.cn.
\IEEEcompsocthanksitem
Zhenxing Niu is with Alibaba Group, Hangzhou, Zhejiang 311121, China.
Email: zhenxingniu@gmail.com.
\IEEEcompsocthanksitem
Qilin Zhang is with ABB Corporate Research Center, Raleigh, NC 27606, USA.
This research work was carried out before his joining of ABB.
Email: samqzhang@gmail.com.
\IEEEcompsocthanksitem
Gang Hua is with Wormpex AI Research, Bellevue, WA 98004, USA.
Email: ganghua@gmail.com}%
%\IEEEcompsocthanksitem Manuscript received MM.DD, 2021; revised MM.DD, 2021.}% <-this % stops an unwanted space
}

\markboth{Journal of \LaTeX\ Class Files,~Vol.~xx, No.~xx, xx~xx}%
%\markboth{IEEE Transactions on Pattern Analysis and Machine Intelligence,~Vol.~xx, No.~xx, xx~xx}%
{Zhou \textit{et al.}: Adversarial Attack and Defense in Deep Ranking}

\IEEEtitleabstractindextext{%
\begin{abstract}
\label{sec:abs}
Deep Neural Network classifiers are vulnerable to adversarial attack, where an
	imperceptible perturbation could result in misclassification.
However, the vulnerability of DNN-based image ranking systems remains
	under-explored.
In this paper, we propose two attacks against deep ranking systems, \emph{i.e.},
	Candidate Attack and Query Attack, that can raise or lower the rank of
	chosen candidates by adversarial perturbations.
Specifically, the expected ranking order is first represented as a set of
	inequalities, and then a triplet-like objective function is designed to
	obtain the optimal perturbation.
\textcolor{blue}{Conversely, an anti-collapse triplet defense
	is proposed to improve the ranking model robustness against all proposed
	attacks, where the model learns to prevent the positive and negative
	samples being pulled close to each other by adversarial attack.}
\textcolor{blue}{To comprehensively measure the empirical adversarial robustness
	of a ranking model with our defense, we propose an empirical
	robustness score, which involves a set of representative attacks against
	ranking models.}
Our adversarial ranking attacks and defenses are evaluated on
	MNIST, Fashion-MNIST, CUB200-2011, CARS196 and Stanford Online Products datasets.
Experimental results demonstrate that a typical deep ranking system can be
	effectively compromised by our attacks.
\textcolor{blue}{Nevertheless, our defense can significantly improve the ranking
	system robustness, and simultaneously mitigate a wide range of attacks.}
\end{abstract}

}

\maketitle

% from template
\IEEEdisplaynontitleabstractindextext

% from template
\IEEEpeerreviewmaketitle

\IEEEraisesectionheading{\section{Introduction}\label{sec:introduction}}
\label{sec:1}

% 1. deep neural networks are vulnerable to adversarial attack

\IEEEPARstart{D}{espite} the successful application in computer vision tasks
such as image classification~\cite{resnet}, Deep Neural Networks (DNNs)
have been found vulnerable to adversarial attacks.
In particular, the DNN's prediction can be arbitrarily changed by just applying
an imperceptible perturbation to the input image~\cite{l-bfgs,fgsm}.
Moreover, such adversarial attacks can effectively compromise the recent
state-of-the-art DNNs such as Inception~\cite{inceptionv2} and
ResNet~\cite{resnet}.
This poses a serious security risk on many DNN-based applications such as face
recognition, where recognition evasion and impersonation can be easily
achieved~\cite{faceblack,phy-crime,advpattern}.

% 2. from classification to ranking, necessity to study, examples

\begin{figure}[t!]
\centering
\includegraphics[width=1.0\columnwidth]{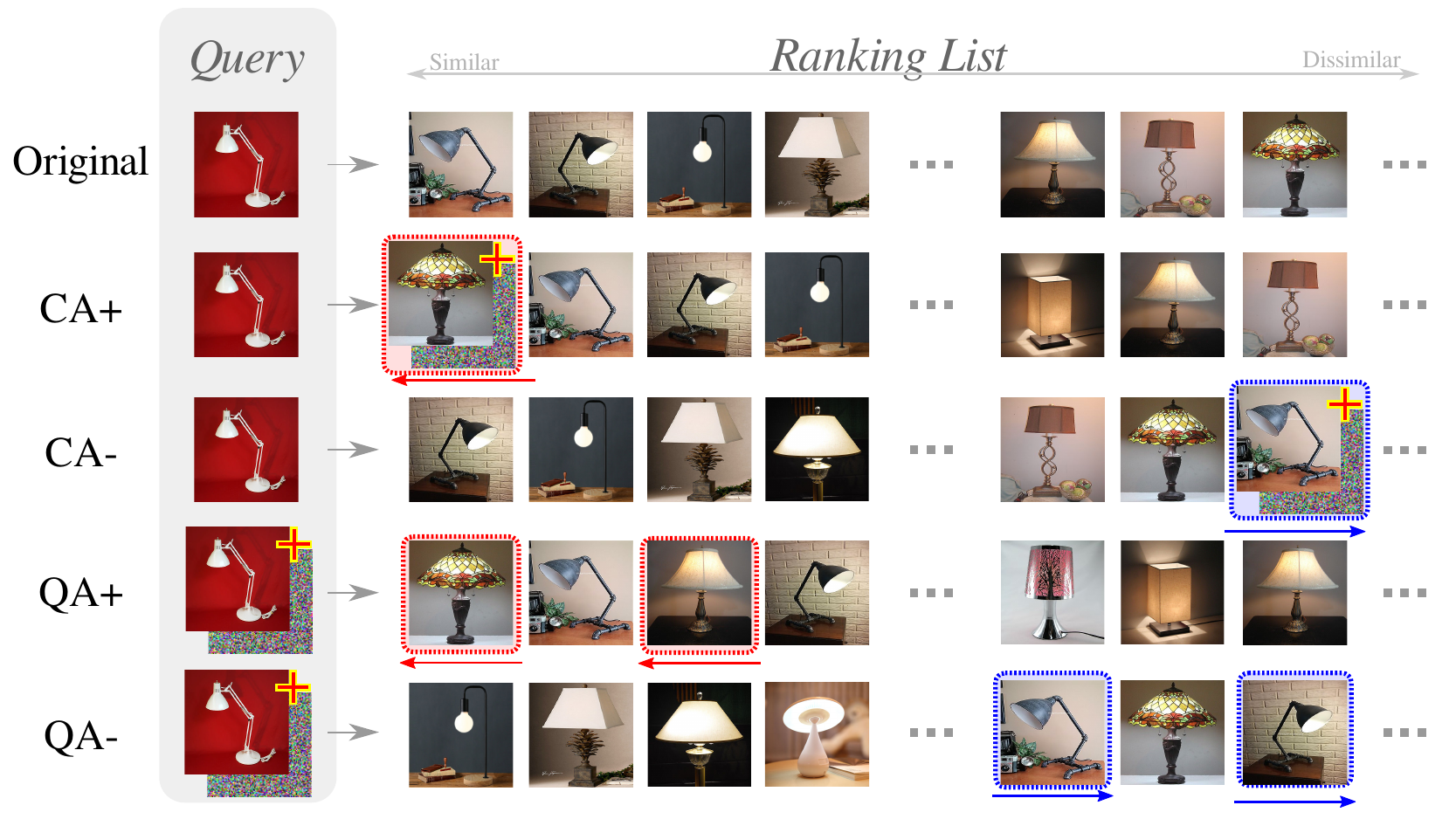}
	\caption{Adversarial ranking attack that can \emph{raise} or \emph{lower}
	the rank of the chosen candidates by adversarial perturbations. In Candidate
	Attack, adversarial perturbation is added to the candidate image, and its rank
	will be \emph{raised} (CA+) or \emph{lowered} (CA-).
	In Query Attack, adversarial perturbation is added to the query image,
	and the ranks of chosen candidates will be \emph{raised} (QA+)
	or \emph{lowered} (QA-).
	\textcolor{blue}{An ideal robust ranking model should be resistant to
	any of these attacks.}
	}
\label{fig:advranking}
\end{figure}

Previous adversarial attacks primarily focus on \emph{classification}.
However, we speculate that DNN-based image ranking
systems~\cite{imagesimilarity,imagesim2}
also suffer from similar vulnerability.
Taking the image-based product search as an example, a fair ranking system
should rank the database products according to their visual similarity to the
query, as shown in \figurename~\ref{fig:advranking} (row 1).
Nevertheless, malicious sellers may attempt to raise the rank of their own
product by adding perturbation to the image (CA+, row 2), or lower the rank of
their competitor’s product (CA-, row 3);
Besides, a ``man-in-the-middle'' attacker (\eg, a malicious advertising
company) could hijack and imperceptibly perturb the query image in order to
promote (QA+, row 4) or impede (QA-, row 5) the sales of specific products.

% 3. the differences between classification and ranking.

Unlike image classifiers where labels are predicted independently, a ranking
model determines the rank of a candidate according to the query as well as all
other candidates based on their relative distance.
Therefore, the existing adversarial attacks against DNN classifier are
incompatible with deep ranking models, and we need to thoroughly study the
\emph{adversarial ranking attack}.

% 4. two ranking attacks CA, QA, relationship, and a in-depth hint about implementation

In this paper, adversarial ranking attack aims to \emph{raise} or \emph{lower}
the ranks of some chosen candidates $C=\{c_1,c_2,\ldots,c_m\}$ with respect to
a specific query set $Q=\{q_1,q_2,\ldots,q_w\}$.
This can be achieved by either Candidate Attack (CA) or Query Attack (QA).
In particular, CA is defined as to raise (\emph{abbr.} CA+) or lower
(\emph{abbr.} CA-) the rank of a single candidate $c$ with respect to the query
set $Q$ by perturbing $c$ itself;
while QA is defined as to raise (\emph{abbr.} QA+) or lower (\emph{abbr.} QA-)
the ranks of a candidate set $C$ with respect to a single query $q$ by
perturbing $q$.
Thus, adversarial ranking attack can be achieved by performing CA on each $c\in
C$, or QA on each $q\in Q$.
In practice, the choice of CA or QA depends on the  accessibility to the
candidate or query respectively. Namely, CA is feasible for a modifiable
candidate, while QA is feasible for a modifiable query.

% 5. in-depth principle of attack

An effective implementation of these attacks is proposed in this paper.
As known, a typical deep ranking model maps samples (\ie, queries and
candidates) to a common embedding space, where the distances among them
determine the final ranking order.
Predictably, a sample's position in the embedding space will be changed by
adding a perturbation to it.
Therefore, the essential of adversarial ranking attack is to find a proper
perturbation, which could push the sample to a desired position that leads to
the expected ranking order.
Specifically, we first represent the expected ranking order as a set of
inequalities.
Subsequently, a triplet-like objective function is designed according to those
inequalities, and combined with Projected Gradient Descent (PGD) to efficiently
obtain the desired adversarial perturbation.

% 6. existing defense does not work when adapted to deep ranking with minor modification

Opposed to these attacks, \emph{adversarial ranking defense} is worth being
investigated especially for security-sensitive applications.
\textcolor{blue}{Until now, adversarial training~\cite{madry} remains to be one
of the most effective defense methods for classification.
However, the defense for deep ranking remains almost uncharted, and we
empirically discover the direct adaptation of adversarial training~\cite{madry}
suffers from model collapse~\cite{facenet}.
Thus, new defense for deep ranking need to be designed.
}

% 6. the defense method (that suppresses embedding shift)

To this end, an \emph{Embedding-Shifted Triplet} (EST) defense is proposed to
simultaneously defend against all attacks.
Note that shifting the embedding position of a sample is the key for any
type of ranking attack.
Although different attacks prefer distinct shift directions (\eg, CA+ and CA-
often prefer opposed shifting directions), a notable shift distance is usually
required.
If the shift distance of embeddings incurred by adversarial perturbation can be
reduced, all attacks are expected to be simultaneously defended.
Specifically, we first create adversarial examples with a maximized shift
distance off their original locations in the embedding space.
Then, the original training samples are directly replaced with their
corresponding adversarial examples during adversarial training.
\textcolor{blue}{Although this defense can moderately improve the model
robustness, it suffers from misleading gradient
and inefficient mini-batch exploitation, which collectively lead to slow
convergence and poor generalization.}

\textcolor{blue}{To address the problems identified from EST, and further
improve ranking model robustness, we propose another adversarial training
defense method named ``Anti-Collapse Triplet'' (ACT).
In particular, for each sample triplet (anchor, positive, negative), the
positive and negative samples are pulled close to each other via adversarial
attack,
while the model learns to separate them apart.
Thus, the ranking model is forced to learn robust representations~\cite{bugfeature} to better
differentiate different samples, lest the adversarial attack collapse them
together again.
This leads to a significant performance improvement over EST defense.}

\textcolor{blue}{In practice, a deep ranking model has zero prior-knowledge on
the type of adversarial attack it will confront with.
Thus, generic robustness against all types of known attacks is important for
a practical defense for real-world applications.
This also requires a defense not to be coupled with any specific attack.
In this paper, we also propose an Empirical Robustness Score (ERS) for deep
ranking models, which is absent from the literature.
It involves evaluating a model with a group of adversarial attacks that are
representative to all existing attacks against deep ranking, including but not
limited to the proposed CA and QA.
}

\textcolor{blue}{Experimental results on five datasets manifest that our
proposed CA and QA can greatly compromise a deep ranking model, and
successfully achieve the attack goals.
Besides, our proposed EST defense can moderately improve model
robustness, and the new ACT defense significantly outperforms the EST defense
both in terms of adversarial robustness against a wide range of attacks (hence
achieves a high ERS) and generalization performance on benign (\ie,
unperturbed) samples.
Thus, ranking models with our ACT defense is generically robust, as such model
is already resistant to a wide range of white-box attacks.}

% 7. contribution

To the best of our knowledge, this is the first work that thoroughly studies
the adversarial ranking attack, defense, and robustness evaluation.
In brief, our contributions are:

\begin{enumerate}[noitemsep,leftmargin=*]
\item The adversarial ranking attack is defined and implemented, which can
	intentionally change the ranking result and raise or lower the ranks of a
		set of selected candidates.
\item Two adversarial ranking defense methods are proposed to improve the
	ranking model robustness, and mitigate all the proposed attacks
		simultaneously.
\item \textcolor{blue}{A comprehensive empirical adversarial robustness evaluation metric for
	deep ranking models is proposed.}
\end{enumerate}

\textcolor{blue}{
This is an extension to the previous conference paper~\cite{advrank}, and the
new major contributions or changes include:
\begin{enumerate}[noitemsep,leftmargin=*]
\item An Anti-Collapse Triplet (ACT) defense, which achieves $60\%\sim540\%$
	robustness improvement on all datasets compared to the EST defense and
		generalizes better.
\item An Empirical Robustness Score (ERS) to comprehensively evaluate the
	robustness of deep ranking models.
To the best of our knowledge, this is the first work of such kind.
\item Experiment and discussion sections are re-written to better focus on
	defense and adversarial robustness.
	Our experimental settings are adjusted to be more compatible with
		existing related deep metric learning works.
\end{enumerate}
}

%%%%%%%%%%%%%%%%%%%%%%%%%%%%%%%%%%%%%%%%%%%%%%%%%%%%%%%%%%%%%%%%%%%%%%%%%%%%%

\section{Related Works}
\label{sec:2}

\textcolor{blue}{
\textbf{Deep Ranking} is generally formularized as the deep metric learning
(DML) problem~\cite{revisiting,dmlreality}, which is
important to a wide range of tasks such as
image retrieval~\cite{imagesim2},
cross-modal retrieval~\cite{hm-lstm,ladderloss},
and face/person recognition~\cite{facenet,9018082}.
Different from the traditional ``learning to rank''~\cite{LTR} methods,
deep ranking methods embed samples into a common space, and subsequently
determine the ranking order based on a defined distance metric. 
Recent works in DML mainly focus on loss functions, such
as triplet loss~\cite{facenet,imagesim2}, lifted structured loss~\cite{sop},
and margin loss~\cite{distance};
or data mining methods, such as semi-hard~\cite{facenet} and distance-weighted~\cite{distance} mining.
More details can be found in surveys~\cite{revisiting,dmlreality}.
}

\textbf{Adversarial Attack.}
Szegedy~\etal~\cite{l-bfgs} find DNN susceptible to imperceptible adversarial
due to its intriguing ``blind spot'' property.
Then Ian, \etal~\cite{fgsm} attribute this to the ``local linearity'' of neural
networks.
Following these, stronger white-box (\ie, model details are known to the adversary)
attacks~\cite{i-fgsm,cw,madry,apgd,lafeat,8807315}
are proposed to effectively compromise the state-of-the-art DNN classifiers.
Besides, adversarial examples are found
transferable~\cite{curlswhey,di-fgsm,ti-fgsm,ila}
among different models.
Image-agnostic universal perturbations are also
discovered~\cite{universal,unsupuap}.
Moreover, score-based and decision-based black-box attacks are proposed~\cite{benchmarking}
to overcome practical limitations such as inaccessibility of gradients.
It is even possible to create physical adversarial
examples~\cite{phy-synth,phy-robust,advpattern}.

\textbf{Adversarial Attack in Deep Ranking.}
For information retrieval systems, the risk of malicious manipulation of ranking
always exists~\cite{advrank-doc,advrank-rec}, and so does deep ranking.
Some existing attacks against deep ranking aim to incur mismatching top-ranked
items~\cite{universalret,flowertower,learn-to-misrank} as long as they mismatch with
the expected ones,
while the others lead to more specific ranking results~\cite{advrank,advorder,advpattern,qair}
beyond a mere mismatch.
These attacks will be reviewed in \secname~\ref{sec:5}.

\textbf{Adversarial Defense} consistently engages in arms race with adversarial
attacks~\cite{benchmarking,adaptive}.
Gradient masking-based defenses can be circumvented~\cite{obfuscated}.
Defensive distillation~\cite{distill2} has been compromised by C\&W~\cite{cw}.
Ensemble of weak defenses are insufficient~\cite{ensembleweak} against adversarial examples.
Other types of defenses may involve
input transformation~\cite{deflecting},
input reconstruction~\cite{magnet},
input replacement~\cite{nndef},
randomization~\cite{adv-bnn,self-ensemble},
feature denoising~\cite{featuredenoise},
%
%verification~\cite{reluplex,deepsafe},
%
%and evidential learning~\cite{edl},
%
but many of them are still susceptible to adaptive attack~\cite{adaptive}.
As an early defense~\cite{l-bfgs}, adversarial
training~\cite{fgsm,madry,advscale,bilateral}
remains very effective~\cite{benchmarking}.

\textcolor{blue}{
\textbf{Adversarial Defense in Deep Ranking} remains mostly uncharted.
In this paper, we identify problems from the defense~\cite{advrank} proposed
in the conference version of this paper
and attempt to eliminate the limitations.
Meanwhile, we propose an Empirical Robustness Score to evaluate the robustness
of deep ranking model.
}

%%%%%%%%%%%%%%%%%%%%%%%%%%%%%%%%%%%%%%%%%%%%%%%%%%%%%%%%%%%%%%%%%%%%%%%%%%%%%%%%

\section{Adversarial Ranking Attack}
\label{sec:3}

% 1. general ranking/embedding framework

Generally, a DNN-based ranking task could be formulated as a metric
learning problem~\cite{imagesimilarity}.
Given the query $q$ and candidate set
$X=\{c_1,c_2,\ldots,c_n\}$, deep ranking aims to learn a mapping $f$,
which is usually implemented by a DNN to map all candidates and query
into a common embedding space, such that the relative distances among
the embedding vectors could satisfy the expected ranking order.
For
instance, if candidate $c_i$ is more similar to the query $q$ than
candidate $c_j$, it is encouraged for the mapping $f$ to satisfy the
inequality $\|f(q)-f(c_i)\|<\|f(q)-f(c_j)\|$,
where $\|\cdot\|$ denotes $\ell_2$ norm (where embedding vectors
are projected onto the unit hypersphere following common practice~\cite{revisiting}).
For brevity, we denote the Euclidean distance $\|f(q)-f(c_i)\|$ as
$d(q,c_i)$.

% 2. principle of ranking attack, and transit to deeper detail

Therefore, adversarial ranking attack should find a proper adversarial
perturbation which changes the ranking order as expected.
For example, if a less relevant $c_j$ is expected to be ranked \emph{ahead}
of a relevant $c_i$, a proper perturbation $r$ should be found to perturb $c_j$, \ie,
$\tilde{c}_j=c_j+r$, such that the inequality $d(q,c_i)<d(q,c_j)$
could be changed into $d(q,c_i)>d(q,\tilde{c}_j)$.
In the following text, we will
describe Candidate Attack and Query Attack in detail.

\subsection{Candidate Attack}
\label{sec:31}

% 1. accurate definition.

Candidate Attack (\textbf{CA}) aims to raise (\emph{abbr.} \textbf{CA+}) or
lower (\emph{abbr.} \textbf{CA-}) the rank of a \emph{single} candidate $c$ with respect to
a set of queries $Q=\{q_1,q_2,\ldots,q_w\}$ by adding perturbation $r$
to the candidate itself, \ie, $\tilde{c}=c+r$.

% 2. intuitive and straightforward goal

% indeed CA perturbes only a single candidate. By independently perturbing
% every c\in C, the ranking of whole set of C will be raised w.r.t all q\in Q.

Let $\text{Rank}_X(q,c)$ denote the rank of the candidate $c$ with respect to
query $q$, where $X$ indicates the set of all candidates, and a smaller
rank value means a higher ranking.
Thus, the \textbf{CA+} that {\it raises} the rank of $c$ with respect to every query $q\in Q$
by perturbation $r$ can be formulated as follows,
\begin{align}
	r &= \argmin_{r\in\Gamma}\sum_{q\in Q}\text{Rank}_X(q,c+r), \label{eq:ca_intuitive} \\
	\Gamma &= \{r \big| \|r\|_\infty\leqslant\varepsilon; r,c+r\in [0,1]^N \}, \label{eq:gamma}
\end{align}
where $\Gamma$ is a $\ell_\infty$-bounded $\varepsilon$-neighbor of $c$,
$\varepsilon \in [0,1]$ is a predefined small positive constant,
the constraint $\|r\|_\infty \leqslant \varepsilon$ limits the perturbation $r$
to be ``visually imperceptible'', and $c+r\in [0,1]^N$ ensures the adversarial example
remains a valid input image.
Although alternative ``imperceptible'' constraints exist
(\eg, $\ell_0$~\cite{sparse}; $\ell_1$~\cite{ead}; 
and $\ell_2$~\cite{cw} variants),
we use the $\ell_\infty$ constraint following~\cite{fgsm,i-fgsm,madry}.

% 2. but it cannot be directly solved

However, the optimization problem Eq.~\eqref{eq:ca_intuitive}--\eqref{eq:gamma}
cannot be directly solved due to the discrete nature of the rank value $\text{Rank}_X(q,c)$.
Instead, a surrogate objective is needed.

% 3. we use the idea from metric learning

In deep metric learning, given two candidates $c_p, c_n \in X$ where $c_p$ is ranked
ahead of $c_n$, \ie, $\text{Rank}_X(q,c_p) < \text{Rank}_X(q,c_n)$, the ranking
order is represented as an inequality $d(q,c_p)<d(q,c_n)$ and formulated
in triplet loss:
\begin{equation}
	L_{\text{trip}}(q,c_p,c_n)=\left[\beta + d(q,c_p) - d(q,c_n)\right]_+,
	\label{eq:triplet}
\end{equation}
where $[\cdot]_+$ denotes $\max(0,\cdot)$, and $\beta$ is
a pre-defined margin constant.
% only give it a name after presenting the idea
This loss function is widely known as the triplet ranking loss~\cite{imagesimilarity,imagesim2,facenet}.

% 4. and convert our problem into solvable form

Similarly, the attacking goal of \textbf{CA+} in Eq.~\eqref{eq:ca_intuitive} can
be readily converted into a series of inequalities, and subsequently turned
into a sum of triplet losses,
\begin{equation}
	L_{\text{CA+}}(\tilde{c},Q;X)=\sum_{q\in Q}\sum_{x\in X}\big[d(q,\tilde{c})-d(q,x)\big]_{+}.
	\label{eq:ca+_loss}
	% \beta is intentionally dropped from our formulation because we don't need it.
\end{equation}
In this way, the original problem in Eq.~\eqref{eq:ca_intuitive}--\eqref{eq:gamma}
can be reformulated into a constrained optimization problem:
\begin{equation}
	r=\argmin_{r\in\Gamma}L_{\text{CA+}}(c+r,Q;X).
	\label{eq:ca_opt}
\end{equation}

% 3. we adapt PGD algorithm to our case

To solve the optimization problem, Projected
Gradient Descent (PGD) method~\cite{madry} (\emph{a.k.a}
the iterative version of FGSM~\cite{fgsm}) can be used.
Note that PGD is one of the most effective first-order
gradient-based algorithms~\cite{benchmarking}, popular among related works
about adversarial attack.

%% note: hinge loss only has subgradients due to the non-smooth zero/origin point.

Specifically, in order to find an adversarial perturbation $r$ to create a desired
adversarial candidate $\tilde{c}=c+r$,
the PGD algorithm alternates two steps at every iteration $t=1, 2,\ldots,\eta$.
Step one updates $\tilde{c}$ according to the gradient of Eq.~\eqref{eq:ca+_loss};
while step two clips the result of step one to fit in the $\varepsilon$-neighboring region $\Gamma$:
\begin{equation}
	\tilde{c}_{t+1}=\text{Clip}_{c,\Gamma}\big\{\tilde{c}_{t}
	-\alpha\text{sign}(\nabla_{\tilde{c}_t}L_{\text{CA+}}(\tilde{c}_t,Q,X))\big\}, \label{eq:PGDsteps}
\end{equation}
where $\alpha$ is a constant hyper-parameter indicating the PGD step size,
and $\tilde{c}_1$ is initialized as $c$.
After $\eta$ iterations, the desired adversarial candidate $\tilde{c}$ is obtained as $\tilde{c}_{\eta}$,
which is optimized to satisfy as many inequalities
as possible.
Each inequality represents a pairwise ranking sub-problem,
hence the adversarial candidate $\tilde{c}$ will be ranked ahead of
other candidates with respect to every specified query $q\in Q$.

% 4. to keep the narration as linear as possible, in the above text we only
%    discuss about CA+. here is CA-

Likewise, the \textbf{CA-} that {\it lowers} the rank of a candidate
$c$ with respect to a set of queries $Q$ can be obtained as:
\begin{equation}
L_{\text{CA-}}(\tilde{c},Q;X)=\sum_{q\in Q}\sum_{x\in X}\big[-d(q,\tilde{c})+d(q,x)\big]_{+}.
	\label{eq:ca-_loss}
\end{equation}

\subsection{Query Attack}
\label{sec:32}

% 1. intuitive objective, and the equivalent, solvable form

Query Attack (\textbf{QA}) aims to raise (\emph{abbr.} \textbf{QA+}) or lower (\emph{abbr.} \textbf{QA-})
the rank of a set of candidates
$C=\{c_1,c_2,\ldots,c_m\}$ with respect to an adversarially perturbed query $\tilde{q}=q+r$.
Thus, \textbf{QA} and \textbf{CA} are two ``symmetric'' attacks.
% "conjugate" lacks a definition, we'd better not mislead the reader to take it seriously. a quoted "symmetric" may be better to be understood
The \textbf{QA-} for {\it lowering} the rank could be formulated as follows:
\begin{equation}
	r=\argmax_{r\in\Gamma}\sum_{c\in C}\text{Rank}_X(q+r,c),
	\label{eq:qa_intuitive}
\end{equation}
where $\Gamma$ is the $\varepsilon$-neighbor of $q$.
Likewise, this objective can be transformed into a constrained optimization problem:
%
% Is there a typo here? Previously it is arg max, here it is arg min?? -- qilin
% No, I confirm it. we cannot maximize (argmax) a hinge loss that staying zero gradient area
% all functions passed to PGD will be minimized for consistency -- m.zhou
\begin{align}
	L_{\text{QA-}}(\tilde{q},C;X)&=\sum_{c\in C}\sum_{x\in X}\big[-d(\tilde{q},c)+d(\tilde{q},x)\big]_{+}, \label{eq:qa-_loss}\\
	r&=\argmin_{r\in\Gamma}L_{\text{QA-}}(q+r,C;X), \label{eq:qa_opt}
\end{align}
and it can be solved with the PGD algorithm. Similarly, the
\textbf{QA+} loss function $L_{\text{QA+}}$ for {\it raising} the rank of $c$
is as follows:
\begin{equation}
L_{\text{QA+}}(\tilde{q},C;X)=\sum_{c\in C}\sum_{x\in X}\big[d(\tilde{q},c)-d(\tilde{q},x)\big]_{+}.
\label{eq:qa+}
\end{equation}

%In contrast, CA does not suffer from this side-effect
%because it is impossible to disrupt the whole ranking list
%by modifying a single candidate.
Unlike \textbf{CA}, the \textbf{QA} perturbs the \emph{query} $q$,
and hence may drastically change its semantics, resulting in abnormal retrieval results.
For instance, after perturbing a ``lamp'' query image,
some totally unrelated candidates (\eg, ``shelf'', ``toaster'', \etc)
may appear in the top return list, which is undesired.
Thus, an ideal query attack should preserve the query
semantics, \ie, the candidates in the set $X{\setminus}C$ \footnote{The complement of the set $C$.}
should retain their original ranks if possible.
To this end, we propose the Semantics-Preserving Query Attack
(\textbf{SP-QA}) by adding an \textbf{SP} term to suppress the semantic
changes of the adversarial query $\tilde{q}$, \ie,
%
%\begin{small}
\begin{equation}
	L_{\text{SP-QA-}}(\tilde{q},C;X)=L_{\text{QA-}}(\tilde{q},C;X) + \xi L_{\text{QA+}}(\tilde{q},C_{\text{SP}};X), \label{eq:spqa}
\end{equation}%
%\end{small}%
%
where $C_{\text{SP}} = \left\{ c \in X{\setminus}C | \text{Rank}_{X
\setminus C}(q,c) \leqslant G \right\}$, \ie, $C_{\text{SP}}$ contains the
top-$G$ most-relevant candidates corresponding to $q$, and the
$L_{\text{QA+}}(q,C_{\text{SP}};X)$ term helps preserve the query semantics by
retaining some $C_{\text{SP}}$ candidates in the retrieved ranking list.
Constant $G$ is a predefined integer; and constant $\xi$
balances the attack effect and semantics preservation.

\textcolor{blue}{
	The SP term in Eq.~\eqref{eq:spqa} is expected to be negligible
	at the beginning of optimization.
	But the ranks of $C_\text{SP}$ are prone to be sacrificed later
	in order to optimize previous loss term.
	As drastic changes in the query semantics is strongly undesired, we set
	$\xi$ as a exponentially changing variable that does not involve in
	back-propagation, \ie,
	\begin{equation}
		\xi = \min\Big(10^9,~\exp\big(\zeta \times L_{\text{QA+}}(\tilde{q},C_{\text{SP}};X)\big) \Big).
		\label{eq:zeta}
	\end{equation}
	where $\zeta$ is a hyper-parameter, and the exponential is
	clipped to $10^9$ for numerical stability.
	Compared to a constant $\xi$ in the conference version, this
	change makes semantics preservation stronger.
}

% we don't have to mention SP-QA+, too easy to obtain
%\[
%L_{\text{SP-QA+}}(q,C;X)=\xi L_{\text{QA+}}(q,C_{\text{SP}};X)+L_{\text{QA+}}(q,C;X)
%\]

%%%%%%%%%%%%%%%%%%%%%%%%%%%%%%%%%%%%%%%%%%%%%%%%%%%%%%%%%%%%%%%%%%%%%%%%%%%%%%%

\section{Defense for Deep Ranking}
\label{sec:4}

% NOTE: the core reference for our defense is \cite{madry}.

\textcolor{blue}{
Adversarial training~\cite{advscale,madry} is a commonly used
defense for classification.
For instance, the Madry defense~\cite{madry} replaces or augments the original
training samples with their adversarial counterparts,
and is regarded as the one of the most effective~\cite{bilateral,obfuscated,benchmarking}
defense methods.
However, when directly adapting such defense to improve ranking
robustness, we empirically discover a primary challenge of
excessively hard (adversarial) training samples causing model
collapse~\cite{facenet} and failing to generalize.
Therefore, a new \emph{generic} defense for deep ranking is preferred.
}

\subsection{Defense with Embedding-Shifted Triplet}
\label{sec:41}
\label{sec:est}

Recall that the principle of an attack against deep ranking is to shift the
embedding of a sample to a proper position.
And a successful attack depends on a notable shift distance as well as a correct
shift direction.
Predictably, as a notable shift distance is indispensable for all types of
\textbf{CA} and \textbf{QA}, a reduction in the embedding shift distance that
adversarial perturbation could incur will lead to a more robust model against all of
them simultaneously.

\textcolor{blue}{
To this end, we propose an ``Embedding-Shifted Triplet'' (EST) defense to adversarially
train the ranking model with adversarial examples with the maximum embedding
shift, namely the distance off their original locations in the embedding space,
\eg, $r = \argmax_{r\in\Gamma} d(q+r, q) \label{eq:es}$
(resembles Feature Adversary~\cite{featureadversary} for classification).
Then we replace original training samples with such adversarial examples 
at each training iteration for adversarial training
following Madry, \etal~\cite{madry}.
Once a model is able to generalize on these adversarial examples,
the shift distance that adversarial perturbation can incur is expected to be implicitly reduced.
}

In brief, a model can be trained with EST as follows:
\begin{align}
	L_{\text{EST}}(q,c_p,c_n) = L_{\text{trip}} \Big( & q + \argmax_{r\in\Gamma}d(q+r,q),\nonumber \\
	& c_p + \argmax_{r\in\Gamma}d(c_p+r,c_p),\nonumber \\
	& c_n + \argmax_{r\in\Gamma}d(c_n+r,c_n) \Big),
	\label{eq:defense}
\end{align}
\textcolor{blue}{which only changes the loss function in the standard deep ranking
model training.
The idea can be adapted to other deep metric learning loss
functions as well.
And this defense method can converge without causing model collapse.
}

\subsubsection{Limitations of Embedding-Shifted Triplet}
\label{sec:411}

\textcolor{blue}{
Although EST can moderately improve ranking model robustness, we find it
slow to converge and poor in performance on unperturbed examples.
Further examination suggests that EST greatly suffers from misleading gradient
and inefficient mini-batch exploitation.
}

\textcolor{blue}{
We denote the embeddings of a sample triplet (\ie, an anchor,
a positive, and a negative sample) as $v_q = f(q)$, $v_p = f(c_p)$,
and $v_n = f(c_n)$, respectively.
When $L_\text{trip}(q, c_p, c_n) > 0$, the gradients of the triplet loss
with respect to these sample embeddings $v_q$, $v_p$, and $v_n$ are
respectively:
\begin{equation}
	\frac{\partial L_\text{trip}}{\partial v_q} =
	\frac{v_q-v_p}{\|v_q-v_p\|} - \frac{v_q-v_n}{\|v_q-v_n\|},
	\label{eq:ga}
\end{equation}
\begin{align}
	\frac{\partial L_\text{trip}}{\partial v_p} &=
	\frac{v_p-v_q}{\|v_q-v_p\|},
	\label{eq:gp}\\
	\frac{\partial L_\text{trip}}{\partial v_n} &=
	\frac{v_q-v_n}{\|v_q-v_n\|}.
	\label{eq:gn}
\end{align}
}

\begin{figure}[t]
	\includegraphics[width=1.0\linewidth]{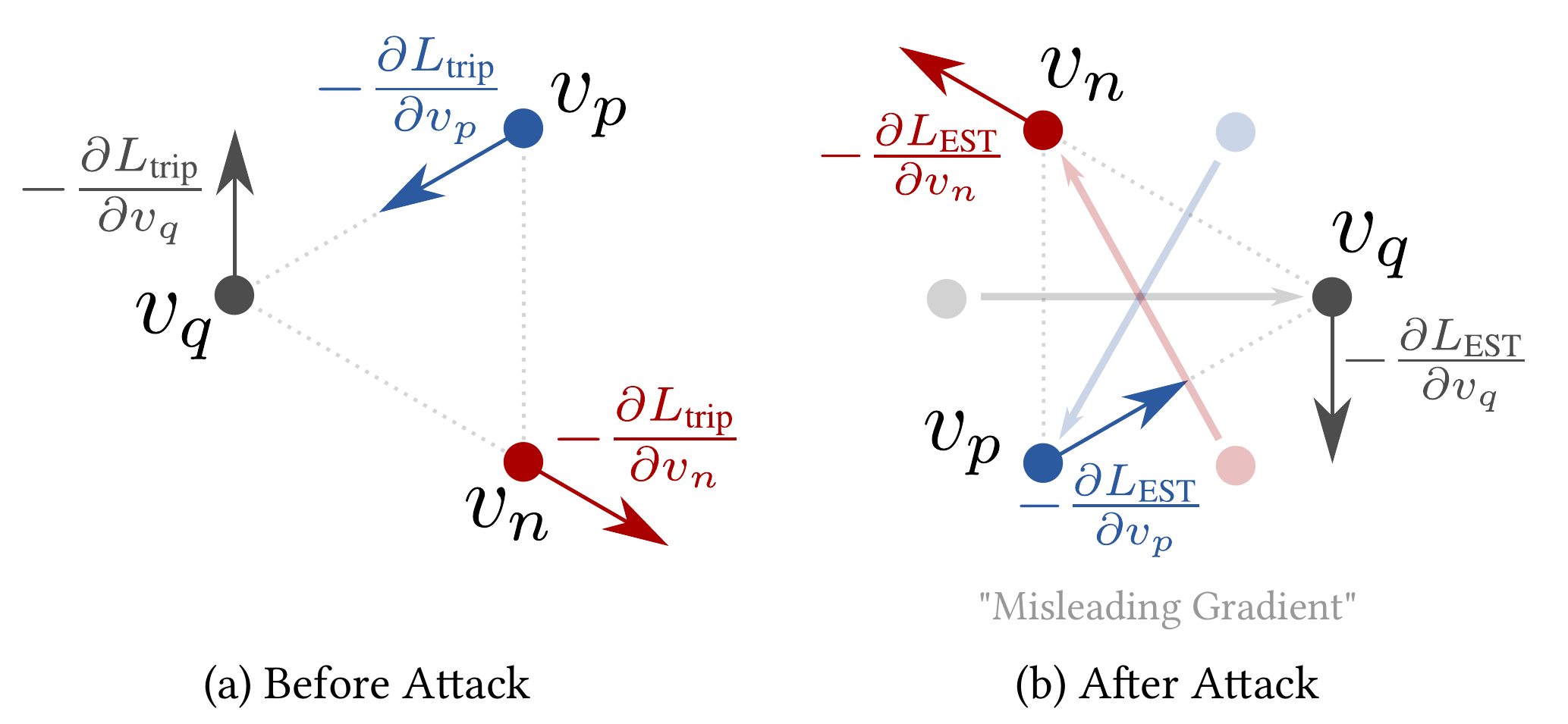}
	\caption{\textcolor{blue}{Misleading Gradient in EST defense~\cite{advrank}.
	With the samples moved far off their original locations in an arbitrary
	direction, the gradient of loss with respect to the embeddings may point
	to the wrong sample cluster.}}
	\label{fig:misgrad}
\end{figure}

\textcolor{blue}{
\textbf{Misleading Gradient.}
The embeddings $v_q$, $v_p$ and $v_n$ of adversarial examples
are moved off their original positions without any restriction
on the directions.
As a result, the embeddings may be near to the cluster of any other
sample class, as shown in \figurename~\ref{fig:misgrad}.
In this case, the gradients may point at wrong directions
(\eg, negative gradient of $v_n$ points towards the cluster of $v_q$ and $v_p$
in \figurename~\ref{fig:misgrad} (b)).
In other words, the gradient vectors can point at ``arbitrary''
directions as there is no restriction on the shifting directions of embedding
vectors, and such ``arbitrary'' directions can even vary across training
iterations.
As a result, the embeddings are moving towards ``arbitrary''
directions during the training process, leading to very slow
convergence rate and poor generalization on benign (\ie, unperturbed) examples.
}

\begin{figure}[t]
	\includegraphics[width=1.0\linewidth]{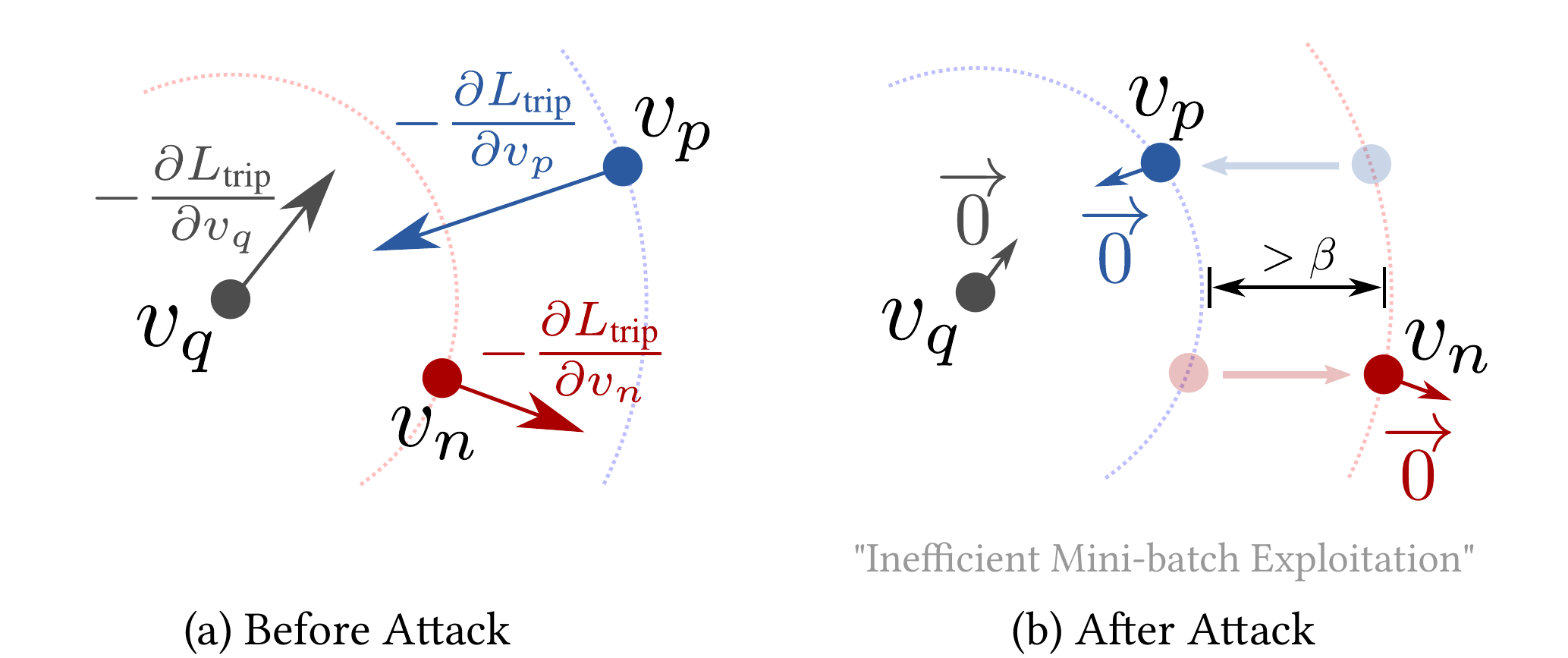}
	\caption{\textcolor{blue}{Inefficient Mini-batch Exploitation in EST defense~\cite{advrank}.
	With the samples moved far off their original locations in an arbitrary
	direction, the originally hard examples from which the model should learn
	will not be involved in the gradients of the loss function.
	}}
	\label{fig:ineff}
\end{figure}

\textcolor{blue}{
\textbf{Inefficient Mini-batch Exploitation.}
As maximizing the embedding shift distance is not an adversarially opposite goal
to minimizing triplet loss, the adversarial examples can lead to either
a larger or smaller loss value.
Namely, the originally easy samples with $L_\text{trip}=0$ can be turned into
hard samples (with a large loss).
Although sufficiently learned by the model, these samples will be moved
along ``arbitrary'' directions which is unnecessary.
Besides, the originally hard samples with a large $L_\text{trip}$ can be turned
into easy samples (with zero loss), as shown in \figurename~\ref{fig:ineff}.
In this way, the training samples from which it should learn will not be
involved into the gradients of the loss function.
In brief, the EST cannot help the model efficiently exploit the information
from mini-batches, and will converge slowly and generalize poorly due to
low-quality gradients.
}

\textcolor{blue}{
In short, we learn that an adversarial training-based defense for
deep ranking should be free from misleading gradients, and inefficient
exploitation of mini-batches.
Meanwhile, it should not create excessively hard adversarial examples to
trigger ranking model collapse~\cite{facenet}.
To seek for a better defense for deep ranking, how can these three conditions
be satisfied simultaneously?
}

\subsubsection{Mitigation of Limitations of EST}
\label{sec:412}

\textcolor{blue}{
To alleviate the problem of misleading gradient, we slightly modify the
EST defense into the ``Revised EST'' (REST) defense,
where the query sample $q$ is not replaced with its adversarial counterpart:
\begin{align}
	L_{\text{REST}}(q,c_p,c_n) = L_{\text{trip}} \Big( & q,\nonumber \\
	& c_p + \argmax_{r\in\Gamma}d(c_p+r,c_p),\nonumber \\
	& c_n + \argmax_{r\in\Gamma}d(c_n+r,c_n) \Big).
	\label{eq:rest}
\end{align}
According to Eq.~\eqref{eq:ga}--\eqref{eq:gn},
the gradients will be stabilized by the $v_q$ of benign example,
and will sometimes point at a proper direction (\eg, negative gradient of $v_n$
will not point towards the cluster of $v_q$ and $v_p$).
A positive effect is expected from such subtle and careful change.
}

\textcolor{blue}{
To improve the efficiency of mini-batch exploitation for EST, a simple and
straightforward mitigation is to increase the margin hyper-parameter $\beta$ in
the triplet loss.
In this way, more samples from which the model should learn will involve in
the gradient computation.
}

\textcolor{blue}{
In addition, a defense to directly ``Suppress the Embedding Shift''
(SES)\footnote{This is discussed in the supplementary material of Mo, \etal~\cite{advrank}.}
can circumvent both of limitations:
\begin{equation}
	L_\text{SES} = L_\text{trip}(q,c_p,c_n) + \sum_{x\in\{q,c_p,c_n\}}
	\max_{r\in\Gamma} d(x+r, x).
\end{equation}
All these mitigations will be examined in \secname~\ref{sec:7}.
}

\subsection{Defense with Anti-Collapse Triplet}
\label{sec:42}
\label{sec:act}

\textcolor{blue}{
Instead of mitigating limitations of EST, we attempt
to address them from the root.
In this paper, we present a new defense method that adversarially train
deep ranking models with ``Anti-Collapse Triplet'' (ACT), where the positive
and negative sample embeddings are ``collapsed'' (\ie, pulled close) together
through adversarial attack, and then the ranking model is trained to separate
them with the triplet loss. 
The ACT satisfies the three conditions in \secname~\ref{sec:411}.
}

\textcolor{blue}{
Specifically, given a triplet ($q$, $c_p$, $c_n$), we first find a pair
of adversarial positive and negative examples
$(\overrightarrow{c_p}, \overleftarrow{c_n}) = (c_p + \overrightarrow{r_p},
c_n + \overleftarrow{r_n})$, so that their embedding
vectors are ``collapsed'' together (the distance between them is minimized):
\begin{equation}
	(\overrightarrow{r_p}, \overleftarrow{r_n}) = \mathop{\arg\!\min}_{
		r_p\in\Gamma_{c_p}, r_n\in\Gamma_{c_n}}
	\|f(c_p+r_p) - f(c_n+r_n)\| .
	\label{eq:pnp}
\end{equation}
Subsequently, ($\overrightarrow{c_p}$, $\overleftarrow{c_n}$) and the original query
$q$ are fed into the triplet loss function as the ACT defense:
\begin{equation}
	L_\text{ACT}(q,c_p,c_n) = L_\text{trip}(q, 
	\overrightarrow{c_p}, \overleftarrow{c_n}).
	\label{eq:act}
\end{equation}
}

\begin{figure}[t]
	\includegraphics[width=1.0\linewidth]{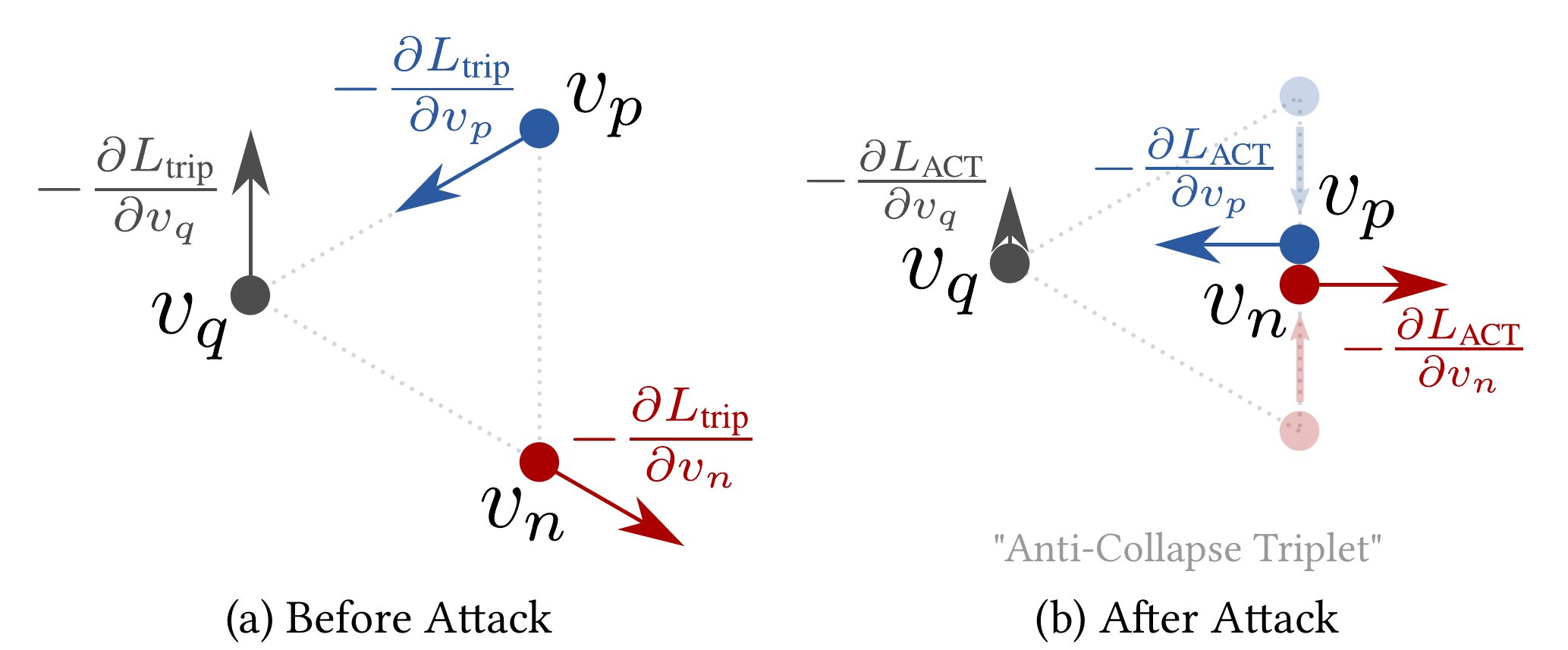}
	\caption{\textcolor{blue}{Gradient Direction of Anti-Collapse Triplet (ACT).
	ACT does not suffer from misleading gradient or inefficient mini-batch
	exploitation. It does not create excessively hard adversarial example
	neither.
	}}
	\label{fig:act}
\end{figure}

\textcolor{blue}{
During the training process, the model is forced to learn robust
representations~\cite{bugfeature} in order to differentiate and separate
the collapsed
positive and negative samples, lest them be collapsed again through
non-robust representations by the next round of adversarial attack.
Meanwhile, the robust feature will also help generalization on benign examples.
}

\textcolor{blue}{
Unlike EST, ACT will not suffer from ``misleading gradients''.
According to Eq.~\eqref{eq:gp}--\eqref{eq:gn}, the negative gradient for $v_p$
and $v_n$ will largely point at a proper direction (\eg, that
of $v_p$ points toward $v_q$ of unperturbed query, and that of $v_n$ points
as opposed to $v_q$) whether the adversarial attack successfully ``collapse''
$v_p$ and $v_n$ together or not, because the $v_p$ and $v_n$ are moved along
a deterministic direction.
When the ``collapse'' is successful, even if the directions of $v_p$ and $v_n$
slightly deflected due to the optimization algorithm (\ie, the projection step
of PGD), the norm of gradient for $v_q$ will be negligible
($\partial{L_\text{ACT}}/\partial v_q \approx 0$) regardless of the gradient
direction.
When ($\overrightarrow{c_p}$, $\overleftarrow{c_n}$) are merely slightly
pulled closer to each other, the model is already somewhat resistant to the
attack, and the gradient directions are still largely correct.
}

\textcolor{blue}{
The ACT does not suffer from inefficient mini-batch exploitation neither,
as shown in \figurename~\ref{fig:act}.
It will turn both easy and hard samples into moderately hard samples
(so that $L_\text{ACT}>0$), and will not create excessively hard examples
because the attack stops when $v_p=v_n$ (where $L_\text{ACT}=\beta$).
Thus, the gradient quality will not degrade, as the model will not omit
any sample from which it should learn.
Automatically stopping at $v_p=v_n$ ($L_\text{ACT}=\beta$) also means
that ACT will be unlikely to incur model collapse~\cite{facenet}.
}

\section{Adversarial Robustness Evaluation}
\label{sec:5}

\textcolor{blue}{
In real-world ranking applications, the ranking model has zero prior knowledge
on the exact type of attack it will confront with.
Thus, a practical defense should not couple with any specific attack,
and should be generically robust to a wide range of attacks~\cite{bugfeature}.
As a robustness evaluation metric for deep ranking is absent from the
literature, we propose an ``Empirical Robustness Score'' (ERS) to
comprehensively evaluate the empirical adversarial robustness.
}

\textcolor{blue}{
In particular, ERS adopts the following attacks:
\begin{enumerate}[noitemsep,leftmargin=*]
	\item \textbf{CA+ ($w{=}1$).}
		CA+ with $w>1$ will be more difficult, hence showing lower efficacy.
		Expectedly, a model resistant to CA+ ($w{=}1$) will be more resistant
		to CA+ ($w{>}1$).
		Thus, CA+ ($w{=}1$) is representative for the CA+ family.
	\item \textbf{CA- ($w{=}1$).}
		It is chosen for similar reasons.
	\item \textbf{QA+ ($m{=}1$).}
		The semantics-preserving term is discarded to make the attack
		much easier.
		Expectedly, a model resistant to QA+ will be more resistant to SP-QA+.
	\item \textbf{QA- ($m{=}1$).}
		It is chosen for similar reasons.
	\item \textbf{TMA,} namely the Targeted Mismatch Attack using global
		descriptor~\cite{flowertower}, which aims to increase the cosine
		similarity between $\tilde{q}$ and a randomly chosen ``target''
		$q_t$:
		\begin{equation}
			L_\text{TMA}(\tilde{q}, q_t) = 1 - f^{\sf T}(\tilde{q}) f(q_t).
			\label{eq:tma}
		\end{equation}
	\item \textbf{ES,} namely the Embedding Shift attack used in the EST
		defense, where an
		adversarial query $\tilde{q}=q+\argmax_{r\in\Gamma}d(q,q+r)$
		is fed into the model to incur a large embedding shift distance,
		and sometimes a mismatching top-$1$ retrieval as well.
	\item \textbf{LTM,} namely the Learning-To-Misrank~\cite{learn-to-misrank}
		attack which aims to perturb the ranking system output by minimizing
		the distance of unmatched pairs while maximizing the distance of matched
		pairs, as follows:
		\begin{equation}
			L_\text{LTM}(\tilde{q}) = [\max_{c_n\in X_n} d(\tilde{q},c_n)
			- \min_{c_p\in X_p} d(\tilde{q},c_p)]_+,
			\label{eq:ltm}
		\end{equation}
		where $X_n$ and $X_p$ are the sets containing all candidates of
		different class and the same class, respectively.
	\item \textbf{GTM,} a new ``Greedy Top-1 Misranking'' attack
		which aims to reduce the distance between adversarial query
		$\tilde{q}$ and the most confusing negative sample (\ie, the
		closest candidate to $q$ in a different class):
		\begin{equation}
			L_\text{GTM}(\tilde{q}) = d\big(\tilde{q},
			\argmin_{c_n\in X_n} d(q, c_n)
			\big).
			\label{eq:gtm}
		\end{equation}
		The efficacy of attack is measured with Recall@1 as well.
	\item \textbf{GTT,} a new ``Greedy Top-1 Translocation'' attack
		simplified from \cite{qair}, which aims to move the top-1
		candidate out of the top-ranked items with the following objective:
		\begin{equation}
			L_\text{GTT}(\tilde{q}) = L_\text{QA-}(\tilde{q},
			\{ \argmin_{c\in X} d(q, c)\}; X).
			\label{eq:gtt}
		\end{equation}
\end{enumerate}
Similar to the robustness of deep classifier~\cite{benchmarking},
that of deep ranking models can also be reflected by the reduction
in the efficacy of the above adversarial attacks.
Namely, a robust ranking model should prevent \textbf{CA+} and \textbf{QA+}
from moving selected candidates towards the topmost part of ranking list;
\textbf{CA-} and \textbf{QA-} from moving selected candidates towards the
bottommost part of ranking list;
\textbf{TMA} from achieving a high cosine similarity;
\textbf{ES} from incurring a large embedding shift distance or reducing
the recall performance;
\textbf{LTM} and \textbf{GTM} from reducing the recall performance;
\textbf{GTT} from moving the original top-$1$ candidate out from the top-$k$
results (extremely difficult with a small $k$).
The concrete quantitative evaluation protocol of these attacks 
will be detailed in \secname~\ref{sec:6}.
}

\textcolor{blue}{
After evaluating each attack against the model, the score of the corresponding
attack (\eg, success rate) will be normalized within $[0,100]$.
Finally the ERS is calculated as the average score across all attacks.
Thus, a high ERS is preferred for a robust ranking model that is resistant
to a wide range of attacks (even including unknown attacks).
}

\textcolor{blue}{Although there are related attacks focusing on
transferability~\cite{learn-to-misrank}, universal perturbation~\cite{universalret},
and even black-box attacks~\cite{qair},
ERS only involves simple white-box attacks representing or simplified from
them, but not any attack in a complicated form.
The reason is that finding a transferable, universal, or black-box perturbation
with an optionally complicated goal is much more difficult than finding a
per-model, per-sample, or white-box perturbation with a simplified goal, respectively.
Additionally, when a model is already resistant to the simpler attack,
a more complicated attack is more likely to fail.
Thus, none of a complicated attack is included in ERS, and the performance
of the simple white-box attacks are already representative enough for
adversarial robustness evaluation.
}

\textcolor{blue}{
In the end, we review all existing attacks against deep ranking
and discuss which attacks in ERS can represent them for robustness evaluation:
\begin{enumerate}[noitemsep,leftmargin=*]
	\item QAIR~\cite{qair} (CVPR'21) is a black-box attack to subvert
		the top-$k$ retrieval results, where the union between the
		original top-$k$ samples and the top-$k$ samples with adversarial 
		query is expected to be an empty set.
		We simplify QAIR into GTT to examine whether a model
		can retain the original top-$1$ sample within the top-$k$
		results.
	\item Bai, \etal~\cite{metric1} (T-PAMI'21) propose metric
		attacks, where the ``non-targeted'' attack is identical to ES,
		while the ``targeted'' attack can be represented by TMA or GTM.
	\item Learning-To-Misrank~\cite{learn-to-misrank} (CVPR'20)
		is directly adopted as a part of ERS evaluation in its simplest form.
		%
%	\item Bouniot, \etal~\cite{metric2} (CVPRw'20) present metric adversarial attacks,
%		where the ``Self Metric Attack'' is identical to ES, and the
%		``Furthest-Negative Attack'' is equivalent to LTM.
%		%
	\item DPQN~\cite{advdpqn} (AAAI'20) presents an attack whose goal formulation
		is identical to ES.
	\item AdvPattern~\cite{advpattern} (ICCV'19) introduces two concepts
		without their concrete white-box implementations.
		The ``Evading Attack'' to demote the rank of the selected candidate
		can be represented by CA-, QA-, ES, and LTM due to the similarity in
		the goal.
		According to their formulations, the ``Impersonation Attack'' to
		promote the rank of a selected candidate
		while demoting the rank of another selected candidate
		can be represented by CA+, QA+, TMA and GTM due to the similarity
		in the goal.
	\item Targeted Mismatch Attack~\cite{flowertower} (ICCV'19) is directly
		adopted in its simplest form (with global descriptor) as a part of
		our ERS evaluation, because it is
		``suitable when all parameters of the retrieval system are known''.
	\item Li, \etal~\cite{universalret} (ICCV'19) propose a universal perturbation
		to ``corrupt as much similarity relationships as possible in the data
		distribution'', which can be represented by ES. Besides, their formulation
		for ``corrupting pair-wise relationship'' is very similar to LTM.
%	\item PIRE~\cite{advrank-ut2} (ICMR'19) attack is identical to ES in formula.
%	\item Zhao, \etal~\cite{zhao2019unsupervised} (ArXiv) presents a GAN~\cite{GAN}-based
%		attack whose goal formulation is identical to ES.
%	\item ODFA~\cite{ofda} (ArXiv) presents an attack equivalent to ES.
\end{enumerate}
A model achieving a high ERS is expected to be robust against all attacks
mentioned above.
}

%%%%%%%%%%%%%%%%%%%%%%%%%%%%%%%%%%%%%%%%%%%%%%%%%%%%%%%%%%%%%%%%%%%%%%%%%%%%%

\section{Experiments}
\label{sec:6}
\label{sec:exp}

% BEGIN LYX
\begin{table*}[t]
\caption{\textcolor{blue}{
Adversarial Ranking Attack \& Defense with Ranking Models on Various Datasets.
For SP-QA+ and SP-QA-, we denote the worst rank percentile of $C_\text{SP}$
among four settings of $m$ as $\hat{R}_\text{GT}$ (the lower the better).
Mark ``$\uparrow$'' means larger values are preferred for a robust model,
while ``$\downarrow$'' means smaller values are preferred for a robust model.
The optional superscript or subscript to the arrows means the upper or lower
bound of the value, respectively.
	}}%
\label{tab:atkdef}
\vskip -1em
\resizebox{1.0\linewidth}{!}{%
\setlength{\tabcolsep}{0.25em}
\renewcommand{\arraystretch}{1.25} 
\begin{tabular}{cccc|cccc|c|rrrr|rrrr|rrrrr|rrrrr}

\toprule

\rowcolor{SteelBlue!33} & & & & \multicolumn{4}{c|}{\textbf{Benign Example}} & & \multicolumn{4}{c|}{\textbf{CA+ $\uparrow^{(50)}$}} & \multicolumn{4}{c|}{\textbf{CA- $\downarrow_{(0)}$}} & \multicolumn{5}{c|}{\textbf{SP-QA+ $\uparrow^{(50)}$}} & \multicolumn{5}{c}{\textbf{SP-QA- $\downarrow_{(0)}$}}\tabularnewline
\cline{5-8} \cline{6-8} \cline{7-8} \cline{8-8} \cline{10-27} \cline{11-27} \cline{12-27} \cline{13-27} \cline{14-27} \cline{15-27} \cline{16-27} \cline{17-27} \cline{18-27} \cline{19-27} \cline{20-27} \cline{21-27} \cline{22-27} \cline{23-27} \cline{24-27} \cline{25-27} \cline{26-27} \cline{27-27} 
\rowcolor{SteelBlue!33} \multirow{-2}{*}{\textbf{Dataset}}& \multirow{-2}{*}{\textbf{Model}} & \multirow{-2}{*}{\textbf{Loss}} & \multirow{-2}{*}{\textbf{Defense}} & \multicolumn{1}{c|}{R@1$\uparrow$} & \multicolumn{1}{c|}{R@2$\uparrow$} & \multicolumn{1}{c|}{mAP$\uparrow$} & NMI$\uparrow$ & \multirow{-2}{*}{$\varepsilon$} & \multicolumn{1}{r|}{$w{=}1$} & \multicolumn{1}{r|}{$2$} & \multicolumn{1}{r|}{$5$} & $10$ & \multicolumn{1}{r|}{$w{=}1$} & \multicolumn{1}{r|}{$2$} & \multicolumn{1}{r|}{$5$} & $10$ & \multicolumn{1}{r|}{$m{=}1$} & \multicolumn{1}{r|}{$2$} & \multicolumn{1}{r|}{$5$} & \multicolumn{1}{r|}{$10$} & $\hat{R}_{\text{GT}}$ & \multicolumn{1}{r|}{$m{=}1$} & \multicolumn{1}{r|}{$2$} & \multicolumn{1}{r|}{$5$} & \multicolumn{1}{r|}{$10$} & $\hat{R}_{\text{GT}}$\tabularnewline

\midrule

\multirow{6}{*}{MNIST} & \multirow{6}{*}{C2F2} & \multirow{6}{*}{Triplet} & \multirow{2}{*}{\xmark} & \multirow{2}{*}{\textbf{99.0}} & \multirow{2}{*}{\textbf{99.4}} & \multirow{2}{*}{\textbf{98.7}} & \multirow{2}{*}{84.7} & 8/255 & 41.8 & 43.7 & 45.1 & 45.6 & 4.9 & 4.6 & 4.5 & 4.5 & 44.8 & 46.5 & 47.9 & 48.8 & 0.3 & 1.7 & 1.4 & 1.2 & 1.1 & 0.3\tabularnewline
 &  &  &  &  &  &  &  & 77/255 & 3.3 & 10.3 & 14.1 & 15.9 & 69.9 & 69.6 & 69.2 & 69.1 & 29.2 & 35.7 & 41.6 & 44.8 & 0.7 & 2.7 & 2.2 & 1.9 & 1.8 & 0.7\tabularnewline
\cline{5-27} \cline{6-27} \cline{7-27} \cline{8-27} \cline{9-27} \cline{10-27} \cline{11-27} \cline{12-27} \cline{13-27} \cline{14-27} \cline{15-27} \cline{16-27} \cline{17-27} \cline{18-27} \cline{19-27} \cline{20-27} \cline{21-27} \cline{22-27} \cline{23-27} \cline{24-27} \cline{25-27} \cline{26-27} \cline{27-27} 
 &  &  & \multirow{2}{*}{$\checkmark$} & \multirow{2}{*}{98.3} & \multirow{2}{*}{99.0} & \multirow{2}{*}{91.3} & \multirow{2}{*}{80.7} & 8/255 & 41.1 & 41.9 & 42.2 & 42.3 & 3.0 & 2.7 & 2.5 & 2.5 & 46.2 & 47.4 & 48.7 & 49.0 & 0.2 & 1.5 & 1.3 & 1.2 & 1.2 & 0.4\tabularnewline
 &  &  &  &  &  &  &  & 77/255 & 6.8 & 11.5 & 14.6 & 15.8 & 36.0 & 33.9 & 32.2 & 31.7 & 32.6 & 37.9 & 43.2 & 45.6 & 0.7 & 3.6 & 2.9 & 2.4 & 2.3 & 0.8\tabularnewline
\cline{5-27} \cline{6-27} \cline{7-27} \cline{8-27} \cline{9-27} \cline{10-27} \cline{11-27} \cline{12-27} \cline{13-27} \cline{14-27} \cline{15-27} \cline{16-27} \cline{17-27} \cline{18-27} \cline{19-27} \cline{20-27} \cline{21-27} \cline{22-27} \cline{23-27} \cline{24-27} \cline{25-27} \cline{26-27} \cline{27-27} 
\rowcolor{red!10}\cellcolor{white} & \cellcolor{white} & \cellcolor{white} & & & & & & 8/255 & 48.2 & 48.3 & 48.9 & 49.2 & 2.7 & 2.7 & 2.6 & 2.6 & 48.8 & 48.6 & 49.4 & 49.6 & 0.0 & 0.7 & 0.6 & 0.6 & 0.6 & 0.0\tabularnewline
\rowcolor{red!10}\cellcolor{white} & \cellcolor{white} & \cellcolor{white} & \multirow{-2}{*}{$\bigstar$} & \multirow{-2}{*}{98.6} & \multirow{-2}{*}{99.1} & \multirow{-2}{*}{98.1} & \multirow{-2}{*}{\textbf{86.4}} & 77/255 & 33.4 & 37.2 & 41.3 & 43.9 & 7.6 & 7.5 & 7.5 & 7.5 & 36.7 & 39.8 & 44.2 & 46.1 & 0.4 & 2.7 & 2.2 & 1.9 & 1.8 & 0.6\tabularnewline

\hline 

\multirow{6}{*}{Fashion} & \multirow{6}{*}{C2F2} & \multirow{6}{*}{Triplet} & \multirow{2}{*}{\xmark} & \multirow{2}{*}{\textbf{87.6}} & \multirow{2}{*}{\textbf{92.7}} & \multirow{2}{*}{\textbf{84.9}} & \multirow{2}{*}{\textbf{77.8}} & 8/255 & 31.7 & 35.8 & 38.1 & 38.9 & 12.7 & 12.4 & 12.3 & 12.3 & 44.5 & 47.0 & 48.7 & 49.0 & 0.3 & 1.7 & 1.4 & 1.2 & 1.1 & 0.3\tabularnewline
 &  &  &  &  &  &  &  & 77/255 & 1.0 & 13.2 & 20.2 & 22.8 & 95.0 & 95.0 & 95.0 & 95.0 & 40.3 & 42.6 & 46.2 & 47.9 & 0.7 & 2.5 & 2.4 & 2.2 & 2.2 & 0.8\tabularnewline
\cline{5-27} \cline{6-27} \cline{7-27} \cline{8-27} \cline{9-27} \cline{10-27} \cline{11-27} \cline{12-27} \cline{13-27} \cline{14-27} \cline{15-27} \cline{16-27} \cline{17-27} \cline{18-27} \cline{19-27} \cline{20-27} \cline{21-27} \cline{22-27} \cline{23-27} \cline{24-27} \cline{25-27} \cline{26-27} \cline{27-27} 
 &  &  & \multirow{2}{*}{$\checkmark$} & \multirow{2}{*}{78.6} & \multirow{2}{*}{86.8} & \multirow{2}{*}{64.6} & \multirow{2}{*}{64.9} & 8/255 & 42.0 & 43.2 & 43.7 & 43.9 & 3.6 & 3.2 & 3.0 & 3.0 & 46.7 & 48.3 & 49.3 & 49.4 & 0.4 & 1.9 & 1.6 & 1.4 & 1.4 & 0.6\tabularnewline
 &  &  &  &  &  &  &  & 77/255 & 12.1 & 19.3 & 23.2 & 24.5 & 32.7 & 31.8 & 31.1 & 30.9 & 42.2 & 44.7 & 47.1 & 48.4 & 0.5 & 1.9 & 1.5 & 1.3 & 1.3 & 0.4\tabularnewline
\cline{5-27} \cline{6-27} \cline{7-27} \cline{8-27} \cline{9-27} \cline{10-27} \cline{11-27} \cline{12-27} \cline{13-27} \cline{14-27} \cline{15-27} \cline{16-27} \cline{17-27} \cline{18-27} \cline{19-27} \cline{20-27} \cline{21-27} \cline{22-27} \cline{23-27} \cline{24-27} \cline{25-27} \cline{26-27} \cline{27-27} 
\rowcolor{red!10}\cellcolor{white} & \cellcolor{white} & \cellcolor{white} & & & & & & 8/255 & 47.7 & 48.3 & 48.9 & 49.2 & 2.1 & 2.0 & 2.0 & 2.0 & 48.6 & 49.0 & 49.4 & 49.7 & 0.1 & 0.9 & 0.8 & 0.7 & 0.6 & 0.1\tabularnewline
\rowcolor{red!10}\cellcolor{white} & \cellcolor{white} & \cellcolor{white} & \multirow{-2}{*}{$\bigstar$} & \multirow{-2}{*}{79.4} & \multirow{-2}{*}{87.9} & \multirow{-2}{*}{71.6} & \multirow{-2}{*}{69.6} & 77/255 & 34.7 & 38.8 & 42.5 & 44.6 & 11.3 & 10.9 & 10.9 & 10.9 & 43.3 & 45.4 & 47.4 & 48.5 & 0.3 & 2.2 & 1.8 & 1.5 & 1.4 & 0.3\tabularnewline

\hline 

\multirow{6}{*}{CUB} & \multirow{6}{*}{RN18} & \multirow{6}{*}{Triplet} & \multirow{2}{*}{\xmark} & \multirow{2}{*}{\textbf{53.9}} & \multirow{2}{*}{\textbf{66.4}} & \multirow{2}{*}{\textbf{26.1}} & \multirow{2}{*}{\textbf{59.5}} & 2/255 & 0.2 & 5.5 & 13.5 & 17.8 & 99.6 & 99.3 & 98.9 & 98.7 & 13.4 & 23.3 & 32.2 & 39.1 & 0.4 & 17.2 & 11.5 & 7.0 & 6.8 & 0.6\tabularnewline
 &  &  &  &  &  &  &  & 8/255 & 0.0 & 5.0 & 13.2 & 17.6 & 100.0 & 100.0 & 100.0 & 100.0 & 23.5 & 29.2 & 37.1 & 42.1 & 0.8 & 11.9 & 10.8 & 6.7 & 6.0 & 0.8\tabularnewline
\cline{5-27} \cline{6-27} \cline{7-27} \cline{8-27} \cline{9-27} \cline{10-27} \cline{11-27} \cline{12-27} \cline{13-27} \cline{14-27} \cline{15-27} \cline{16-27} \cline{17-27} \cline{18-27} \cline{19-27} \cline{20-27} \cline{21-27} \cline{22-27} \cline{23-27} \cline{24-27} \cline{25-27} \cline{26-27} \cline{27-27} 
 &  &  & \multirow{2}{*}{$\checkmark$} & \multirow{2}{*}{8.5} & \multirow{2}{*}{13.0} & \multirow{2}{*}{2.6} & \multirow{2}{*}{25.2} & 2/255 & 8.8 & 13.3 & 18.8 & 21.4 & 79.2 & 77.2 & 75.4 & 74.7 & 11.4 & 16.5 & 26.2 & 32.6 & 0.4 & 40.4 & 30.4 & 19.0 & 13.7 & 0.5\tabularnewline
 &  &  &  &  &  &  &  & 8/255 & 2.7 & 8.4 & 14.8 & 18.0 & 97.9 & 97.6 & 97.2 & 97.1 & 13.6 & 17.6 & 28.1 & 33.7 & 0.5 & 37.9 & 28.9 & 16.6 & 10.7 & 0.7\tabularnewline
\cline{5-27} \cline{6-27} \cline{7-27} \cline{8-27} \cline{9-27} \cline{10-27} \cline{11-27} \cline{12-27} \cline{13-27} \cline{14-27} \cline{15-27} \cline{16-27} \cline{17-27} \cline{18-27} \cline{19-27} \cline{20-27} \cline{21-27} \cline{22-27} \cline{23-27} \cline{24-27} \cline{25-27} \cline{26-27} \cline{27-27} 
\rowcolor{red!10}\cellcolor{white} & \cellcolor{white} & \cellcolor{white} & & & & & & 2/255 & 39.0 & 42.1 & 44.2 & 45.1 & 5.4 & 4.8 & 4.5 & 4.5 & 41.2 & 44.1 & 46.1 & 47.8 & 0.1 & 1.9 & 1.4 & 1.1 & 0.9 & 0.1\tabularnewline
\rowcolor{red!10}\cellcolor{white} & \cellcolor{white} & \cellcolor{white} & \multirow{-2}{*}{$\bigstar$} & \multirow{-2}{*}{27.5} & \multirow{-2}{*}{38.2} & \multirow{-2}{*}{12.2} & \multirow{-2}{*}{43.0} & 8/255 & 15.5 & 21.9 & 28.8 & 32.5 & 37.7 & 34.1 & 31.8 & 31.2 & 25.9 & 32.9 & 39.9 & 43.4 & 0.1 & 7.1 & 4.0 & 2.7 & 2.1 & 0.2\tabularnewline

\hline 

\multirow{6}{*}{CARS} & \multirow{6}{*}{RN18} & \multirow{6}{*}{Triplet} & \multirow{2}{*}{\xmark} & \multirow{2}{*}{\textbf{62.5}} & \multirow{2}{*}{\textbf{74.0}} & \multirow{2}{*}{\textbf{23.8}} & \multirow{2}{*}{\textbf{57.0}} & 2/255 & 0.5 & 10.9 & 21.9 & 26.6 & 99.6 & 99.2 & 98.8 & 98.6 & 31.6 & 37.6 & 43.2 & 45.9 & 0.3 & 10.0 & 7.6 & 5.6 & 4.8 & 0.6\tabularnewline
 &  &  &  &  &  &  &  & 8/255 & 0.2 & 10.6 & 21.5 & 26.2 & 100.0 & 100.0 & 99.9 & 99.8 & 45.4 & 46.9 & 47.8 & 49.0 & 0.3 & 3.8 & 3.8 & 2.8 & 2.7 & 0.6\tabularnewline
\cline{5-27} \cline{6-27} \cline{7-27} \cline{8-27} \cline{9-27} \cline{10-27} \cline{11-27} \cline{12-27} \cline{13-27} \cline{14-27} \cline{15-27} \cline{16-27} \cline{17-27} \cline{18-27} \cline{19-27} \cline{20-27} \cline{21-27} \cline{22-27} \cline{23-27} \cline{24-27} \cline{25-27} \cline{26-27} \cline{27-27} 
 &  &  & \multirow{2}{*}{$\checkmark$} & \multirow{2}{*}{30.7} & \multirow{2}{*}{41.0} & \multirow{2}{*}{5.6} & \multirow{2}{*}{31.8} & 2/255 & 18.7 & 22.4 & 25.6 & 27.0 & 39.0 & 35.7 & 33.7 & 33.2 & 24.7 & 31.7 & 39.6 & 43.5 & 0.1 & 11.6 & 6.7 & 3.8 & 2.9 & 0.2\tabularnewline
 &  &  &  &  &  &  &  & 8/255 & 1.2 & 3.4 & 6.7 & 8.5 & 98.1 & 97.5 & 97.3 & 97.1 & 6.9 & 12.2 & 22.8 & 32.7 & 0.2 & 44.1 & 33.8 & 19.5 & 12.8 & 0.4\tabularnewline
\cline{5-27} \cline{6-27} \cline{7-27} \cline{8-27} \cline{9-27} \cline{10-27} \cline{11-27} \cline{12-27} \cline{13-27} \cline{14-27} \cline{15-27} \cline{16-27} \cline{17-27} \cline{18-27} \cline{19-27} \cline{20-27} \cline{21-27} \cline{22-27} \cline{23-27} \cline{24-27} \cline{25-27} \cline{26-27} \cline{27-27} 
\rowcolor{red!10}\cellcolor{white} & \cellcolor{white} & \cellcolor{white} & & & & & & 2/255 & 40.2 & 43.5 & 45.6 & 46.9 & 5.1 & 4.6 & 4.4 & 4.3 & 42.0 & 44.7 & 46.9 & 48.3 & 0.1 & 1.7 & 1.2 & 0.9 & 0.8 & 0.1\tabularnewline
\rowcolor{red!10}\cellcolor{white} & \cellcolor{white} & \cellcolor{white} & \multirow{-2}{*}{$\bigstar$} & \multirow{-2}{*}{43.4} & \multirow{-2}{*}{54.6} & \multirow{-2}{*}{11.8} & \multirow{-2}{*}{42.9} & 8/255 & 18.0 & 25.6 & 33.5 & 37.2 & 32.2 & 28.7 & 27.0 & 26.3 & 32.2 & 37.8 & 43.0 & 45.8 & 0.1 & 4.8 & 3.2 & 2.3 & 2.0 & 0.1\tabularnewline

\hline 

\multirow{6}{*}{SOP} & \multirow{6}{*}{RN18} & \multirow{6}{*}{Triplet} & \multirow{2}{*}{\xmark} & \multirow{2}{*}{\textbf{62.9}} & \multirow{2}{*}{\textbf{68.5}} & \multirow{2}{*}{\textbf{39.2}} & \multirow{2}{*}{\textbf{87.4}} & 2/255 & 5.0 & 10.6 & 18.4 & 22.7 & 55.7 & 48.3 & 43.1 & 41.2 & 12.5 & 20.2 & 30.1 & 36.6 & 0.1 & 18.9 & 12.8 & 7.6 & 6.3 & 0.3\tabularnewline
 &  &  &  &  &  &  &  & 8/255 & 0.1 & 3.2 & 10.1 & 15.0 & 99.3 & 98.7 & 98.1 & 97.9 & 19.9 & 27.6 & 35.4 & 39.6 & 0.2 & 19.0 & 14.0 & 11.0 & 10.9 & 1.1\tabularnewline
\cline{5-27} \cline{6-27} \cline{7-27} \cline{8-27} \cline{9-27} \cline{10-27} \cline{11-27} \cline{12-27} \cline{13-27} \cline{14-27} \cline{15-27} \cline{16-27} \cline{17-27} \cline{18-27} \cline{19-27} \cline{20-27} \cline{21-27} \cline{22-27} \cline{23-27} \cline{24-27} \cline{25-27} \cline{26-27} \cline{27-27} 
 &  &  & \multirow{2}{*}{$\checkmark$} & \multirow{2}{*}{46.0} & \multirow{2}{*}{51.4} & \multirow{2}{*}{24.5} & \multirow{2}{*}{84.7} & 2/255 & 39.8 & 43.0 & 45.4 & 46.4 & 4.6 & 3.7 & 3.4 & 3.3 & 37.6 & 41.1 & 44.6 & 46.7 & 0.0 & 1.9 & 1.4 & 1.0 & 0.9 & 0.0\tabularnewline
 &  &  &  &  &  &  &  & 8/255 & 12.5 & 18.5 & 25.2 & 28.7 & 43.6 & 36.4 & 32.0 & 30.7 & 12.1 & 19.7 & 30.2 & 36.8 & 0.1 & 22.2 & 13.4 & 8.7 & 7.1 & 0.2\tabularnewline
\cline{5-27} \cline{6-27} \cline{7-27} \cline{8-27} \cline{9-27} \cline{10-27} \cline{11-27} \cline{12-27} \cline{13-27} \cline{14-27} \cline{15-27} \cline{16-27} \cline{17-27} \cline{18-27} \cline{19-27} \cline{20-27} \cline{21-27} \cline{22-27} \cline{23-27} \cline{24-27} \cline{25-27} \cline{26-27} \cline{27-27} 
\rowcolor{red!10}\cellcolor{white} & \cellcolor{white} & \cellcolor{white} & & & & & & 2/255 & 45.0 & 47.9 & 49.4 & 49.9 & 1.7 & 1.5 & 1.4 & 1.3 & 42.4 & 44.7 & 46.8 & 48.0 & 0.0 & 1.0 & 0.8 & 0.7 & 0.6 & 0.0\tabularnewline
\rowcolor{red!10}\cellcolor{white} & \cellcolor{white} & \cellcolor{white} & \multirow{-2}{*}{$\bigstar$} & \multirow{-2}{*}{47.5} & \multirow{-2}{*}{52.6} & \multirow{-2}{*}{25.5} & \multirow{-2}{*}{84.9} & 8/255 & 24.1 & 29.9 & 36.9 & 40.1 & 10.5 & 7.8 & 6.3 & 5.9 & 22.8 & 28.8 & 36.9 & 41.3 & 0.0 & 8.2 & 5.3 & 3.6 & 3.0 & 0.1\tabularnewline

\bottomrule

\end{tabular}}
\end{table*}

% END LYX

% 1. overview: what we do in this section, and the measure

To validate the proposed attacks and defenses, and evaluate the ranking models
with ERS, we use five ranking datasets
including MNIST~\cite{mnist}, Fashion-MNIST (Fashion)~\cite{fashion},
CUB200-2011 (CUB)~\cite{cub200}, CARS196 (CARS)~\cite{cars196}, and Stanford Online Product (SOP)~\cite{sop}.
For MNIST and Fashion, we use the same dataset split as in \cite{advrank},
and the same dataset split as in \cite{revisiting} for CUB, CARS, and SOP.
Note, the dataset split for CUB and CARS is zero-shot~\cite{revisiting},
where the classes in training set do not overlap with those in test set.

\textcolor{blue}{
We conduct experiments with Nvidia RTX3090 GPUs and Intel Xeon 6226R CPU.
Our PyTorch~\cite{pytorch}-based code implementation of
the attacks, defenses, and the empirical robustness score
is available as a Python library at \texttt{https://cdluminate.github.io/robrank}.
}

\subsection{Evaluation Protocol}

\subsubsection{Baseline Deep Ranking Model}

\textcolor{blue}{
For MNIST and Fashion, we train a CNN model with $2$ convolution layers and
$2$ fully-connected layers (\emph{abbr.} C2F2), which is the same
as the model used in \cite{madry} except that the output dimension is changed to $D$.
}

\textcolor{blue}{
For CUB, CARS and SOP, we train a ResNet-18~\cite{resnet}
(\emph{abbr.} RN18)
model with the output dimension of the last fully-connected layer
changed as $D$ following \cite{revisiting}.
}

\textcolor{blue}{
We follow the standard deep ranking model training procedure.
All embeddings are projected onto the unit hypersphere~\cite{revisiting}.
In particular, models are fed with ``SPC-2'' mini-batches~\cite{revisiting},
where every mini-batch contains (at least) $2$ samples
for the each sampled class.
}

\textcolor{blue}{
The embedding space dimension is set as $D=512$ for all models.
The margin $\beta$ is set as $0.2$ by default following \cite{revisiting}.
The C2F2 model is trained with the Adam~\cite{adam} optimizer for $16$ epochs with
batch-size set as $128$, and a constant learning rate $1.0\times 10^{-3}$.
And the ResNet-18 model is trained with Adam optimizer for $150$ epochs
with batch-size $112$ following \cite{revisiting}, and a constant learning rate
$1.0\times 10^{-5}$.
The model performance on benign (\ie, unperturbed) examples is evaluated
in Recall@1 (R@1), Recall@2 (R@2), mAP and NMI following \cite{revisiting}.
All these metrics are scaled to $[0,100]$, where higher values are preferred.
}

\subsubsection{Evaluation of Candidate Attack \& Query Attack}
\label{sec:612}

We conduct attacks on the corresponding test dataset (used as $X$).
For each candidate $c$, its \emph{normalized} rank (\ie, ranking percentile)
is calculated as $R(q,c) = \frac{\text{Rank}_X(q,c)}{|X|}\times 100\%$
where $c\in X$, and $|X|$ is the length of full ranking list.
Thus, $R(q,c)\in [0,1]$, and a top ranked $c$ will have a small $R(q,c)$.
The attack effectiveness can be measured by the magnitude of change in
$R(q,c)$.
We omit the percent sign ``$\%$'' for brevity.

\textbf{Performance Metric of CA\&QA.}
To measure the performance of a single CA, we average the rank of
candidate $c$ across every query $q \in Q$, \ie, $R_\text{CA}(c)=\sum_{q\in Q}{R}(q,c)/w$.
Similarly, the performance of a single QA can be measured by the average
rank across every candidate $c \in C$, \ie, $R_\text{QA}(q)=\sum_{c\in C}{R}(q,c)/m$.
Then, the overall performance of an attack is reported as the mean of
$R_\text{CA}(c)$ or $R_\text{QA}(q)$ over $T$ times of independent trials,
accordingly.
%

% 2. sampling attacking targets
% Compared to randomly chosen queries, attacking with
% the top-$m$ samples as the target is believed to be simpler because the top-$m$
% targets often have high correlation with each other.

\textbf{Sampling for CA+ \& QA+.}
% CA+: X->random->(1 candidate as c), X->random->(w queries as Q)
% QA+: X->random->(1 query as q),     X->random->(m candidates as C)
For CA+, the query set $Q$ is randomly sampled from $X$.
Likewise, for QA+, the candidate set $C$ is randomly sampled from $X$.
Before attack, both the $R_\text{CA}(c)$ and $R_\text{QA}(q)$ will
approximate to $50\%$, and the attacks should significantly \emph{decrease}
the value towards $0\%$.

\textbf{Sampling for CA- \& QA-.}
% CA-: X->random->(1 candidate as c), X->random-within-top1%-closet-to(c)->(w queries as Q)
% QA-: X->random->(1 query as q),     X->random-within-top1%-close-to(q)->(m candidates as C)
%
We expect an attacker in practice prefer to lower some top ranked candidates
than further lowering candidates that are already away from the top part
of ranking list.
Thus, the $Q$ for CA- and the $C$ for QA- should be selected from the top
ranked samples (top-$1\%$ in our experiments) in $X$.
Formally, given the candidate $c$ for CA-,
we randomly sample the $w$ queries from
$\left\{ q \in X | R(c,q) \leqslant 1\% \right\}$ as $Q$.
Given the query $q$ for QA-,
$m$ candidates are randomly sampled from
$\left\{ c \in X | R(q,c) \leqslant 1\% \right\}$ as $C$.
Without attack, both the $R_\text{CA}(c)$ and $R_\text{QA}(q)$ will be close to $0\%$,
and the attacks should significantly \emph{increase} the value towards $100\%$.

% 3. attack / adv training: PGD parameter settings and misc

\textbf{Parameters for CA\&QA.}
We conduct CA with $w \in \{1,2,5,10\}$ queries, and QA with
$m \in \{1,2,5,10\}$ candidates, respectively. In SP-QA, we let $G=5$.
The parameter $\zeta$ in SP is empirically set to \verb|2e4| for MNIST;
\verb|4e4| for Fashion and CUB; and \verb|7e4| for CARS and SOP in order
to keep $C_\text{SP}$ within the top-$1\%$ ranked samples.
We perform $T=|X|$ times of attack to obtain the reported performance.

%During these attacks, the ranks of $C_\text{SP}$ are kept within $0.1$ despite
%of the extreme difficulty. Namely, the query semantics is unchanged.
%In practice we set e=??, which can simultaneously preserve the query semantic and effectively conduct attack.

\subsubsection{Hyper-Parameters for Projected Gradient Descent}

We use PGD for any attack mentioned in this paper.
Specifically, we investigate attacks of different
$\varepsilon \in \{8/255, 77/255\}$ on MNIST and Fashion, and
$\varepsilon \in \{2/255, 8/244\}$ on CUB, CARS and SOP following~\cite{advrank,madry,advscale}.
The number of PGD iterations is a constant $\eta=32$, where the size of each step is
$\alpha=\max(1/255,\text{round}(\varepsilon*255/25)/255)$.

% 4. defense

\subsubsection{Defense \& Empirical Robustness Score}
\label{sec:614}

\textcolor{blue}{
\textbf{Adversarial Defense.}
Following the procedure of Madry defense~\cite{madry}, we use a strong adversary
(\ie, $\varepsilon=77/255$ on MNIST and Fashion datasets; $\varepsilon=8/255$
on CUB200, CARS196, ans SOP)
to create adversarial examples for adversarial training based on standard
deep ranking~\cite{revisiting}.
}

\textcolor{blue}{
\textbf{Scores for ERS.}
Let $z$ be the performance score of any attack.
Specifically,
\textbf{(1)} we evaluate CA and QA as described in \secname~\ref{sec:612}.
The ranking percentile of CA- or QA- is normalized as $100-z$,
while the score of CA+ or QA+ is normalized as $2z$;
\textbf{(2)} The TMA is evaluated in cosine similarity.
It is normalized as $100(1-z)$ as its value lie within $[0,1]$ in most cases;
\textbf{(3)} The ES is evaluated in embedding shift distance (denoted as ES:D) and R@1 (denoted as ES:R).
The shift distance is normalized as $100*(1-z/2)$, while the R@1 is directly used;
\textbf{(4)} The LTM and GTM are both evaluated in R@1, which is directly used;
\textbf{(5)} The GTT is measured in the success rate (scaled to $[0,100]$ so directly
used for ERS) that
the top-$1$ sample is retained within the top-$k$ ($k=4$) result after attack.
We do not choose $k=1$ in order to lower the difficulty.
The score of any attack is averaged over $T=|X|$ times of trials.
Finally, the average of normalized scores across all attacks is ERS.
}

\begin{figure*}[t]
	\includegraphics[width=1.0\linewidth]{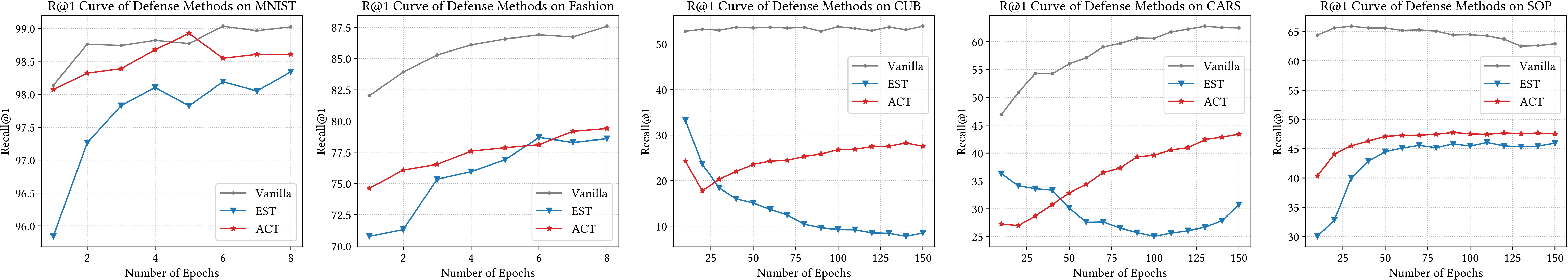}
	\caption{\textcolor{blue}{R@1 Curves of Defense Methods on Various Datasets.
	ACT consistently converges faster and generalizes better than EST
	on any dataset.
	}}
	\label{fig:expr1}
\end{figure*}

\begin{figure*}[t]
	\centering
	\includegraphics[width=0.8\linewidth]{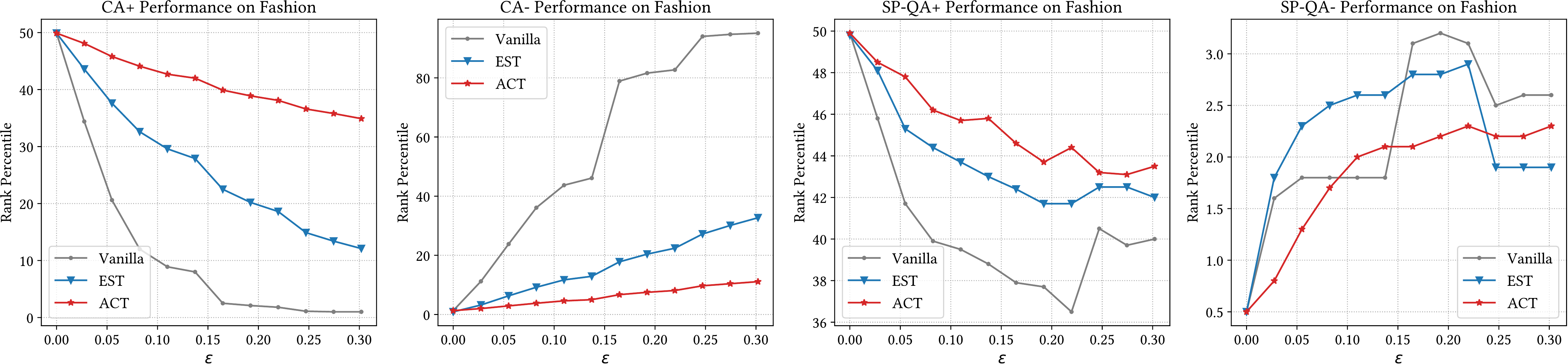}
\caption{\textcolor{blue}{Adversarial Ranking Attacks on Different Models with Varying $\varepsilon$.
	ACT consistently manifests better robustness than EST against any attack.
	}}
\label{fig:mn4atk}
\end{figure*}

\textcolor{blue}{
\textbf{Marks.}
To ease the comparison among different defenses in the following tables,
we mark the vanilla (\ie, without any defense) model as ``\xmark'',
EST as ``$\checkmark$'',
REST as ``$\checkmark^\ast$'',
SES as ``$\circ$'',
and ACT as ``$\bigstar$''.
Superscript ``$^\beta$'' to a mark indicates the usage of a non-default
margin $\beta$.}

\subsection{Candidate Attack \& Query Attack}
\label{sec:62}

% 1. mnist: experiment setup

\textbf{On MNIST Dataset.}
\textcolor{blue}{
We first train a vanilla (\ie, without defense) C2F2 ranking model.
Its retrieval performance can be found in the ``Benign Example'' columns of \tabname~\ref{tab:atkdef}.
Then we conduct adversarial ranking attacks against this model.}

% 2. mnist: ca+

\textcolor{blue}{
The attack results are presented in \tabname~\ref{tab:atkdef}.
For example, a strong \textbf{CA+} with $\varepsilon=77/255$ and $w=1$
can raise the rank $R_\text{CA}(c)$ from $50\%$ to $3.3\%$, which is close
to the top part of ranking list.
As expected, the attack has a weaker effect with
a $\varepsilon=8/255$ constraint, but remains effective.
Likewise, the rank of $c$ can be raised to $10.3\%$, $14.1\%$, $15.9\%$
for $w=2,5,10$ chosen queries, respectively.
This means that CA+ with more selected queries is more difficult.
We speculate such difficulty mainly stems from
geometric restriction\footnote{For instance, a candidate $c$ cannot be close
to $q_1$ and $q_2$ simultaneously when the distance between $q_1$ and $q_2$ is large.} in the embedding space
and optimization difficulty\footnote{The PGD optimizer is based on the first-order gradient.}.
}

% 3. mnist: ca-

\textcolor{blue}{
Meanwhile, a strong \textbf{CA-} for $w=1$
can lower the rank $R_\text{CA}(c)$ from $2.0\%$ to $69.9\%$.
The \textbf{CA-} for $w=2,5,10$
are similarly effective, and a larger $w$ makes the attack more difficult
due to the same reasons as \textbf{CA+}.
Note, the degradation of \textbf{CA-} performance with a large $w$
is relatively small because the selected queries $Q$
are highly correlated to each other due to our sampling method.
}

% 4. mnist: sp-qa

\textcolor{blue}{
The results of \textbf{SP-QA+} and \textbf{SP-QA-} are also shown in \tabname~\ref{tab:atkdef}.
For instance, the \textbf{SP-QA+} ($m=1$) can raise the rank of $c$ from $50\%$ to $29.2\%$,
while keeping the rank of $C_\text{SP}$ at $0.7\%$.
The \textbf{SP-QA-} ($m=1$) can lower the rank of $c$ from $0.5\%$ to $2.7\%$,
while retaining $C_\text{SP}$ in the top-ranked candidates.
This means \textbf{SP-QA} can raise or lower the rank of $C$
(with a less dramatic effect compared to \textbf{CA})
while largely retaining the query semantics.
The difficulty stems from \textbf{SP} term in Eq.~\eqref{eq:spqa},
and the \textbf{SP-QA} effectiveness is inversely correlated with $\zeta$.
Predictably, the effect can be boosted with a smaller $\zeta$,
at a cost of a larger change in query semantics.
Such trade-off for \textbf{SP-QA} between the extent of rank change
and the extent of semantics change depends on the attacker.}

% 1. other datasets: experiment setting

\textbf{On the Other Datasets.}
\textcolor{blue}{
We train the vanilla deep ranking models with C2F2 architecture on Fashion,
and with RN18 architecture on CUB, CARS, and SOP.
Their corresponding ranking performance on benign examples can be found in \tabname~\ref{tab:atkdef}.
Then we conduct attacks against these models.
}

% 2. ca+ and ca-

\textcolor{blue}{
The attack results are available in \tabname~\ref{tab:atkdef}.
As shown, the CA+ and CA- consistently achieve better effect
on datasets harder than MNIST.
For example, in a strong \textbf{CA+} ($w=1$) on SOP, the rank $R_\text{CA}(c)$ can be
raised to $0.1\%$, almost reaching the top.
In a strong \textbf{CA-} ($w=1$) on SOP, the rank of $c$
can be lowered to $99.3\%$, almost reaching the bottom.
We speculate that the dataset difficult partly contributes to the effectiveness
of CA, since it is difficult for a model to converge into an ideal state.
Moreover, the input dimension (\ie, $1\times 28\times 28$ for C2F2,
$3\times 224\times 224$ for RN18) is another reason why CA is more effective
on RN18 than C2F2.
According to Ian, \etal~\cite{fgsm}, it is easier to drive the
neural network output into a ``locally linear area'' 
with a higher input dimension.
Thus, CA can be easy to succeed on models with a high dimensional input.
And models like C2F2 are actually not easy to attack.
}

% BEGIN LYX
\begin{table*}[t]

\caption{\textcolor{blue}{Adversarial Robustness Evaluation for Deep Ranking Models on Various Datasets.
Mark ``$\uparrow$'' means larger values are preferred for a robust model,
while ``$\downarrow$'' means smaller values are preferred for a robust model.
The ERS is calculated as the average normalized score across all attacks involved.
}}
\label{tab:rob}
\vskip -1em
\resizebox{1.0\linewidth}{!}{%
\setlength{\tabcolsep}{0.3em}
\renewcommand{\arraystretch}{1.25}
\begin{tabular}{cccc|cccc|rrrrr|rrrrr||r}

\toprule

\rowcolor{SteelBlue!33} & & & & \multicolumn{4}{c|}{\textbf{Benign Example}} & \multicolumn{10}{c||}{\textbf{White-Box Attacks for Robustness Evaluation}} & \tabularnewline
\cline{5-18} \cline{6-18} \cline{7-18} \cline{8-18} \cline{9-18} \cline{10-18} \cline{11-18} \cline{12-18} \cline{13-18} \cline{14-18} \cline{15-18} \cline{16-18} \cline{17-18} \cline{18-18} 
\rowcolor{SteelBlue!33} \multirow{-2}{*}{\textbf{Dataset}} & \multirow{-2}{*}{\textbf{Model}} & \multirow{-2}{*}{\textbf{Loss}} & \multirow{-2}{*}{\textbf{Defense}} & R@1 $\uparrow$ & R@2 $\uparrow$ & mAP $\uparrow$ & NMI $\uparrow$ & CA+ $\uparrow$ & CA-\textbf{ $\downarrow$} & QA+ $\uparrow$ & QA-\textbf{ $\downarrow$} & TMA\textbf{ $\downarrow$} & ES:D\textbf{ $\downarrow$} & ES:R $\uparrow$ & LTM $\uparrow$ & GTM $\uparrow$ & GTT $\uparrow$ & \multirow{-2}{*}{\textbf{ERS} $\uparrow$}\tabularnewline

\midrule

 & & & \xmark & \textbf{99.0} & \textbf{99.4} & \textbf{98.7} & 84.7 & 3.3 & 69.9 & 3.7 & 83.8 & 0.940 & 1.314 & 0.6 & 21.7 & 10.5 & 0.0 & 13.3\tabularnewline
 &  &  & $\checkmark$ & 98.3 & 99.0 & 91.3 & 80.7 & 6.8 & 36.0 & 15.3 & 51.3 & 0.920 & 0.572 & 78.4 & 58.6 & 31.6 & 0.0 & 40.5\tabularnewline
\rowcolor{red!10}\cellcolor{white} \multirow{-3}{*}{MNIST} & \cellcolor{white} \multirow{-3}{*}{C2F2} & \cellcolor{white} \multirow{-3}{*}{Triplet} & $\bigstar$ & 98.6 & 99.1 & 98.1 & \textbf{86.4} & \textbf{33.4} & \textbf{7.6} & \textbf{35.7} & \textbf{3.8} & \textbf{0.145} & \textbf{0.259} & \textbf{93.2} & \textbf{96.6} & \textbf{96.1} & \textbf{1.1} & \textbf{78.6}\tabularnewline

\hline 

\multirow{3}{*}{Fashion} & \multirow{3}{*}{C2F2} & \multirow{3}{*}{Triplet} & \xmark & \textbf{87.6} & \textbf{92.7} & \textbf{84.9} & \textbf{77.8} & 1.0 & 95.0 & 0.5 & 94.2 & 0.993 & 1.531 & 0.1 & 0.8 & 6.7 & 0.0 & 4.5\tabularnewline
 &  &  & $\checkmark$ & 78.6 & 86.8 & 64.6 & 64.9 & 12.1 & 32.7 & 19.6 & 49.6 & 0.955 & \textbf{0.381} & 57.2 & 22.4 & 17.6 & 0.0 & 36.4\tabularnewline
\rowcolor{red!10}\cellcolor{white} & \cellcolor{white} & \cellcolor{white} & $\bigstar$ & 79.4 & 87.9 & 71.6 & 69.6 & \textbf{34.7} & \textbf{11.3} & \textbf{39.1} & \textbf{9.0} & \textbf{0.216} & 0.450 & \textbf{58.5} & \textbf{66.2} & \textbf{68.0} & \textbf{0.5} & \textbf{67.7}\tabularnewline

\hline 

\multirow{3}{*}{CUB} & \multirow{3}{*}{RN18} & \multirow{3}{*}{Triplet} & \xmark & \textbf{53.9} & \textbf{66.4} & \textbf{26.1} & \textbf{59.5} & 0.0 & 100.0 & 0.0 & 99.9 & 0.883 & 1.762 & 0.0 & 0.0 & 14.1 & 0.0 & 3.8\tabularnewline
 &  &  & $\checkmark$ & 8.5 & 13.0 & 2.6 & 25.2 & 2.7 & 97.9 & 0.4 & 97.3 & 0.848 & 1.576 & 1.4 & 0.0 & 4.0 & 0.0 & 5.3\tabularnewline
\rowcolor{red!10}\cellcolor{white} & \cellcolor{white} & \cellcolor{white} & $\bigstar$ & 27.5 & 38.2 & 12.2 & 43.0 & \textbf{15.5} & \textbf{37.7} & \textbf{15.1} & \textbf{32.2} & \textbf{0.472} & \textbf{0.821} & \textbf{11.1} & \textbf{9.4} & \textbf{14.9} & \textbf{1.0} & \textbf{33.9}\tabularnewline

\hline 

\multirow{3}{*}{CARS} & \multirow{3}{*}{RN18} & \multirow{3}{*}{Triplet} & \xmark & \textbf{62.5} & \textbf{74.0} & \textbf{23.8} & \textbf{57.0} & 0.2 & 100.0 & 0.1 & 99.6 & 0.874 & 1.816 & 0.0 & 0.0 & 13.4 & 0.0 & 3.6\tabularnewline
 &  &  & $\checkmark$ & 30.7 & 41.0 & 5.6 & 31.8 & 1.2 & 98.1 & 0.4 & 91.8 & 0.880 & 1.281 & 2.9 & 0.7 & 8.2 & 0.0 & 7.3\tabularnewline
\rowcolor{red!10}\cellcolor{white} & \cellcolor{white} & \cellcolor{white} & $\bigstar$ & 43.4 & 54.6 & 11.8 & 42.9 & \textbf{18.0} & \textbf{32.3} & \textbf{17.5} & \textbf{30.5} & \textbf{0.383} & \textbf{0.763} & \textbf{16.3} & \textbf{15.3} & \textbf{20.7} & \textbf{1.6} & \textbf{38.6}\tabularnewline

\hline 

\multirow{3}{*}{SOP} & \multirow{3}{*}{RN18} & \multirow{3}{*}{Triplet} & \xmark & \textbf{62.9} & \textbf{68.5} & \textbf{39.2} & \textbf{87.4} & 0.1 & 99.3 & 0.2 & 99.1 & 0.845 & 1.685 & 0.0 & 0.0 & 6.3 & 0.0 & 4.0\tabularnewline
 &  &  & $\checkmark$ & 46.0 & 51.4 & 24.5 & 84.7 & 12.5 & 43.6 & 10.6 & 34.8 & 0.468 & 0.830 & 9.6 & 7.2 & 17.3 & 3.8 & 31.7\tabularnewline
\rowcolor{red!10}\cellcolor{white} & \cellcolor{white} & \cellcolor{white} & $\bigstar$ & 47.5 & 52.6 & 25.5 & 84.9 & \textbf{24.1} & \textbf{10.5} & \textbf{22.7} & \textbf{9.4} & \textbf{0.253} & \textbf{0.532} & \textbf{21.2} & \textbf{21.6} & \textbf{27.8} & \textbf{15.3} & \textbf{50.8}\tabularnewline

\bottomrule

\end{tabular}}

\end{table*}

\begin{figure*}[t]
	\includegraphics[width=1.0\linewidth]{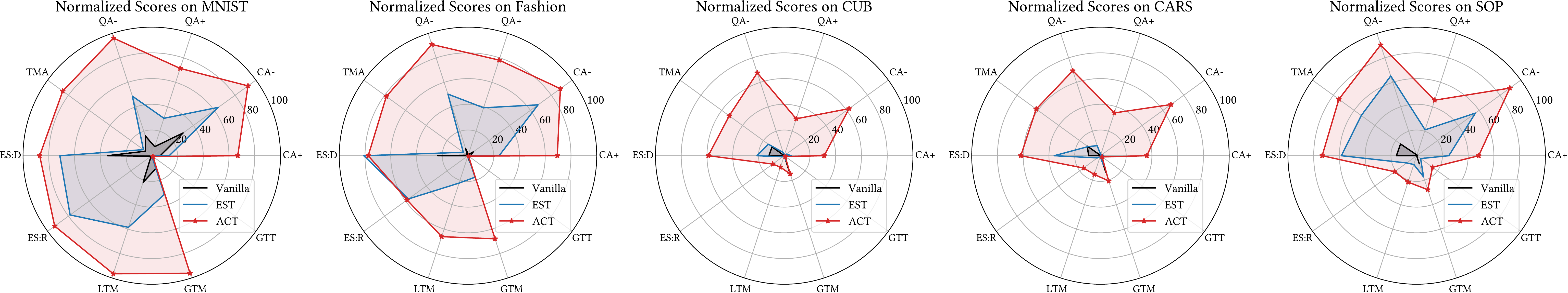}
	\caption{\textcolor{blue}{Comparison of Individual Normalized Attack Scores among Different
	Defense Methods.
	Every score is normalized within $[0,100]$.
	}}
	\label{fig:rob}
\end{figure*}
% END LYX

% 3. qa+ and qa-

\textcolor{blue}{
The SP-QA+ and SP-QA- are also effective on the rest datasets as expected.
Note, the SP-QA with a large $\varepsilon$ (\eg, $8/255$) is sometimes less
effective than that with a small $\varepsilon$ (\eg, $2/255$).
Because it is easier to significantly change the query semantics under a
large $\varepsilon$, meanwhile triggering a very strong semantics-preserving
penalty that immediately dominates the loss.
Thus, the optimizer will temporarily stop raising or lowering $C$ to
pull $C_\text{SP}$ back to the top of ranking list.
As a special characteristic, the SP-QA performance does not necessarily peak at a large $\varepsilon$.}

% summary for sec:62

\textcolor{blue}{
In summary, deep ranking models are vulnerable to adversarial ranking attack.
Our proposed CA and QA are particularly effective when the input
dimension is large, or when the corresponding dataset is difficult.
Previous study~\cite{advrank} also suggests that CA and QA are effective
regardless of the choice on metric learning loss function used for training,
or the distance metric in embedding space.
Therefore, we speculate that models used in realistic
applications are vulnerable, because they are
usually trained on larger-scale and more difficult datasets.
}

\subsection{Defending against CA \& QA}
\label{sec:63}

% 1. migration from attack to defense

\textcolor{blue}{
Deep ranking models are vulnerable to adversarial ranking attacks as suggested
by above experiments.
Whereas attacks may cause security or fairness concerns, a defense
to make a ranking model resistant to the attacks is necessary.
}

% 2. adversarial training: overall pattern

\textcolor{blue}{
To validate the proposed defenses,
we train ranking models on the five datasets with them, respectively.
The R@1 curves of these defensive models are presented in \figurename~\ref{fig:expr1}.
As a commonly expected phenomenon, adversarial training leads to notable
ranking performance drop on benign examples, and
this is particularly distinct on difficult datasets.
For example, while an RN18 without defense can achieve an R@1 of $62.9\%$
on SOP, an RN18 with EST only achieves R@1 of $46.0\%$ with EST, or only $47.5\%$ with ACT.
}

% 3. adversarial training: detailed difference

\textcolor{blue}{
As discussed in \secname~\ref{sec:4}, the EST defense~\cite{advrank} suffers
from misleading gradient and inefficient mini-batch
exploitation.
The corresponding model is expected to suffer from slow
convergence, and the R@1 curves in \figurename~\ref{fig:expr1} attest this.
Notably, EST leads to more pronounced performance drop under
the zero-shot setting (\ie, the CUB and CARS datasets).
From the figure, we also note that the proposed ACT defense, being free from
problems identified in EST, converges
much faster than EST in terms of R@1, and achieves a higher performance on all datasets.
The complete ranking performance of models with defense is presented in
\tabname~\ref{tab:atkdef}.
}

% 4. fashion-mnist figure

\textcolor{blue}{
Then we examine the defense methods with CA and SP-QA.
As shown in \figurename~\ref{fig:mn4atk}, we conduct attacks on the models with
$\varepsilon$ varying from $0$ to $77/255$ (with a $7/255$ interval) on
Fashion.
As an overall trend, the effect of an attack increases with the $\varepsilon$
increasing.
Through training with EST, the model gains a moderate robustness
against the attacks.
Besides, being free from the problems in EST, ACT outperforms EST by a large margin
in terms of resistance to these attacks.
}

% 5. defense on all datasets

\textcolor{blue}{
The complete results on all datasets can be found in \tabname~\ref{tab:atkdef}.
In general, the EST can achieve a moderate level of robustness
against CA and SP-QA, but at a high cost of ranking performance.
On the other hand, the newly proposed ACT can achieve significantly
higher robustness, while generalizes better on benign examples (especially
under zero-shot settings such as CUB and CARS).
ACT overwhelmingly outperforms the EST defense.
}

%Further analysis suggests that the weights in the first convolution layer of the defensive model are
%closer to $0$ and have smaller variance than those of the vanilla model,
%which may help resist adversarial perturbation from changing the layer
%outputs into the local linear area of ReLU~\cite{fgsm}.

% 6. misc findings

\textcolor{blue}{
Apart from these, the previous study~\cite{advrank} discover the
performance of attacks and defenses vary across different
embedding distance metrics (\eg, Euclidean distance or cosine distance),
or across different metric learning loss functions.
We leave further investigation for future study.
}

% BEGIN LYX
\begin{table*}
\caption{\textcolor{blue}{Robustness with Revised EST (REST) on Various Datasets.
	Its comparison with EST attests the efficacy of mitigation
	of misleading gradients.
	}}
\label{tab:rest}
\vskip -1em
\resizebox{1.0\linewidth}{!}{%
\setlength{\tabcolsep}{0.3em}
\renewcommand{\arraystretch}{1.25}
\begin{tabular}{cccc|rrrr|rrrrr|rrrrr||c}

\toprule

\rowcolor{SteelBlue!33} & & & & \multicolumn{4}{c|}{\textbf{Benign Example}} & \multicolumn{10}{c||}{\textbf{White-Box Attacks for Robustness Evaluation}} & \tabularnewline
\cline{5-18} \cline{6-18} \cline{7-18} \cline{8-18} \cline{9-18} \cline{10-18} \cline{11-18} \cline{12-18} \cline{13-18} \cline{14-18} \cline{15-18} \cline{16-18} \cline{17-18} \cline{18-18}
\rowcolor{SteelBlue!33} \multirow{-2}{*}{\textbf{Dataset}} & \multirow{-2}{*}{\textbf{Model}} & \multirow{-2}{*}{\textbf{Loss}} & \multirow{-2}{*}{\textbf{Defense}} & R@1 $\uparrow$ & R@2 $\uparrow$ & mAP $\uparrow$ & NMI $\uparrow$ & CA+ $\uparrow$ & CA-\textbf{ $\downarrow$} & QA+ $\uparrow$ & QA-\textbf{ $\downarrow$} & TMA\textbf{ $\downarrow$} & ES:D\textbf{ $\downarrow$} & ES:R $\uparrow$ & LTM $\uparrow$ & GTM $\uparrow$ & GTT $\uparrow$ & \multirow{-2}{*}{\textbf{ERS} $\uparrow$}\tabularnewline

\midrule 

\multirow{2}{*}{MNIST} & \multirow{2}{*}{C2F2} & \multirow{2}{*}{Triplet} & $\checkmark$ & 98.3 & 99.0 & 91.3 & 80.7 & 6.8 & 36.0 & 15.3 & 51.3 & 0.920 & 0.572 & 78.4 & 58.6 & 31.6 & 0.0 & 40.5\tabularnewline
 &  &  & $\checkmark^{\ast}$ & \textbf{98.8} & \textbf{99.3} & \textbf{98.3} & \textbf{87.5} & \textbf{26.9} & \textbf{9.5} & \textbf{32.9} & \textbf{6.4} & \textbf{0.391} & \textbf{0.291} & \textbf{88.4} & \textbf{90.8} & \textbf{89.4} & \textbf{0.1} & \textbf{71.9}\tabularnewline
\hline
\multirow{2}{*}{Fashion} & \multirow{2}{*}{C2F2} & \multirow{2}{*}{Triplet} & $\checkmark$ & 78.6 & 86.8 & 64.6 & 64.9 & 12.1 & 32.7 & 19.6 & 49.6 & 0.955 & 0.381 & \textbf{57.2} & 22.4 & 17.6 & 0.0 & 36.4\tabularnewline
 &  &  & $\checkmark^{\ast}$ & \textbf{79.8} & \textbf{87.7} & \textbf{71.1} & \textbf{69.4} & \textbf{28.3} & \textbf{17.0} & \textbf{36.8} & \textbf{13.6} & \textbf{0.701} & \textbf{0.360} & 51.1 & \textbf{49.4} & \textbf{57.2} & \textbf{0.1} & \textbf{56.9}\tabularnewline
\hline
\multirow{2}{*}{CUB} & \multirow{2}{*}{RN18} & \multirow{2}{*}{Triplet} & $\checkmark$ & 8.5 & 13.0 & 2.6 & 25.2 & 2.7 & 97.9 & 0.4 & 97.3 & 0.848 & 1.576 & 1.4 & 0.0 & 4.0 & 0.0 & 5.3\tabularnewline
 &  &  & $\checkmark^{\ast}$ & \textbf{13.6} & \textbf{20.8} & \textbf{5.9} & \textbf{34.2} & \textbf{9.4} & \textbf{40.2} & \textbf{11.2} & \textbf{40.5} & \textbf{0.810} & \textbf{0.590} & \textbf{7.0} & \textbf{1.1} & \textbf{7.7} & \textbf{0.1} & \textbf{26.6}\tabularnewline
\hline
\multirow{2}{*}{CARS} & \multirow{2}{*}{RN18} & \multirow{2}{*}{Triplet} & $\checkmark$ & 30.7 & 41.0 & 5.6 & 31.8 & 1.2 & 98.1 & 0.4 & 91.8 & 0.880 & 1.281 & 2.9 & 0.7 & 8.2 & 0.0 & 7.3\tabularnewline
 &  &  & $\checkmark^{\ast}$ & \textbf{31.9} & \textbf{42.3} & \textbf{7.0} & \textbf{34.3} & \textbf{7.6} & \textbf{45.1} & \textbf{7.8} & \textbf{44.6} & \textbf{0.735} & \textbf{0.642} & \textbf{7.1} & \textbf{5.4} & \textbf{12.3} & \textbf{0.4} & \textbf{26.1}\tabularnewline
\hline
\multirow{2}{*}{SOP} & \multirow{2}{*}{RN18} & \multirow{2}{*}{Triplet} & $\checkmark$ & 46.0 & 51.4 & 24.5 & 84.7 & 12.5 & 43.6 & 10.6 & 34.8 & 0.468 & 0.830 & 9.6 & 7.2 & 17.3 & 3.8 & 31.7\tabularnewline
 &  &  & $\checkmark^{\ast}$ & \textbf{47.2} & \textbf{52.3} & \textbf{25.3} & \textbf{84.9} & \textbf{22.0} & \textbf{13.6} & \textbf{20.4} & \textbf{11.3} & \textbf{0.362} & \textbf{0.500} & \textbf{18.3} & \textbf{20.5} & \textbf{24.7} & \textbf{10.1} & \textbf{47.2}\tabularnewline

\bottomrule
\end{tabular}}
\end{table*}

% END LYX
% BEGIN LYX
\begin{table*}
\caption{\textcolor{blue}{Robustness with EST in a Larger Margin on Fashion.
	The comparison attests the efficacy of mitigation of inefficient
	mini-batch exploitation in EST.}}
	\label{tab:estb}
\vskip -1em
\resizebox{1.0\linewidth}{!}{%
\setlength{\tabcolsep}{0.3em}
\renewcommand{\arraystretch}{1.25}
\begin{tabular}{cccc|rrrr|rrrrr|rrrrr||c}

	\toprule

\rowcolor{SteelBlue!33} & & & & \multicolumn{4}{c|}{\textbf{Benign Example}} & \multicolumn{10}{c||}{\textbf{White-Box Attacks for Robustness Evaluation}} & \tabularnewline
\cline{5-18} \cline{6-18} \cline{7-18} \cline{8-18} \cline{9-18} \cline{10-18} \cline{11-18} \cline{12-18} \cline{13-18} \cline{14-18} \cline{15-18} \cline{16-18} \cline{17-18} \cline{18-18}
\rowcolor{SteelBlue!33} \multirow{-2}{*}{\textbf{Margin $\beta$}} & \multirow{-2}{*}{\textbf{Model}} & \multirow{-2}{*}{\textbf{Loss}} & \multirow{-2}{*}{\textbf{Defense}} & R@1 $\uparrow$ & R@2 $\uparrow$ & mAP $\uparrow$ & NMI $\uparrow$ & CA+ $\uparrow$ & CA-\textbf{ $\downarrow$} & QA+ $\uparrow$ & QA-\textbf{ $\downarrow$} & TMA\textbf{ $\downarrow$} & ES:D\textbf{ $\downarrow$} & ES:R $\uparrow$ & LTM $\uparrow$ & GTM $\uparrow$ & GTT $\uparrow$ & \multirow{-2}{*}{\textbf{ERS} $\uparrow$}\tabularnewline

\midrule 

$0.2$ & C2F2 & Triplet & $\checkmark$ & 78.6 & 86.8 & \textbf{64.6} & \textbf{64.9} & 12.1 & 32.7 & 19.6 & 49.6 & 0.955 & \textbf{0.381} & 57.2 & 22.4 & 17.6 & 0.0 & 36.4\tabularnewline
\hline
$0.4$ & C2F2 & Triplet & $\checkmark^{\beta}$ & \textbf{81.0} & \textbf{88.6} & 64.0 & 63.5 & 17.5 & 36.8 & 28.1 & 40.7 & 0.764 & 0.662 & \textbf{63.1} & 35.9 & 22.8 & 0.0 & 42.6\tabularnewline
$0.6$ & C2F2 & Triplet & $\checkmark^{\beta}$ & 78.0 & 86.4 & 62.5 & 61.5 & 26.9 & 28.6 & 32.0 & 27.3 & 0.533 & 0.649 & 60.1 & 50.8 & 38.7 & 0.0 & 52.6\tabularnewline
$0.8$ & C2F2 & Triplet & $\checkmark^{\beta}$ & 77.1 & 85.6 & 60.7 & 61.7 & \textbf{35.7} & \textbf{18.5} & \textbf{40.0} & \textbf{14.4} & \textbf{0.232} & 0.434 & 57.9 & \textbf{53.5} & \textbf{53.9} & 0.0 & \textbf{63.9}\tabularnewline
$1.0$ & C2F2 & Triplet & $\checkmark^{\beta}$ & 73.9 & 84.1 & 57.2 & 60.3 & 28.3 & 26.5 & 34.5 & 24.3 & 0.422 & 0.565 & 45.2 & 45.3 & 37.9 & 0.0 & 53.3\tabularnewline

\bottomrule

\end{tabular}}
\end{table*}

% END LYX

% BEGIN LYX
\begin{table*}
\caption{\textcolor{blue}{Comparison among All Defenses and Their Variants. 
The best result in each column is highlighted in bold font,
and the second best is underscored.
}}
\label{tab:alldef}
\vskip -1em
\resizebox{1.0\linewidth}{!}{%
\setlength{\tabcolsep}{0.3em}
\renewcommand{\arraystretch}{1.25}
\begin{tabular}{cccc|rrrr|rrrrr|rrrrr||c}

	\toprule

\rowcolor{SteelBlue!33} & & & & \multicolumn{4}{c|}{\textbf{Benign Example}} & \multicolumn{10}{c||}{\textbf{White-Box Attacks for Robustness Evaluation}} & \tabularnewline
\cline{5-18} \cline{6-18} \cline{7-18} \cline{8-18} \cline{9-18} \cline{10-18} \cline{11-18} \cline{12-18} \cline{13-18} \cline{14-18} \cline{15-18} \cline{16-18} \cline{17-18} \cline{18-18}
\rowcolor{SteelBlue!33} \multirow{-2}{*}{\textbf{Dataset}} & \multirow{-2}{*}{\textbf{Model}} & \multirow{-2}{*}{\textbf{Loss}} & \multirow{-2}{*}{\textbf{Defense}} & R@1 $\uparrow$ & R@2 $\uparrow$ & mAP $\uparrow$ & NMI $\uparrow$ & CA+ $\uparrow$ & CA-\textbf{ $\downarrow$} & QA+ $\uparrow$ & QA-\textbf{ $\downarrow$} & TMA\textbf{ $\downarrow$} & ES:D\textbf{ $\downarrow$} & ES:R $\uparrow$ & LTM $\uparrow$ & GTM $\uparrow$ & GTT $\uparrow$ & \multirow{-2}{*}{\textbf{ERS} $\uparrow$}\tabularnewline

\midrule 

\multirow{6}{*}{Fashion} & \multirow{6}{*}{C2F2} & \multirow{6}{*}{Triplet} & $\checkmark$ & 78.6 & 86.8 & 64.6 & 64.9 & 12.1 & 32.7 & 19.6 & 49.6 & 0.955 & 0.381 & 57.2 & 22.4 & 17.6 & 0.0 & 36.4\tabularnewline
 &  &  & $\checkmark^{\beta}$ & 77.1 & 85.6 & 60.7 & 61.7 & \uline{35.7} & 18.5 & 40.0 & 14.4 & \uline{0.232} & 0.434 & \uline{57.9} & \uline{53.5} & 53.9 & 0.0 & \uline{63.9}\tabularnewline
 &  &  & $\checkmark^{\ast}$ & \textbf{79.8} & \uline{87.7} & 71.1 & 69.4 & 28.3 & 17.0 & 36.8 & 13.6 & 0.701 & \uline{0.360} & 51.1 & 49.4 & \textbf{57.2} & 0.1 & 56.9\tabularnewline
 &  &  & $\checkmark^{\ast,\beta}$ & \uline{79.6} & 87.6 & \textbf{72.8} & \textbf{70.8} & \uline{35.7} & \uline{16.6} & \uline{40.5} & 15.7 & 0.246 & 0.369 & 51.3 & 49.1 & 55.3 & 0.0 & 63.3\tabularnewline
 &  &  & $\circ$ & 71.2 & 83.2 & 56.6 & 59.5 & \textbf{38.0} & 17.8 & \textbf{44.7} & \uline{13.0} & 0.964 & \textbf{0.022} & 51.4 & 49.6 & 51.2 & \uline{0.2} & 59.0\tabularnewline
\rowcolor{red!10}\cellcolor{white} & \cellcolor{white} & \cellcolor{white} & $\bigstar$ & 79.4 & \textbf{87.9} & \uline{71.6} & \uline{69.6} & 34.7 & \textbf{11.3} & 39.1 & \textbf{9.0} & \textbf{0.216} & 0.450 & \textbf{58.5} & \textbf{66.2} & \textbf{68.0} & \textbf{0.5} & \textbf{67.7}\tabularnewline

\bottomrule
\end{tabular}}
\end{table*}

% END LYX

\subsection{Adversarial Robustness Evaluation}
\label{sec:64}

\textcolor{blue}{
As discussed in \secname~\ref{sec:5}, a practical defense for deep ranking
should be resistant to as more types of attacks as possible.
Thus, we evaluate the defenses with the proposed ERS.
}

\textcolor{blue}{
The performance of all attacks involved in ERS on the ranking models
is available in \tabname~\ref{tab:rob}. These scores are also normalized
and visualized in \figurename~\ref{fig:rob}.
Take the MNIST dataset as an example, as shown in \tabname~\ref{tab:rob}.
Compared to the EST defense, 
the ACT defense (1) effectively reduces the efficacy for CA+, CA-, QA+ and QA- to change
the rank of selected candidates; 
(2) only allows the cosine distance between two random samples to increase to $0.145$ for TMA;
(3) suppresses the maximum embedding shift distance to merely $0.259$ for ES:D;
(4) retains a much higher R@1 for ES:R, LTM, and GTM;
(5) retains the original top-$1$ candidate within the top-$4$ results at a success rate of $1.1\%$ in GTT.
Notably, ACT is the only method that consistently achieves
a non-zero performance in GTT which is extremely difficult.
}

\textcolor{blue}{
From the table and figure, the proposed ACT significantly outperforms
EST (state-of-the-art ranking defense) in defending against all attacks involved,
while achieving a higher generalization performance on benign examples.
By comparing the ERS scores, it is noted that ACT defense achieves at least $60\%$
and at most $540\%$ improvement over the EST defense.
This attests our analysis of EST in \secname~\ref{sec:4}.}

% i.e. Discussion
\section{Discussions}
\label{sec:7}
\label{sec:dis}

\textcolor{blue}{
In this section, we first conduct further experiments to verify the analysis
on the characteristics of EST.
Such analysis is also the foundation of the ACT defense.
In the end, we examine the relationship between adversarial robustness and
some commonly concerned factors in deep metric learning.
}

\textcolor{blue}{
Apart from these, as discussed in \secname~\ref{sec:3}, an \text{SP} loss term is
introduced in \textbf{SP-QA} to balance semantics preservation and the actual
attack goal.
By comparing the performance between \textbf{SP-QA} ($m=1$) in \tabname~\ref{tab:atkdef}
and that of pure \textbf{QA} ($m=1$) in \tabname~\ref{tab:rob}, it is clear that
\textbf{SP-QA} with a large $\zeta$ can be
much harder to achieve than pure \textbf{QA} (\ie, $\xi=0$).
}

% BEGIN LYX
\begin{table*}
\caption{\textcolor{blue}{Robustness with ACT in Different Anti-Collapse Strength (Different Margins in ACT).}}
\label{tab:actb}
\vskip -1em

\resizebox{1.0\linewidth}{!}{%
\setlength{\tabcolsep}{0.3em}
\renewcommand{\arraystretch}{1.25} 
\begin{tabular}{cccc|rrrr|rrrrr|rrrrr||c}

	\toprule

\rowcolor{SteelBlue!33} & & & & \multicolumn{4}{c|}{\textbf{Benign Example}} & \multicolumn{10}{c||}{\textbf{White-Box Attacks for Robustness Evaluation}} &\tabularnewline
\cline{5-18} \cline{6-18} \cline{7-18} \cline{8-18} \cline{9-18} \cline{10-18} \cline{11-18} \cline{12-18} \cline{13-18} \cline{14-18} \cline{15-18} \cline{16-18} \cline{17-18} \cline{18-18} 
\rowcolor{SteelBlue!33} \multirow{-2}{*}{\textbf{Margin $\beta$}} & \multirow{-2}{*}{\textbf{Model}} & \multirow{-2}{*}{\textbf{Loss}} & \multirow{-2}{*}{\textbf{Defense}} & R@1 $\uparrow$ & R@2 $\uparrow$ & mAP $\uparrow$ & NMI $\uparrow$ & CA+ $\uparrow$ & CA-\textbf{ $\downarrow$} & QA+ $\uparrow$ & QA-\textbf{ $\downarrow$} & TMA\textbf{ $\downarrow$} & ES:D\textbf{ $\downarrow$} & ES:R $\uparrow$ & LTM $\uparrow$ & GTM $\uparrow$ & GTT $\uparrow$ &  \multirow{-2}{*}{\textbf{ERS} $\uparrow$}\tabularnewline

\midrule

\rowcolor{red!10}\cellcolor{white}$0.2$ & \cellcolor{white}C2F2 & \cellcolor{white}Triplet & $\bigstar$ & \textbf{79.4} & \textbf{87.9} & \textbf{71.6} & 69.6 & 34.7 & \textbf{11.3} & 39.1 & \textbf{9.0} & 0.216 & 0.450 & 58.5 & \textbf{66.2} & \textbf{68.0} & \textbf{0.5} & 67.7\tabularnewline
\hline
$0.4$ & C2F2 & Triplet & $\bigstar^{\beta}$ & 78.5 & 87.0 & 69.6 & 68.2 & 37.3 & 11.9 & 40.7 & 9.2 & 0.172 & \textbf{0.399} & \textbf{59.0} & 63.0 & 66.8 & 0.4 & \textbf{68.7}\tabularnewline
$0.6$ & C2F2 & Triplet & $\bigstar^{\beta}$ & 78.0 & 86.4 & 68.3 & \textbf{69.7} & \textbf{38.6} & 11.8 & \textbf{42.3} & 9.4 & \textbf{0.169} & 0.409 & 52.7 & 63.2 & 66.1 & 0.4 & 68.6\tabularnewline
$0.8$ & C2F2 & Triplet & $\bigstar^{\beta}$ & 77.6 & 85.8 & 65.1 & 65.0 & 35.8 & 16.2 & 39.7 & 13.9 & 0.217 & 0.416 & 50.0 & 54.7 & 57.2 & 0.1 & 64.0\tabularnewline
$1.0$ & C2F2 & Triplet & $\bigstar^{\beta}$ & 77.2 & 86.1 & 62.9 & 64.3 & 35.3 & 18.5 & 39.6 & 15.6 & 0.205 & 0.443 & 42.5 & 50.9 & 46.6 & 0.1 & 61.3\tabularnewline

\bottomrule
\end{tabular}}
\end{table*}

% END LYX
% BEGIN LYX
\begin{table*}
\caption{\textcolor{blue}{Adversarial Robustness with Different Models or Triplet Sampling Strategies on CUB Dataset.
	}}
\label{tab:dmlfact}
\vskip -1em
\resizebox{1.0\linewidth}{!}{%
\setlength{\tabcolsep}{0.3em}
\renewcommand{\arraystretch}{1.25}
\begin{tabular}{cccc|rrrr|rrrrr|rrrrr||c}

	\toprule

\rowcolor{SteelBlue!33} & & & & \multicolumn{4}{c|}{\textbf{Benign Example}} & \multicolumn{10}{c||}{\textbf{White-Box Attacks for Robustness Evaluation}} & \tabularnewline
\cline{5-18} \cline{6-18} \cline{7-18} \cline{8-18} \cline{9-18} \cline{10-18} \cline{11-18} \cline{12-18} \cline{13-18} \cline{14-18} \cline{15-18} \cline{16-18} \cline{17-18} \cline{18-18}
\rowcolor{SteelBlue!33}\multirow{-2}{*}{\textbf{Group}} & \multirow{-2}{*}{\textbf{Model}} & \multirow{-2}{*}{\textbf{Loss}} & \multirow{-2}{*}{\textbf{Defense}} & R@1 $\uparrow$ & R@2 $\uparrow$ & mAP $\uparrow$ & NMI $\uparrow$ & CA+ $\uparrow$ & CA-\textbf{ $\downarrow$} & QA+ $\uparrow$ & QA-\textbf{ $\downarrow$} & TMA\textbf{ $\downarrow$} & ES:D\textbf{ $\downarrow$} & ES:R $\uparrow$ & LTM $\uparrow$ & GTM $\uparrow$ & GTT $\uparrow$ & \multirow{-2}{*}{\textbf{ERS} $\uparrow$}\tabularnewline

	\midrule

\multirow{3}{*}{\#0} & \multirow{3}{*}{RN18} & \multirow{3}{*}{Triplet} & \xmark & \textbf{53.9} & \textbf{66.4} & \textbf{26.1} & \textbf{59.5} & 0.0 & 100.0 & 0.0 & 99.9 & 0.883 & 1.762 & 0.0 & 0.0 & 14.1 & 0.0 & 3.8\tabularnewline
 &  &  & $\checkmark$ & 8.5 & 13.0 & 2.6 & 25.2 & 2.7 & 97.9 & 0.4 & 97.3 & 0.848 & 1.576 & 1.4 & 0.0 & 4.0 & 0.0 & 5.3\tabularnewline
\rowcolor{red!10}\cellcolor{white} & \cellcolor{white} & \cellcolor{white} & $\bigstar$ & 27.5 & 38.2 & 12.2 & 43.0 & \textbf{15.5} & \textbf{37.7} & \textbf{15.1} & \textbf{32.2} & \textbf{0.472} & \textbf{0.821} & \textbf{11.1} & \textbf{9.4} & \textbf{14.9} & \textbf{1.0} & \textbf{33.9}\tabularnewline
\hline
\multirow{9}{*}{\#1} & \multirow{3}{*}{RN50} & \multirow{9}{*}{Triplet} & \xmark & \textbf{53.3} & \textbf{65.6} & \textbf{26.4} & \textbf{58.9} & 0.3 & 100.0 & 0.1 & 99.4 & 0.909 & 1.766 & 0.0 & 0.0 & 14.5 & 0.0 & 3.7\tabularnewline
 &  &  & $\checkmark$ & 18.7 & 26.7 & 5.6 & 33.4 & 3.8 & 82.2 & 4.0 & 79.6 & 0.859 & 1.113 & 4.7 & 0.3 & 7.0 & 0.0 & 12.4\tabularnewline
 &  &  & $\bigstar$ & 31.7 & 40.8 & 14.3 & 45.7 & \textbf{16.9} & \textbf{35.0} & \textbf{16.0} & \textbf{31.3} & \textbf{0.410} & \textbf{0.842} & \textbf{14.9} & \textbf{11.0} & \textbf{18.1} & \textbf{1.2} & \textbf{36.2}\tabularnewline
\cline{2-2} \cline{4-19} \cline{5-19} \cline{6-19} \cline{7-19} \cline{8-19} \cline{9-19} \cline{10-19} \cline{11-19} \cline{12-19} \cline{13-19} \cline{14-19} \cline{15-19} \cline{16-19} \cline{17-19} \cline{18-19} \cline{19-19}
 & \multirow{3}{*}{IBN} &  & \xmark & \textbf{53.5} & \textbf{66.1} & \textbf{27.4} & \textbf{59.1} & 0.1 & 100.0 & 0.1 & 99.6 & 0.876 & 1.752 & 0.1 & 0.0 & 13.7 & 0.0 & 3.9\tabularnewline
 &  &  & $\checkmark$ & 24.0 & 33.7 & 8.3 & 38.4 & 3.2 & 80.6 & 5.4 & 64.8 & 0.950 & \textbf{0.555} & 9.3 & 0.7 & 9.3 & 0.0 & 16.8\tabularnewline
 &  &  & $\bigstar$ & 28.9 & 38.7 & 12.4 & 43.3 & \textbf{17.4} & \textbf{36.7} & \textbf{17.2} & \textbf{29.2} & \textbf{0.445} & 0.790 & \textbf{14.6} & \textbf{12.3} & \textbf{15.8} & \textbf{1.7} & \textbf{36.4}\tabularnewline
\cline{2-2} \cline{4-19} \cline{5-19} \cline{6-19} \cline{7-19} \cline{8-19} \cline{9-19} \cline{10-19} \cline{11-19} \cline{12-19} \cline{13-19} \cline{14-19} \cline{15-19} \cline{16-19} \cline{17-19} \cline{18-19} \cline{19-19}
 & \multirow{3}{*}{Mnas} &  & \xmark & \textbf{54.2} & \textbf{66.2} & \textbf{26.5} & \textbf{59.9} & 0.1 & 100.0 & 0.1 & 99.6 & 0.860 & 1.785 & 0.0 & 0.0 & \textbf{13.6} & 0.0 & 3.9\tabularnewline
 &  &  & $\checkmark$ & 17.4 & 25.0 & 4.4 & 30.2 & 0.6 & 99.0 & 0.4 & 98.0 & 0.828 & 1.598 & 1.2 & 0.0 & 6.3 & 0.0 & 5.0\tabularnewline
 &  &  & $\bigstar$ & 23.3 & 32.1 & 10.0 & 40.9 & \textbf{17.4} & \textbf{32.3} & \textbf{18.3} & \textbf{27.2} & \textbf{0.482} & \textbf{0.736} & \textbf{13.6} & \textbf{9.0} & 12.8 & \textbf{0.9} & \textbf{36.3}\tabularnewline
\hline
\multirow{9}{*}{\#2} & \multirow{9}{*}{RN18} & \multirow{3}{*}{\vtop{\hbox{\strut ~~~~Triplet}\hbox{\strut (Semihard)}}} & \xmark & \textbf{39.7} & \textbf{51.3} & \textbf{19.1} & \textbf{51.8} & 0.2 & 100.0 & 0.2 & 99.8 & 0.820 & 1.681 & 0.4 & 0.0 & 12.1 & 0.0 & 4.8\tabularnewline
 &  &  & $\checkmark$ & 4.5 & 7.7 & 1.7 & 22.9 & 5.6 & 99.9 & 5.9 & 83.6 & 0.907 & 1.821 & 1.2 & 0.4 & 2.1 & 0.0 & 6.1\tabularnewline
 &  &  & $\bigstar$ & 20.3 & 28.8 & 9.1 & 39.3 & \textbf{21.0} & \textbf{27.4} & \textbf{22.3} & \textbf{21.2} & \textbf{0.436} & \textbf{0.694} & \textbf{11.0} & \textbf{9.0} & \textbf{12.9} & \textbf{1.6} & \textbf{39.4}\tabularnewline
\cline{3-19} \cline{4-19} \cline{5-19} \cline{6-19} \cline{7-19} \cline{8-19} \cline{9-19} \cline{10-19} \cline{11-19} \cline{12-19} \cline{13-19} \cline{14-19} \cline{15-19} \cline{16-19} \cline{17-19} \cline{18-19} \cline{19-19}
 &  & \multirow{3}{*}{\vtop{\hbox{\strut ~~~Triplet}\hbox{\strut (Softhard)}}} & \xmark & \textbf{53.4} & \textbf{65.3} & \textbf{25.9} & \textbf{60.1} & 0.0 & 100.0 & 0.0 & 99.5 & 0.932 & 1.353 & 0.1 & 0.0 & 12.5 & 0.0 & 5.2\tabularnewline
 &  &  & $\checkmark$ & 37.6 & 48.7 & 14.6 & 46.4 & 1.1 & 92.2 & 1.3 & 80.5 & 0.984 & \textbf{0.376} & 11.9 & 3.1 & 12.0 & 0.0 & 14.2\tabularnewline
 &  &  & $\bigstar$ & 39.4 & 50.2 & 18.6 & 51.3 & \textbf{6.8} & \textbf{61.5} & \textbf{5.2} & \textbf{60.4} & \textbf{0.506} & 1.032 & \textbf{12.8} & \textbf{11.3} & \textbf{17.7} & \textbf{0.3} & \textbf{24.2}\tabularnewline
\cline{3-19} \cline{4-19} \cline{5-19} \cline{6-19} \cline{7-19} \cline{8-19} \cline{9-19} \cline{10-19} \cline{11-19} \cline{12-19} \cline{13-19} \cline{14-19} \cline{15-19} \cline{16-19} \cline{17-19} \cline{18-19} \cline{19-19}
 &  & \multirow{3}{*}{\vtop{\hbox{\strut ~~~Triplet}\hbox{\strut (Distance)}}} & \xmark & \textbf{48.2} & \textbf{60.7} & \textbf{23.4} & \textbf{56.6} & 0.0 & 100.0 & 0.0 & 100.0 & 0.830 & 1.747 & 0.1 & 0.0 & \textbf{13.5} & 0.0 & 4.3\tabularnewline
 &  &  & $\checkmark$ & 6.3 & 10.8 & 2.1 & 30.2 & 7.3 & 99.2 & 8.1 & \textbf{29.0} & 1.000 & \textbf{0.013} & \textbf{2.6} & 0.2 & 1.5 & 0.0 & \textbf{20.6}\tabularnewline
 &  &  & $\bigstar$ & 18.5 & 25.4 & 4.9 & 29.1 & \textbf{13.5} & \textbf{95.3} & \textbf{8.6} & 97.5 & \textbf{0.517} & 1.698 & 1.6 & \textbf{0.3} & 9.4 & 0.0 & 12.6\tabularnewline
 
	\bottomrule

\end{tabular}}
\end{table*}

% END LYX

\subsection{Characteristics of EST \& ACT}
\label{sec:71}

\textcolor{blue}{
In \secname~\ref{sec:41}, we present the mitigations to the
misleading gradient and inefficient mini-batch exploitation for EST,
correspondingly.
The mitigation of the former issue is the Revised EST (REST),
and the mitigation of the later one is to enlarge the margin hyper-parameter $\beta$.
To validate the underlying ideas, we train defensed ranking models with
these mitigations, and then evaluate them with ERS.
}

\textbf{Mitigation of Misleading Gradient.}
\textcolor{blue}{
We train ranking models on all datasets with REST and compare the result
with those of EST, as shown in \tabname~\ref{tab:rest}.
Clearly, the sole mitigation of misleading gradient leads to significant
improvement in both ranking performance on benign example and the resistance
to all attacks. A significantly higher ERS is achieved on every dataset,
with at least $49\%$ and at most $402\%$ improvement over EST, nearly
reaching that of ACT.
}

\textbf{Mitigation of Inefficient Mini-batch Exploitation.}
\textcolor{blue}{
We enlarge the margin $\beta$ and train models on the Fashion dataset, and evaluate the
resulting models with the proposed ERS.
The results can be found in \tabname~\ref{tab:estb}.
Clearly, this mitigation can also lead to significant improvement in robustness
against all types of attacks, but results in a drop in benign example ranking
performance.
The robustness peaks at $\beta=0.8$, but the ranking performance does not peak
at the meantime.
}

\textbf{Mitigations Combined.}
\textcolor{blue}{
We combine the two mitigations together, and compare the performance
with other defense methods in \tabname~\ref{tab:alldef}.
As shown in the $4$-th row of the table, the combination achieves a higher
ranking performance, as well as a significant improvement on robustness,
but is still outperformed by ACT.
This is because ACT eliminates these problems
instead of merely mitigating them.}

\textbf{SES Defense.}
\textcolor{blue}{The SES discussed in \secname~\ref{sec:41} can lead
to competitive robustness, but meanwhile a significant ranking performance drop.
Notably, SES is particularly resistant to the ES attack (embedding
shift is suppressed to $0.022$), but is relatively weak against some other attacks.
Hence, suppressing the embedding
shift that adversarial perturbation could incur is not the only condition
to achieve a robust model.
We speculate that solely reducing the ES may not necessarily introduce robustness,
because the embedding shift can also be reduced
by ``shrinking'' the embedding space.}

\textbf{Anti-Collapse Strength of ACT.}
\textcolor{blue}{
The margin parameter $\beta$ has an extra meaning in our ACT defense, \ie,
the ``strength'' to separate the adversarially ``collapsed'' positive
and negative samples.
To better understand the margin for ACT,
we conduct experiments with various margins,
as shown in \tabname~\ref{tab:actb}.
From the table, we discover that a slightly larger margin (\eg, $0.4$) can
further boost the model robustness, as the model is forced to learn more
robust representations.
However, an excessively large margin will harm the generalization performance
as expected in the literature of deep ranking.
Thus, there is a trade-off between generalization performance and robustness
as they do not peak at the same time.
}

\textcolor{blue}{
In summary, all these experiments attest our analysis on the EST in
\secname~\ref{sec:4}. Being free from problems identified in EST, our 
ACT greatly outperforms EST in various aspects.
}

\subsection{Robustness with Other Models / Triplet Samplers}
\label{sec:72}

\textcolor{blue}{
In deep metric learning, the model and the triplet sampling strategy
may greatly impact the performance, but their impact on adversarial
robustness remains unexplored.
To this end, we follow \cite{revisiting} and train models on CUB with
different architectures including ResNet-50 (RN50)~\cite{resnet},
Inception-BN (IBN)~\cite{inceptionv2}, and MnasNet-1.0 (Mnas)~\cite{mnasnet});
or with different triplet sampling strategies including semi-hard~\cite{facenet},
soft-hard~\cite{revisiting}, and distance-weighted~\cite{distance} triplet sampling.
Then we evaluate the resulting models with ERS, as shown in \tabname~\ref{tab:dmlfact}.
Note, with an aligned number of training epochs across all models in a
specific architecture, the models without defense suffer from overfitting,
but the models with defense need to be sufficiently trained in order to gain
enough robustness as well as performance on benign examples.
Results in the table suggests that ACT outperforms EST in virtually any setting.
}
 
\textcolor{blue}{
By comparing group \#0 and group \#1, 
we discover that model capacity alone benefits adversarial
robustness for adversarial training.
For example, compared to RN18 which achieves an ERS of $5.3$ for EST or $33.9$
for ACT, RN50 achieves an ERS of $12.4$ for EST or $36.2$ for ACT.
A similar effect is also shown by IBN, which has a larger model capacity
than RN18.
This observation is consistent with Madry's conclusion~\cite{madry}.
On the other hand, although the Mnas model has a lower model capacity than RN18,
it achieves a comparable robustness due to its better architecture.}

\textcolor{blue}{
By comparing group \#0 (uniform sampling) and group \#2, we note that 
triplet sampling greatly impacts adversarial robustness.
Compared to uniform sampling, the negative samples from semi-hard
sampler ($-\beta<d(q,c_p)-d(q,c_n)<0$)
are easier to be collapsed with a positive sample, so the model
will learn more robust representations to separate them (hence a higher ERS);
The positive and negative samples from soft-hard sampler are initially further
from each other, hence are less likely to be collapsed together (hence a lower ERS);
The distance-weighted sampler has a higher chance to select very far negative
samples that are even harder to be collapsed with the positive sample (hence a even lower ERS).
Namely, the success rate for the attack to ``collapse'' the positive and negative
samples will affect the model's focus on learning robust representations to
separate them apart.
}

\subsection{Defense with FGSM instead of PGD}

\begin{table*}
\caption{\textcolor{blue}{Defense Methods using Adversarial Examples Created Using FGSM.}}
\label{tab:fgsm}
\vskip -1em
\resizebox{1.0\linewidth}{!}{%
\setlength{\tabcolsep}{0.3em}
\renewcommand{\arraystretch}{1.25} 
\begin{tabular}{cccc|rrrr|rrrrr|rrrrr||c}

\toprule

\rowcolor{SteelBlue!33} & & & & \multicolumn{4}{c|}{\textbf{Benign Example}} & \multicolumn{10}{c||}{\textbf{White-Box Attacks for Robustness Evaluation}} & \tabularnewline
\cline{5-18} \cline{6-18} \cline{7-18} \cline{8-18} \cline{9-18} \cline{10-18} \cline{11-18} \cline{12-18} \cline{13-18} \cline{14-18} \cline{15-18} \cline{16-18} \cline{17-18} \cline{18-18} 
\rowcolor{SteelBlue!33} \multirow{-2}{*}{\textbf{Dataset}}& \multirow{-2}{*}{\textbf{Model}}  & \multirow{-2}{*}{\textbf{Loss}}  & \multirow{-2}{*}{\textbf{Defense}}  & R@1 $\uparrow$ & R@2 $\uparrow$ & mAP $\uparrow$ & NMI $\uparrow$ & CA+ $\uparrow$ & CA-\textbf{ $\downarrow$} & QA+ $\uparrow$ & QA-\textbf{ $\downarrow$} & TMA\textbf{ $\downarrow$} & ES:D\textbf{ $\downarrow$} & ES:R $\uparrow$ & LTM $\uparrow$ & GTM $\uparrow$ & GTT $\uparrow$ & \multirow{-2}{*}{\textbf{ERS} $\uparrow$}\tabularnewline

\midrule

\multirow{2}{*}{MNIST} & \multirow{2}{*}{C2F2} & \multirow{2}{*}{Triplet} & $\checkmark$ (FGSM) & 98.5 & 99.1 & 95.2 & 93.2 & 3.3 & 55.8 & 3.4 & 71.5 & 0.967 & 0.633 & 63.2 & 22.8 & 8.4 & 0.0 & 25.2\tabularnewline
 &  &  & $\bigstar$ (FGSM) & 98.9 & 99.3 & 98.8 & 92.9 & 9.9 & 46.1 & 11.1 & 50.7 & 0.709 & 1.241 & 10.0 & 66.4 & 25.9 & 0.0 & 31.5\tabularnewline
\hline 
\multirow{2}{*}{Fashion} & \multirow{2}{*}{C2F2} & \multirow{2}{*}{Triplet} & $\checkmark$ (FGSM) & 83.6 & 89.9 & 71.8 & 69.1 & 2.5 & 74.5 & 2.3 & 83.3 & 0.980 & 1.037 & 16.2 & 7.2 & 10.3 & 0.0 & 13.5\tabularnewline
 &  &  & $\bigstar$ (FGSM) & 83.7 & 90.1 & 77.0 & 74.3 & 8.5 & 63.2 & 11.1 & 71.5 & 0.804 & 1.357 & 7.2 & 22.0 & 17.1 & 0.0 & 20.3\tabularnewline
\hline 
\multirow{2}{*}{CUB} & \multirow{2}{*}{RN18} & \multirow{2}{*}{Triplet} & $\checkmark$ (FGSM) & 47.7 & 60.3 & 23.6 & 57.7 & 0.0 & 100.0 & 0.0 & 99.5 & 0.918 & 1.729 & 0.0 & 0.0 & 12.8 & 0.0 & 3.5\tabularnewline
 &  &  & $\bigstar$ (FGSM) & 31.4 & 41.9 & 14.1 & 46.3 & 6.6 & 70.2 & 4.9 & 68.0 & 0.598 & 1.184 & 5.6 & 3.8 & 13.6 & 0.1 & 18.9\tabularnewline
\hline 
\multirow{2}{*}{CARS} & \multirow{2}{*}{RN18} & \multirow{2}{*}{Triplet} & $\checkmark$ (FGSM) & 62.3 & 73.9 & 22.3 & 55.5 & 0.1 & 100.0 & 0.1 & 99.5 & 0.905 & 1.762 & 0.0 & 0.0 & 12.2 & 0.0 & 3.4\tabularnewline
 &  &  & $\bigstar$ (FGSM) & 43.6 & 54.9 & 12.1 & 42.3 & 9.1 & 50.3 & 8.5 & 54.4 & 0.495 & 0.989 & 8.5 & 7.0 & 17.3 & 0.4 & 26.5\tabularnewline
\hline 
\multirow{2}{*}{SOP} & \multirow{2}{*}{RN18} & \multirow{2}{*}{Triplet} & $\checkmark$ (FGSM) & 58.4 & 63.8 & 34.8 & 86.6 & 1.3 & 97.3 & 0.3 & 95.4 & 0.837 & 1.487 & 0.0 & 0.0 & 5.6 & 0.0 & 5.8\tabularnewline
 &  &  & $\bigstar$ (FGSM) & 53.6 & 59.0 & 30.3 & 85.7 & 15.9 & 20.4 & 14.0 & 21.2 & 0.348 & 0.668 & 12.1 & 13.3 & 20.2 & 8.9 & 40.5\tabularnewline

\bottomrule

\end{tabular}}
\end{table*}

\textcolor{blue}{
Recall that Madry defense~\cite{madry} involves creating adversarial examples
using PGD at every iteration during the adversarial training process.
As PGD involves multiple times
of model forward and backward propagation, the time consumption for creating
adversarial examples can be much higher than training a vanilla model.
Hence, we replace the PGD algorithm with FGSM~\cite{fgsm}, namely its single-iteration
version (much faster) and evaluate the defense methods accordingly.}

\textcolor{blue}{
As shown in \tabname~\ref{tab:fgsm}, defense with
FGSM leads to a better performance on benign examples, but meanwhile
a lower robustness compared to those with PGD in \tabname~\ref{tab:rob}.
Since FGSM is known to be much weaker than PGD in effect of attack,
we speculate the reason is that the adversarial attack failing to achieve its goal
(\eg, ``collapse'' the positive and negative samples) results in
inefficient learning of robust representations.
Thus, similar to the discussion on different triplet samplers,
the effectiveness of attack for our defense
is an important factor that impacts robustness.
}

\section{Conclusion}
\label{sec:8}

\textcolor{blue}{
Deep ranking models are vulnerable to adversarial perturbations that could
intentionally change the ranking result.
In this paper, we define and implement \emph{adversarial ranking attack}
(including candidate attack and query attack) that can compromise
deep ranking models.
We also propose two adversarial ranking defense methods, namely embedding-shifted triplet and anti-collapse triplet.
The first defense method can significantly suppress embedding shift distance and moderately
improve the ranking model robustness.
Being free of problems identified from the first defense including
misleading gradient and insufficient mini-batch exploitation, the second
defense achieves significant improvement in adversarial robustness and
generalization performance.
Moreover, we propose an empirical robustness
score to comprehensively evaluate a defense,
which involves a wide range of representative attacks.}

\textcolor{blue}{
In the potential of future works, we may investigate
(1) whether transferability or universal perturbation still exist in models
with a strong defense;
(2) whether black-box attacks are still possible against a model with a strong defense;
(3) defense and robustness with other metric learning loss functions;
Moreover, as the current adversarial training methods
still suffer from (1) considerable training time cost, (2) performance drop
on benign examples, and (3) insufficient robustness against adversarial attacks,
it is still very important to seek for a better defense for future work.
}

% or


%\appendices

%\section{}

\ifCLASSOPTIONcompsoc
  \section*{Acknowledgments}
\else
  \section*{Acknowledgment}
\fi

This work was supported partly by National Key R\&D Program of China Grant
2018AAA0101400, NSFC Grants 62088102 and 61976171, and Young Elite Scientists
Sponsorship Program by CAST Grant 2018QNRC001.

\ifCLASSOPTIONcaptionsoff
  \newpage
\fi



\bibliographystyle{IEEEtran}
\bibliography{IEEEabrv,bare_jrnl_compsoc}

\begin{thebibliography}{10}
\providecommand{\url}[1]{#1}
\csname url@samestyle\endcsname
\providecommand{\newblock}{\relax}
\providecommand{\bibinfo}[2]{#2}
\providecommand{\BIBentrySTDinterwordspacing}{\spaceskip=0pt\relax}
\providecommand{\BIBentryALTinterwordstretchfactor}{4}
\providecommand{\BIBentryALTinterwordspacing}{\spaceskip=\fontdimen2\font plus
\BIBentryALTinterwordstretchfactor\fontdimen3\font minus
  \fontdimen4\font\relax}
\providecommand{\BIBforeignlanguage}[2]{{%
\expandafter\ifx\csname l@#1\endcsname\relax
\typeout{** WARNING: IEEEtran.bst: No hyphenation pattern has been}%
\typeout{** loaded for the language `#1'. Using the pattern for}%
\typeout{** the default language instead.}%
\else
\language=\csname l@#1\endcsname
\fi
#2}}
\providecommand{\BIBdecl}{\relax}
\BIBdecl

\bibitem{resnet}
K.~He, X.~Zhang, S.~Ren, and J.~Sun, ``Deep residual learning for image
  recognition,'' in \emph{Proc. IEEE Conf. Comput. Vis. Pattern Recognit.},
  2016, pp. 770--778.

\bibitem{l-bfgs}
C.~Szegedy, W.~Zaremba, I.~Sutskever, J.~Bruna, D.~Erhan, I.~Goodfellow, and
  R.~Fergus, ``Intriguing properties of neural networks,'' in \emph{Proc. Int.
  Conf. Learn. Representations}, 2014.

\bibitem{fgsm}
I.~J. Goodfellow, J.~Shlens, and C.~Szegedy, ``Explaining and harnessing
  adversarial examples,'' in \emph{Proc. Int. Conf. Learn. Representations},
  2015.

\bibitem{inceptionv2}
C.~Szegedy, V.~Vanhoucke, S.~Ioffe, J.~Shlens, and Z.~Wojna, ``Rethinking the
  inception architecture for computer vision,'' in \emph{Proc. IEEE Conf.
  Comput. Vis. Pattern Recognit.}, 2016, pp. 2818--2826.

\bibitem{faceblack}
Y.~Dong, H.~Su, B.~Wu, Z.~Li, W.~Liu, T.~Zhang, and J.~Zhu, ``Efficient
  decision-based black-box adversarial attacks on face recognition,'' in
  \emph{Proc. IEEE Conf. Comput. Vis. Pattern Recognit.}, 2019, pp. 7714--7722.

\bibitem{phy-crime}
M.~Sharif, S.~Bhagavatula, L.~Bauer, and M.~K. Reiter, ``Accessorize to a
  crime: Real and stealthy attacks on state-of-the-art face recognition,'' in
  \emph{Proc. ACM SIGSAC Conf. Comput. Communications Security}, 2016, pp.
  1528--1540.

\bibitem{advpattern}
Z.~Wang, S.~Zheng, M.~Song, Q.~Wang, A.~Rahimpour, and H.~Qi, ``Advpattern:
  Physical-world attacks on deep person re-identification via adversarially
  transformable patterns,'' in \emph{Proc. Int. Conf. Comput. Vis.}, 2019, pp.
  8341--8350.

\bibitem{imagesimilarity}
G.~Chechik, V.~Sharma, U.~Shalit, and S.~Bengio, ``Large scale online learning
  of image similarity through ranking,'' \emph{J. Mach. Learn. Research},
  vol.~11, no.~36, pp. 1109--1135, Mar. 2010.

\bibitem{imagesim2}
J.~Wang, Y.~Song, T.~Leung, C.~Rosenberg, J.~Wang, J.~Philbin, B.~Chen, and
  Y.~Wu, ``Learning fine-grained image similarity with deep ranking,'' in
  \emph{Proc. IEEE Conf. Comput. Vis. Pattern Recognit.}, 2014, pp. 1386--1393.

\bibitem{madry}
A.~Madry, A.~Makelov, L.~Schmidt, D.~Tsipras, and A.~Vladu, ``Towards deep
  learning models resistant to adversarial attacks,'' in \emph{Proc. Int. Conf.
  Learn. Representations}, 2018.

\bibitem{facenet}
F.~Schroff, D.~Kalenichenko, and J.~Philbin, ``Facenet: A unified embedding for
  face recognition and clustering,'' in \emph{Proc. IEEE Conf. Comput. Vis.
  Pattern Recognit.}, 2015, pp. 815--823.

\bibitem{bugfeature}
A.~Ilyas, S.~Santurkar, D.~Tsipras, L.~Engstrom, B.~Tran, and A.~Madry,
  ``Adversarial examples are not bugs, they are features,'' in \emph{Proc.
  Conf. Neural Inf. Process. Syst.}, 2019, pp. 125--136.

\bibitem{advrank}
M.~Zhou, Z.~Niu, L.~Wang, Q.~Zhang, and G.~Hua, ``Adversarial ranking attack
  and defense,'' in \emph{Proc. Eur. Conf. Comput. Vis.}, 2020, pp. 781--799.

\bibitem{revisiting}
K.~Roth, T.~Milbich, S.~Sinha, P.~Gupta, B.~Ommer, and J.~P. Cohen,
  ``Revisiting training strategies and generalization performance in deep
  metric learning,'' in \emph{Proc. Int. Conf. Mach. Learn.}, 2020, pp.
  8242--8252.

\bibitem{dmlreality}
K.~Musgrave, S.~Belongie, and S.-N. Lim, ``A metric learning reality check,''
  in \emph{Proc. Eur. Conf. Comput. Vis.}, 2020, pp. 681--699.

\bibitem{hm-lstm}
Z.~Niu, M.~Zhou, L.~Wang, X.~Gao, and G.~Hua, ``Hierarchical multimodal lstm
  for dense visual-semantic embedding,'' in \emph{Proc. Int. Conf. Comput.
  Vis.}, 2017, pp. 1881--1889.

\bibitem{ladderloss}
M.~Zhou, Z.~Niu, L.~Wang, Z.~Gao, Q.~Zhang, and G.~Hua, ``Ladder loss for
  coherent visual-semantic embedding,'' in \emph{Proc. AAAI. Conf. Artif.
  Intell.}, 2020, pp. 13\,050--13\,057.

\bibitem{9018082}
L.~Zhang, Z.~Shi, J.~T. Zhou, M.-M. Cheng, Y.~Liu, J.-W. Bian, Z.~Zeng, and
  C.~Shen, ``Ordered or orderless: A revisit for video based person
  re-identification,'' \emph{IEEE Trans. Pattern Anal. Mach. Intell.}, vol.~43,
  no.~4, pp. 1460--1466, 2021.

\bibitem{LTR}
T.-Y. Liu, ``Learning to rank for information retrieval,'' \emph{Found. Trends
  Inf. Retr.}, vol.~3, no.~3, pp. 225--331, Mar. 2009.

\bibitem{sop}
H.~Oh~Song, Y.~Xiang, S.~Jegelka, and S.~Savarese, ``Deep metric learning via
  lifted structured feature embedding,'' in \emph{Proc. IEEE Conf. Comput. Vis.
  Pattern Recognit.}, 2016, pp. 4004--4012.

\bibitem{distance}
C.-Y. Wu, R.~Manmatha, A.~J. Smola, and P.~Krahenbuhl, ``Sampling matters in
  deep embedding learning,'' in \emph{Proc. Int. Conf. Comput. Vis.}, 2017, pp.
  2840--2848.

\bibitem{i-fgsm}
A.~Kurakin, I.~Goodfellow, and S.~Bengio, ``Adversarial examples in the
  physical world,'' in \emph{Proc. Int. Conf. Learn. Representations
  Workshops}, 2017.

\bibitem{cw}
N.~Carlini and D.~Wagner, ``Towards evaluating the robustness of neural
  networks,'' in \emph{Proc. IEEE Symposium on Security and Privacy}, 2017, pp.
  39--57.

\bibitem{apgd}
F.~Croce and M.~Hein, ``Reliable evaluation of adversarial robustness with an
  ensemble of diverse parameter-free attacks,'' in \emph{Proc. Int. Conf. Mach.
  Learn.}, 2020, pp. 2206--2216.

\bibitem{lafeat}
Y.~Yu, X.~Gao, and C.-Z. Xu, ``Lafeat: Piercing through adversarial defenses
  with latent features,'' in \emph{Proc. IEEE Conf. Comput. Vis. Pattern
  Recognit.}, 2021.

\bibitem{8807315}
S.~Tang, X.~Huang, M.~Chen, C.~Sun, and J.~Yang, ``Adversarial attack type i:
  Cheat classifiers by significant changes,'' \emph{IEEE Trans. Pattern Anal.
  Mach. Intell.}, vol.~43, no.~3, pp. 1100--1109, 2021.

\bibitem{curlswhey}
Y.~Shi, S.~Wang, and Y.~Han, ``Curls \& whey: Boosting black-box adversarial
  attacks,'' in \emph{Proc. IEEE Conf. Comput. Vis. Pattern Recognit.}, 2019,
  pp. 6519--6527.

\bibitem{di-fgsm}
C.~Xie, Z.~Zhang, Y.~Zhou, S.~Bai, J.~Wang, Z.~Ren, and A.~L. Yuille,
  ``Improving transferability of adversarial examples with input diversity,''
  in \emph{Proc. IEEE Conf. Comput. Vis. Pattern Recognit.}, 2019, pp.
  2730--2739.

\bibitem{ti-fgsm}
Y.~Dong, T.~Pang, H.~Su, and J.~Zhu, ``Evading defenses to transferable
  adversarial examples by translation-invariant attacks,'' in \emph{Proc. IEEE
  Conf. Comput. Vis. Pattern Recognit.}, 2019, pp. 4312--4321.

\bibitem{ila}
Q.~Huang, I.~Katsman, H.~He, Z.~Gu, S.~Belongie, and S.-N. Lim, ``Enhancing
  adversarial example transferability with an intermediate level attack,'' in
  \emph{Proc. Int. Conf. Comput. Vis.}, 2019, pp. 4733--4742.

\bibitem{universal}
S.-M. Moosavi-Dezfooli, A.~Fawzi, O.~Fawzi, and P.~Frossard, ``Universal
  adversarial perturbations,'' in \emph{Proc. IEEE Conf. Comput. Vis. Pattern
  Recognit.}, 2017, pp. 1765--1773.

\bibitem{unsupuap}
H.~Liu, R.~Ji, J.~Li, B.~Zhang, Y.~Gao, Y.~Wu, and F.~Huang, ``Universal
  adversarial perturbation via prior driven uncertainty approximation,'' in
  \emph{Proc. Int. Conf. Comput. Vis.}, 2019, pp. 2941--2949.

\bibitem{benchmarking}
Y.~Dong, Q.-A. Fu, X.~Yang, T.~Pang, H.~Su, Z.~Xiao, and J.~Zhu, ``Benchmarking
  adversarial robustness on image classification,'' in \emph{Proc. IEEE Conf.
  Comput. Vis. Pattern Recognit.}, 2020, pp. 321--331.

\bibitem{phy-synth}
A.~Athalye, L.~Engstrom, A.~Ilyas, and K.~Kwok, ``Synthesizing robust
  adversarial examples,'' in \emph{Proc. Int. Conf. Mach. Learn.}, pp.
  284--293.

\bibitem{phy-robust}
K.~Eykholt, I.~Evtimov, E.~Fernandes, B.~Li, A.~Rahmati, C.~Xiao, A.~Prakash,
  T.~Kohno, and D.~Song, ``Robust physical-world attacks on deep learning
  models,'' in \emph{Proc. IEEE Conf. Comput. Vis. Pattern Recognit.}, 2018,
  pp. 1625--1634.

\bibitem{advrank-doc}
G.~Goren, O.~Kurland, M.~Tennenholtz, and F.~Raiber, ``Ranking robustness under
  adversarial document manipulations,'' in \emph{Proc. ACM SIGIR Conf. Research
  \& Development Inf. Retrieval}, 2018, pp. 395--404.

\bibitem{advrank-rec}
X.~He, Z.~He, X.~Du, and T.-S. Chua, ``Adversarial personalized ranking for
  recommendation,'' in \emph{Proc. ACM SIGIR Conf. Research \& Development Inf.
  Retrieval}, 2018, pp. 355--364.

\bibitem{universalret}
J.~Li, R.~Ji, H.~Liu, X.~Hong, Y.~Gao, and Q.~Tian, ``Universal perturbation
  attack against image retrieval,'' in \emph{Proc. Int. Conf. Comput. Vis.},
  2019, pp. 4899--4908.

\bibitem{flowertower}
G.~Tolias, F.~Radenovic, and O.~Chum, ``Targeted mismatch adversarial attack:
  Query with a flower to retrieve the tower,'' in \emph{Proc. Int. Conf.
  Comput. Vis.}, 2019, pp. 5037--5046.

\bibitem{learn-to-misrank}
H.~Wang, G.~Wang, Y.~Li, D.~Zhang, and L.~Lin, ``Transferable, controllable,
  and inconspicuous adversarial attacks on person re-identification with deep
  mis-ranking,'' in \emph{Proc. IEEE Conf. Comput. Vis. Pattern Recognit.},
  2020, pp. 342--351.

\bibitem{advorder}
M.~Zhou, L.~Wang, Z.~Niu, Q.~Zhang, Y.~Xu, N.~Zheng, and G.~Hua, ``Practical
  relative order attack in deep ranking,'' \emph{arXiv:2103.05248}, 2021.

\bibitem{qair}
X.~Li, J.~Li, Y.~Chen, S.~Ye, Y.~He, S.~Wang, H.~Su, and H.~Xue, ``Qair:
  Practical query-efficient black-box attacks for image retrieval,'' in
  \emph{Proc. IEEE Conf. Comput. Vis. Pattern Recognit.}, 2021.

\bibitem{adaptive}
F.~Tramer, N.~Carlini, W.~Brendel, and A.~Madry, ``On adaptive attacks to
  adversarial example defenses,'' in \emph{Proc. Conf. Neural Inf. Process.
  Syst.}, 2020, pp. 1633--1645.

\bibitem{obfuscated}
A.~Athalye, N.~Carlini, and D.~Wagner, ``Obfuscated gradients give a false
  sense of security: Circumventing defenses to adversarial examples,'' in
  \emph{Proc. Int. Conf. Mach. Learn.}, 2018, pp. 274--283.

\bibitem{distill2}
N.~Papernot, P.~McDaniel, X.~Wu, S.~Jha, and A.~Swami, ``Distillation as a
  defense to adversarial perturbations against deep neural networks,'' in
  \emph{Proc. IEEE Symposium on Security and Privacy}, 2016, pp. 582--597.

\bibitem{ensembleweak}
W.~He, J.~Wei, X.~Chen, N.~Carlini, and D.~Song, ``Adversarial example defense:
  Ensembles of weak defenses are not strong,'' in \emph{USENIX Workshop on
  Offensive Technologies}, 2017, p.~15.

\bibitem{deflecting}
A.~Prakash, N.~Moran, S.~Garber, A.~DiLillo, and J.~Storer, ``Deflecting
  adversarial attacks with pixel deflection,'' in \emph{Proc. IEEE Conf.
  Comput. Vis. Pattern Recognit.}, 2018, pp. 8571--8580.

\bibitem{magnet}
D.~Meng and H.~Chen, ``Magnet: A two-pronged defense against adversarial
  examples,'' in \emph{Proc. ACM SIGSAC Conf. Comput. Communications Security},
  2017, pp. 135--147.

\bibitem{nndef}
A.~Dubey, L.~v.~d. Maaten, Z.~Yalniz, Y.~Li, and D.~Mahajan, ``Defense against
  adversarial images using web-scale nearest-neighbor search,'' in \emph{Proc.
  IEEE Conf. Comput. Vis. Pattern Recognit.}, 2019, pp. 8767--8776.

\bibitem{adv-bnn}
X.~Liu, Y.~Li, C.~Wu, and C.-J. Hsieh, ``Adv-bnn: Improved adversarial defense
  through robust bayesian neural network,'' in \emph{Proc. Int. Conf. Learn.
  Representations}, 2019.

\bibitem{self-ensemble}
X.~Liu, M.~Cheng, H.~Zhang, and C.-J. Hsieh, ``Towards robust neural networks
  via random self-ensemble,'' in \emph{Proc. Eur. Conf. Comput. Vis.}, 2018,
  pp. 369--385.

\bibitem{featuredenoise}
C.~Xie, Y.~Wu, L.~v.~d. Maaten, A.~L. Yuille, and K.~He, ``Feature denoising
  for improving adversarial robustness,'' in \emph{Proc. IEEE Conf. Comput.
  Vis. Pattern Recognit.}, 2019, pp. 501--509.

\bibitem{advscale}
A.~Kurakin, I.~Goodfellow, and S.~Bengio, ``Adversarial machine learning at
  scale,'' in \emph{Proc. Int. Conf. Learn. Representations}, 2017.

\bibitem{bilateral}
J.~Wang and H.~Zhang, ``Bilateral adversarial training: Towards fast training
  of more robust models against adversarial attacks,'' in \emph{Proc. Int.
  Conf. Comput. Vis.}, 2019, pp. 6629--6638.

\bibitem{sparse}
F.~Croce and M.~Hein, ``Sparse and imperceivable adversarial attacks,'' in
  \emph{Proc. Int. Conf. Comput. Vis.}, 2019, pp. 4724--4732.

\bibitem{ead}
P.-Y. Chen, Y.~Sharma, H.~Zhang, J.~Yi, and C.-J. Hsieh, ``Ead: elastic-net
  attacks to deep neural networks via adversarial examples,'' in \emph{Proc.
  AAAI. Conf. Artif. Intell.}, 2018, pp. 10--17.

\bibitem{featureadversary}
S.~Sabour, Y.~Cao, F.~Faghri, and D.~J. Fleet, ``Adversarial manipulation of
  deep representations,'' in \emph{Proc. Int. Conf. Learn. Representations},
  2016.

\bibitem{metric1}
S.~Bai, Y.~Li, Y.~Zhou, Q.~Li, and P.~H. Torr, ``Adversarial metric attack and
  defense for person re-identification,'' \emph{IEEE Trans. Pattern Anal. Mach.
  Intell.}, vol.~43, no.~6, pp. 2119--2126, 2021.

\bibitem{advdpqn}
Y.~Feng, B.~Chen, T.~Dai, and S.-T. Xia, ``Adversarial attack on deep product
  quantization network for image retrieval,'' in \emph{Proc. AAAI. Conf. Artif.
  Intell.}, 2020, pp. 10\,786--10\,793.

\bibitem{mnist}
Y.~LeCun, L.~Bottou, Y.~Bengio, P.~Haffner \emph{et~al.}, ``Gradient-based
  learning applied to document recognition,'' \emph{Proceedings of the IEEE},
  vol.~86, no.~11, pp. 2278--2324, 1998.

\bibitem{fashion}
H.~Xiao, K.~Rasul, and R.~Vollgraf, ``Fashion-mnist: a novel image dataset for
  benchmarking machine learning algorithms,'' \emph{arXiv:1708.07747}, 2017.

\bibitem{cub200}
P.~Welinder, S.~Branson, T.~Mita, C.~Wah, F.~Schroff, S.~Belongie, and
  P.~Perona, ``Caltech-ucsd birds 200,'' California Institute of Technology,
  Tech. Rep. CNS-TR-2010-001, 2010.

\bibitem{cars196}
J.~Krause, M.~Stark, J.~Deng, and L.~Fei-Fei, ``3d object representations for
  fine-grained categorization,'' in \emph{Proc. Int. Conf. Comput. Vis.
  Workshops}, 2013, pp. 554--561.

\bibitem{pytorch}
A.~Paszke, S.~Gross, F.~Massa, A.~Lerer, J.~Bradbury, G.~Chanan, T.~Killeen,
  Z.~Lin, N.~Gimelshein, L.~Antiga, A.~Desmaison, A.~Kopf, E.~Yang, Z.~DeVito,
  M.~Raison, A.~Tejani, S.~Chilamkurthy, B.~Steiner, L.~Fang, J.~Bai, and
  S.~Chintala, ``Pytorch: An imperative style, high-performance deep learning
  library,'' in \emph{Proc. Conf. Neural Inf. Process. Syst.}, 2019, pp.
  8024--8035.

\bibitem{adam}
D.~P. Kingma and J.~Ba, ``Adam: A method for stochastic optimization,'' in
  \emph{Proc. Int. Conf. Learn. Representations}, 2015.

\bibitem{mnasnet}
M.~Tan, B.~Chen, R.~Pang, V.~Vasudevan, M.~Sandler, A.~Howard, and Q.~V. Le,
  ``Mnasnet: Platform-aware neural architecture search for mobile,'' in
  \emph{Proc. IEEE Conf. Comput. Vis. Pattern Recognit.}, 2019, pp. 2820--2828.

\end{thebibliography}

%\vskip -5em
\begin{IEEEbiography}[{\includegraphics[width=1in,height=1.25in,clip,keepaspectratio]{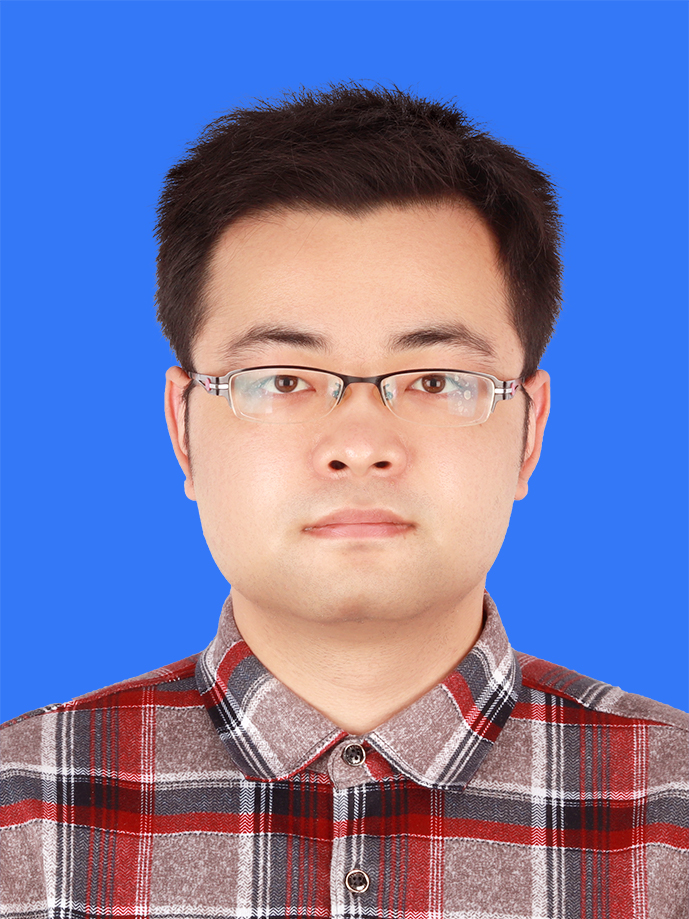}}]{Mo Zhou}
	received the B.S. degree in Electromagnetic Fields and Wireless Technology,
	and the M.S. degree in Pattern Recognition and Intelligent System
	from Xidian University, Xi'an, China, in 2017 and 2020.
	He is currently a Research Assistant with the Institute of Artificial Intelligence and Robotics of Xi'an Jiaotong University, Xi'an, China.
	His research interests include deep learning, computer vision, cross-modal retrieval, and adversarial attack and defense.
\end{IEEEbiography}
%
%\vskip -5em
\begin{IEEEbiography}[{\includegraphics[width=1in,height=1.25in,clip,keepaspectratio]{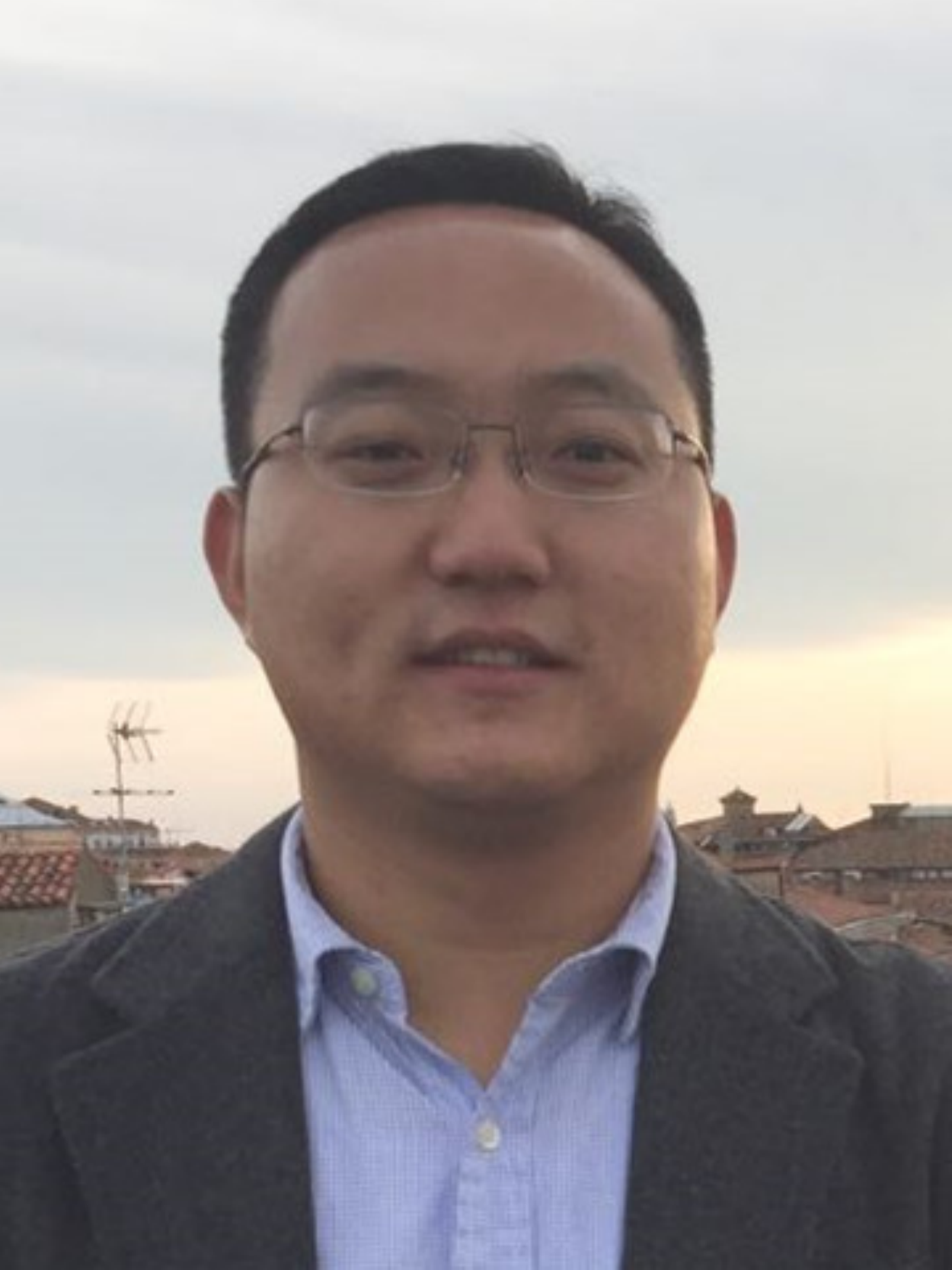}}]{Le Wang}
	(Senior Member, IEEE) received the B.S. and Ph.D. degrees in control science and engineering from Xi'an Jiaotong University, Xi'an, China, in 2008 and 2014, respectively. From 2013 to 2014, he was a Visiting Ph.D. Student with the Stevens Institute of Technology, Hoboken, New Jersey, USA. From 2016 to 2017, he was a Visiting Scholar with Northwestern University, Evanston, Illinois, USA. He is currently an Associate Professor with the Institute of Artificial Intelligence and Robotics, Xi'an Jiaotong University, Xi'an, China. His research interests include computer vision, pattern recognition, and machine learning. He is the author of more than 50 peer reviewed publications in prestigious international journals and conferences. He is an area chair of CVPR'2022.
\end{IEEEbiography}
%
%\vskip -5em
\begin{IEEEbiography}[{\includegraphics[width=1in,height=1.25in,clip,keepaspectratio]{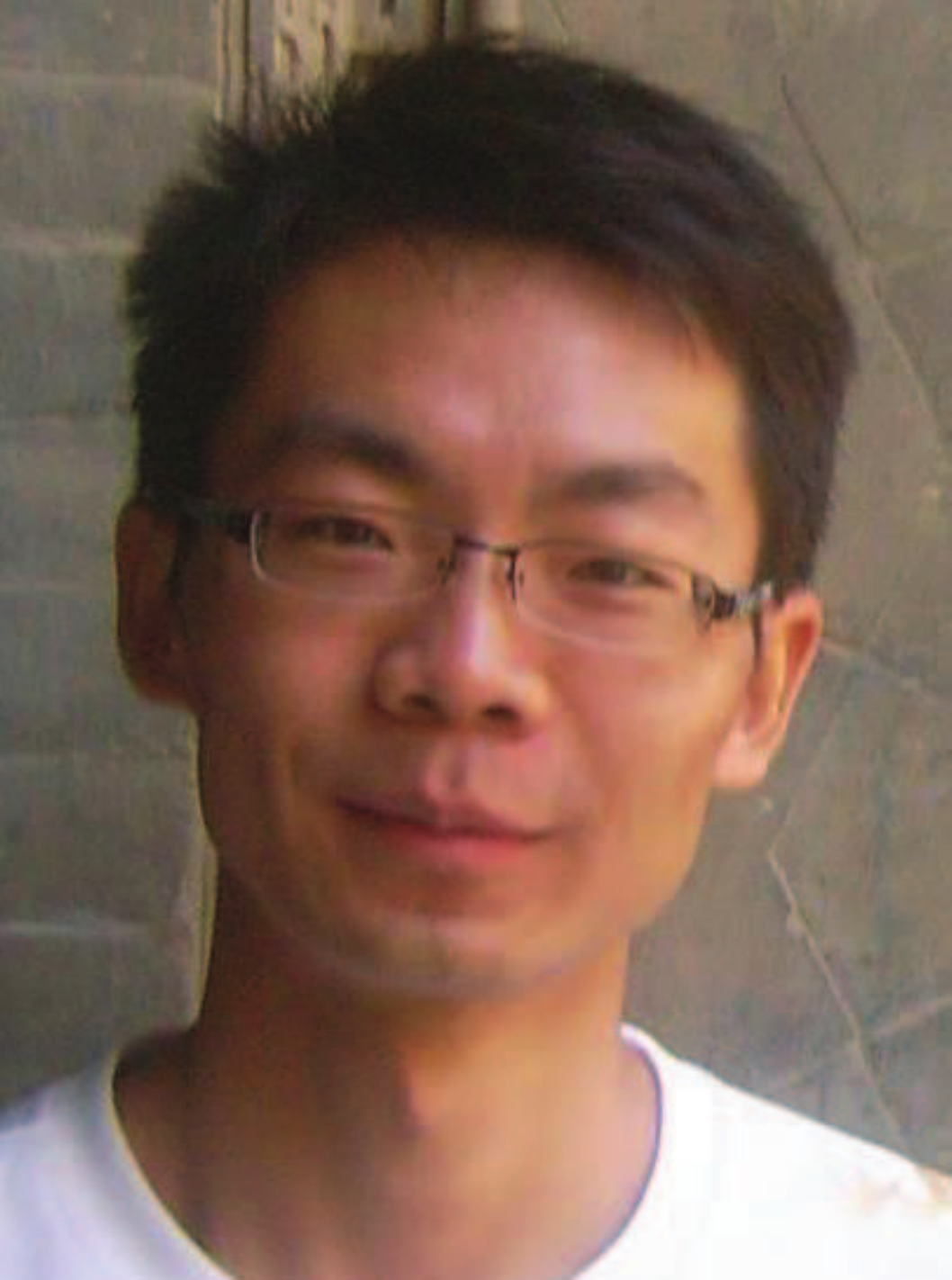}}]{Zhenxing Niu}
	(Member, IEEE) received the Ph.D. degree in Control Science and Engineering from Xidian University, Xi'an, China, in 2012. From 2013 to 2014, he was a visiting scholar with University of Texas at San Antonio, Texas, USA. He is a Researcher at Alibaba Group, Hangzhou, China. Before joining Alibaba Group, he is an Associate Professor of School of Electronic Engineering at Xidian University, Xi'an, China. His research interests include computer vision, machine learning, and their application in object discovery and localization. He served as PC member of CVPR, ICCV, and ACM Multimedia. He is an area chair of CVPR'2022.
\end{IEEEbiography}
\begin{IEEEbiography}[{\includegraphics[width=1in,height=1.25in,clip,keepaspectratio]{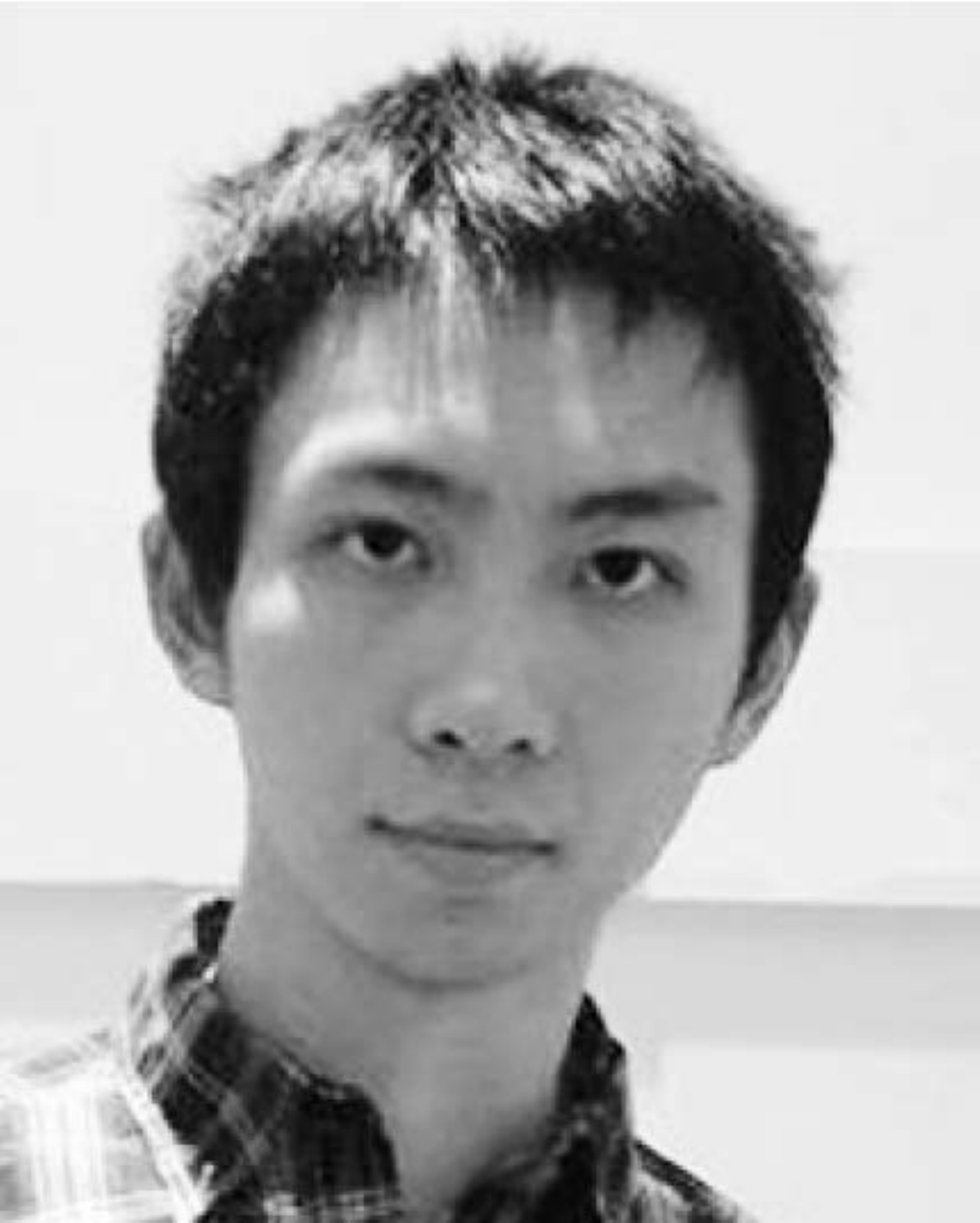}}]{Qilin Zhang}
	(Member, IEEE) received the B.E. degree in Electrical Information Engineering from the University of Science and Technology of China, Hefei, China, in 2009, and the M.S. degree in Electrical and Computer Engineering from University of Florida, Gainesville, Florida, USA, in 2011, and the Ph.D. degree in Computer Science from Stevens Institute of Technology, Hoboken, New Jersey, USA, in 2016.  He is currently a Senior Research Scientist at ABB Corporate Research Center in Raleigh, North Carolina, USA. His research interests include computer vision, machine learning and multimedia signal processing. He is the author of more than 30 peer reviewed publications in prestigious international journals and conferences.
\end{IEEEbiography}
%
%\vskip -5em
\begin{IEEEbiography}[{\includegraphics[width=1in,height=1.25in,clip,keepaspectratio]{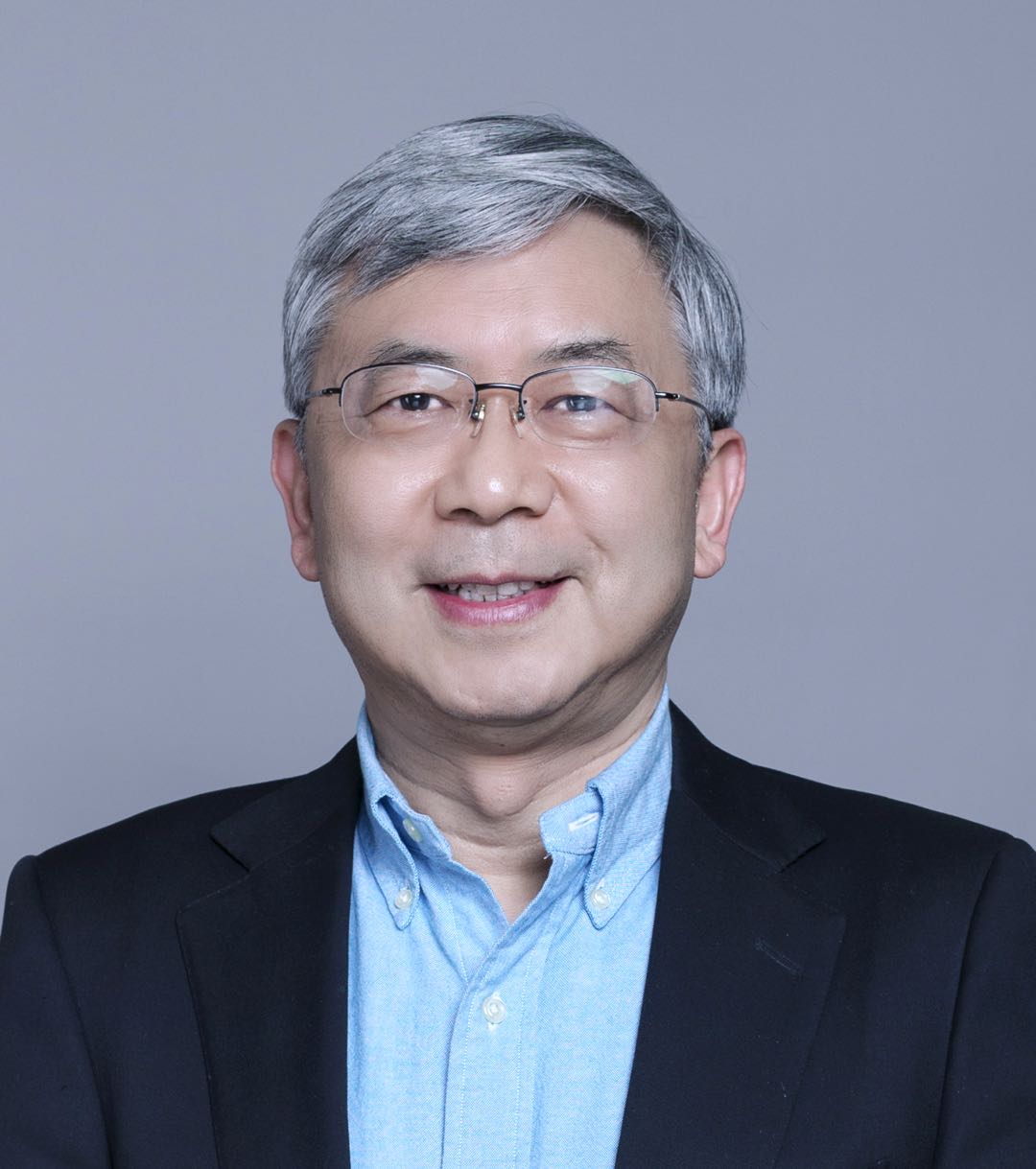}}]{Nanning Zheng}
	(Fellow, IEEE) graduated in 1975  from  the  Department  of  Electrical  Engineering,  Xi'an  Jiaotong  University (XJTU),  received the ME degree in Information and Control Engineering  from  Xi'an  Jiaotong  University  in 1981, and a Ph.D. degree in Electrical Engineering from Keio University in 1985. He is currently a Professor and the Director with the Institute of Artificial Intelligence and Robotics of Xi'an Jiaotong University.  His  research  interests  include  computer vision, pattern recognition, computational intelligence, and hardware implementation of intelligent systems. Since 2000, he has been the Chinese representative on the Governing Board of the International Association for Pattern Recognition. He became a member of the Chinese Academy Engineering in 1999. He is a fellow of the IEEE.
\end{IEEEbiography}
%
%\vskip -5em
\begin{IEEEbiography}[{\includegraphics[width=1in,height=1.25in,clip,keepaspectratio]{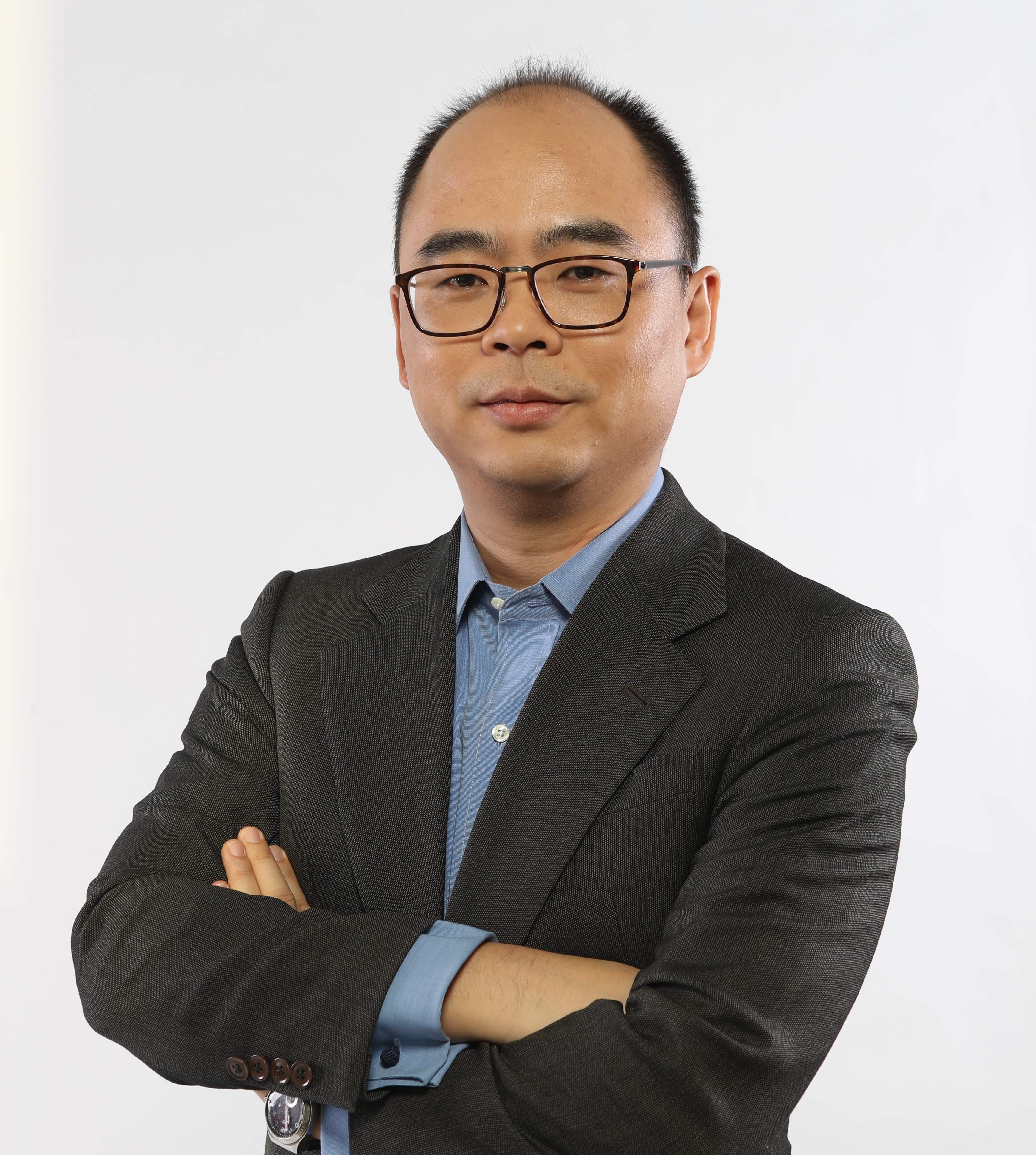}}]{Gang Hua}
(Fellow, IEEE) was enrolled in the Special Class for the Gifted Young of Xi'an Jiaotong University (XJTU), Xi'an, China, in 1994 and received the B.S. degree in Automatic Control Engineering from XJTU in 1999. He received the M.S. degree in Control Science and Engineering in 2002 from XJTU, and the Ph.D. degree in Electrical Engineering and Computer Science at Northwestern University, Evanston, Illinois, USA, in 2006. He is currently the Vice President and Chief Scientist of Wormpex AI Research. Before that, he served in various roles at Microsoft (2015-18) as the Science/Technical Adviser to the CVP of the Computer Vision Group, Director of Computer Vision Science Team in Redmond and Taipei ATL, and Principal Researcher/Research Manager at Microsoft Research. He was an Associate Professor at Stevens Institute of Technology (2011-15). During 2014-15, he took an on leave and worked on the Amazon-Go project. He was an Visiting Researcher (2011-14) and a Research Staff Member (2010-11) at IBM Research T. J. Watson Center, a Senior Researcher (2009-10) at Nokia Research Center Hollywood, and a Scientist (2006-09) at Microsoft Live labs Research. He is an associate editor of TIP, TCSVT, CVIU, IEEE Multimedia, TVCJ and MVA. He also served as the Lead Guest Editor on two special issues in TPAMI and IJCV, respectively. He is a general chair of ICCV'2025. He is a program chair of CVPR'2019\&2022. He is an area chair of CVPR'2015\&2017, ICCV'2011\&2017, ICIP'2012\&2013\&2016, ICASSP'2012\&2013, and ACM MM 2011\&2012\&2015\&2017. He is the author of more than 200 peer reviewed publications in prestigious international journals and conferences. He holds 19 US patents and has 15 more US patents pending. He is the recipient of the 2015 IAPR Young Biometrics Investigator Award for his contribution on Unconstrained Face Recognition from Images and Videos, and a recipient of the 2013 Google Research Faculty Award. He is an IEEE Fellow, an IAPR Fellow, and an ACM Distinguished Scientist.
\end{IEEEbiography}

%\newpage


%\vfill


\end{document}